\documentclass[10pt, journal, compsoc, cspaper, twoside, twocolumn]{IEEEtran}

\usepackage{hyperref}       
\ifCLASSOPTIONcompsoc
  \usepackage[nocompress]{cite}
\else
  \usepackage{cite}
\fi
\ifCLASSINFOpdf
   \usepackage[pdftex]{graphicx}
\else
   \usepackage[dvips]{graphicx}
\fi

\usepackage{amsmath}
\usepackage{gensymb}

\usepackage{algorithmic}
\usepackage{algorithm}

\usepackage{makecell}
\usepackage{diagbox}
\usepackage{xcolor}
\usepackage{booktabs} 
\usepackage{multirow}
\usepackage{threeparttable}
\usepackage{array}

\ifCLASSOPTIONcompsoc
  \usepackage[caption=false,font=footnotesize,labelfont=sf,textfont=sf]{subfig}
\else
  \usepackage[caption=false,font=footnotesize]{subfig}
\fi

\usepackage[]{footmisc}
\usepackage{fixltx2e}

\usepackage{stfloats}
\usepackage{url}

\newtheorem{corollary}{Corollary}
\newtheorem{conjecture}{Conjecture}
\newtheorem{theorem}{Theorem}
\newtheorem{property}{P}
\newtheorem{hypothese}{H}
\newtheorem{condition}{Condition}

\begin{document}
\title{Respecting Time Series Properties \\ Makes Deep Time Series Forecasting Perfect}

\author{Li~Shen, Yuning~Wei and Yangzhu~Wang
\IEEEcompsocitemizethanks{\IEEEcompsocthanksitem Li Shen, Yuning Wei and Yangzhu Wang were with Beihang University, Beijing, China. Email:\{shenli, yuning, wangyangzhu\}@buaa.edu.cn \protect\\
}
\thanks{Manuscript received (Corresponding authors: Yuning Wei.) \\ Digital Object Identifier no.}}

%
%

\markboth{IEEE Transactions on Knowledge and Data Engineering}%
{Shen et al.: Respecting Time Series Properties Makes Deep Time Series Forecasting Perfect}

\IEEEtitleabstractindextext{%
\begin{abstract}
How to handle time features shall be the core question of any time series forecasting model. Ironically, it is often ignored or misunderstood by deep-learning based models, even those baselines which are state-of-the-art. This behavior makes their inefficient, untenable and unstable. In this paper, we rigorously analyze three prevalent but deficient/unfounded deep time series forecasting mechanisms or methods from the view of time series properties, including normalization methods, multivariate forecasting and input sequence length. Corresponding corollaries and solutions are given on both empirical and theoretical basis. We thereby propose a novel time series forecasting network, i.e. RTNet, on the basis of aforementioned analysis. It is general enough to be combined with both supervised and self-supervised forecasting format. Thanks to the core idea of respecting time series properties, no matter in which forecasting format, RTNet shows obviously superior forecasting performances compared with dozens of other SOTA time series forecasting baselines in three real-world benchmark datasets. By and large, it even occupies less time complexity and memory usage while acquiring better forecasting accuracy. The source code is available at \url{https://github.com/OrigamiSL/RTNet}.
\end{abstract}

\begin{IEEEkeywords}
Time series forecasting, deep learning, contrastive learning
\end{IEEEkeywords}}

\maketitle

\IEEEdisplaynontitleabstractindextext
\IEEEpeerreviewmaketitle

\IEEEraisesectionheading{\section{Introduction}}
\IEEEPARstart{T}{he} research of deep time series forecasting is meeting a rapid development period. Diverse networks, including CNN \cite{scinet,TCN}, RNN \cite{LSTM_TKDE,deepar,LSTNet}, Transformer \cite{informer,logtrans}, GNN \cite{GraphAttention_TKDE,StemGNN,MTGNN}, etc., are employed to seek high time series forecasting accuracy. Mature techniques or mechanisms from other AI fields are also borrowed to further improve their performances \cite{BN,LN,frequencyTKDE,simclr}. However, with more and more state-of-the-art methods joint in time series forecasting, we shall also question whether they are really suitable for time series forecasting or whether they should be transformed before employment. Hardly does any deep time series forecasting method ever discuss or even notice this question, which makes their forecasting performances imperfect. Therefore, we believe that it is time to complement this research blank space, i.e. how to respect time series properties while doing time series forecasting.\par
Respecting time series properties actually shall be the core idea of any time series forecasting model. So how to reveal it in practice? Is it equal to force model to learn disentangled seasonal and trend representations \cite{trendbook,trend,N_BEATS, Cost}? Or is it equal to designing stronger extractor to learn deeper time features \cite{TFT, LSTNet} or so-called `long-term dependency' \cite{scinet, informer}? The results of this paper illustrate that methods above are fair but far from enough. They either only take ideal but not practical time series forecasting conditions into consideration or deal with time series forecasting problems only from the point of view of properties of neural networks. However, real-world time series forecasting situations are always much more complicated than imagined. They are often non-stationary, unbounded and composed of complex anomalies. Although those forecasting networks are much stronger than traditional forecasting methods, their performances are far from their full potentials in virtue of their ignorance of time series properties.\par
In this paper, we will first list some long-established time series properties and make some rational hypotheses. Then:\par
\begin{enumerate}
	\item \label{point a} We rigorously analyze three widely-used but deficient/unfounded deep time series forecasting mechanisms, including normalization methods, multivariate forecasting and input sequence length. Several corollaries are drawn to point out their application conditions/methods in time series forecasting.\par
	\item We propose RTNet, a novel `\emph{\textbf{R}}especting \emph{\textbf{T}}ime' time series forecasting \emph{\textbf{N}}etwork. It is simple, efficient and general enough to couple with both end-to-end (supervised) and contrastive learning based (self-supervised) forecasting formats.
	Besides corollaries mentioned in Point \ref{point a}, it respects time series properties also in many other network details.
	\item We perform validation experiments to empirically examine our proposed corollaries and ideas. Moreover, abundant and diverse experiments on three benchmark datasets demonstrate the outstanding forecasting capability of RTNet even compared with dozens of SOTA baselines. 
\end{enumerate}

\section{Respect Time Series Properties}
\label{section2}
Firstly, some properties of time series and reasonable hypotheses are shown below as prerequisites of following statements. P\ref{p1}-P\ref{p2} are properties and H\ref{h1}-H\ref{h4} are hypotheses.\par
\begin{property}
	\label{p1}
	Time series elements are not scale invariant in forecasting tasks based on the homogeneity of linear projection.
\end{property}
\par
\begin{property}
	\label{p2}
	Time series have intrinsic causality \cite{box1968,box2015}.
\end{property}
\begin{hypothese}
	\label{h1}
	There is no supremum/infimum for time series \cite{TS2Vec}.
\end{hypothese}
\begin{hypothese}
	\label{h2}
	Random anomalies always exist in time series \cite{AnomalyTransformer}.
\end{hypothese}
\begin{hypothese}
	\label{h3}
	Variates of multivariate time series may be irrelevant to some other variates.
\end{hypothese}
\begin{hypothese}
	\label{h4}
	Time series are partially auto-regressive and not always stationary in practice \cite{TNC}.
\end{hypothese}
\par
Based on these reasonable hypotheses and properties, we point out some common problems of time series forecasting networks and present corresponding schemes to solve them. \par
\subsection{Normalization}
\label{section2.1}
Some of time series forecasting models \cite{informer, logtrans, N_HiTS} apply normalization methods \cite{BN,LN} which are widely used in other AI fields to solve Internal Covariate Shift (ICS) problem. However, neither models applying normalization methods nor models without them explain why they adopt them or not. In fact, normalization methods are not all suitable for time series forecasting tasks. We analyze the most typical three normalization methods, i.e., Batch Normalization \cite{BN}, Layer Normalization \cite{LN} and Weight Normalization \cite{WN}, here to seek the most suitable kind of normalization formats for time series forecasting models.\par
\subsubsection{Batch Normalization (BN)}
Batch Normalization (BN) computes the normalization statistics over all the hidden units in each mini-batch $B$, which is presented in Eq.\ref{eq1}. $m$ denotes the size of mini-batch, $\mu_B$ is the mean of mini-batch and $\sigma_B^2$ is the variance. $\hat{x_i}$ is the normalized value and ${y}_i$ is the scaled and shifted value. $\gamma/\beta$ is the re-scaling/re-centering parameter to be learned.\par

\begin{align}
	\label{eq1}
	\mu_B &= \frac{1}{m} \sum\nolimits_{i\in[1,m]}{x_i},\ \sigma_B^2 = \frac{1}{m} \sum\nolimits_{i\in[1,m]}{(x_i - \mu_B)^2}  \notag\\
	\hat{x}_i &= \frac{x_i - \mu_B}{\sqrt{\sigma_B^2 + \epsilon}},\	{y}_i = \gamma \hat{x}_i + \beta
\end{align}

\par
From Eq.\ref{eq1}, it can be deduced that the basic prerequisite of applying BN to a specific neural network is that the scale disparity of the hidden units in the same position among different instances of any batch can be ignored. Nevertheless, based on P\ref{p1}, time series forecasting tasks do not satisfy this prerequisite in that the scale disparity cannot be neglected. Actually, based on H\ref{h2} and H\ref{h4}, real-world time series are not always stationary, meaning that the statistical characteristics of a local time series window will change with time so that scale disparity always happens and even might be immense due to H\ref{h1}.\par
\subsubsection{Layer Normalization (LN)}
Similar to BN, Layer Normalization re-parameterizes the normalization statistics over all hidden units in the same layer. Its application prerequisite is also alike: the scale disparity of hidden units in the same layer can be ignored. So this prerequisite also fails to fit P\ref{p1}. Besides, another prerequisite of applying LN is that gain and bias parameters of hidden units in the same layer can be shared. Whereas, owing to P\ref{p2}, the application of LN will bring future information leakage problem if applied in causal architectures like Transformer decoder.\par
\subsubsection{Weight Normalization (WN)}
WN re-parameterizes weight vectors rather than hidden units like BN/LN. The weight vector $w$ is divided into a parameter unit vector $\hat{v}$ and a scalar parameter $g$ after applying WN as Eq.\ref{eq7}. $\hat{v}$ is composed of a vector $v$ and its Euclidean norm $\Vert v\Vert$. WN distinguishes itself from other two methods in not changing distributions of hidden units directly so that it is not invariant to data re-parameterization \cite{LN} and respects the causality of time series, i.e. it does not violates P\ref{p1} and P\ref{p2}.\par
\begin{eqnarray}
	\label{eq7}
	w=g\hat{v}=\frac{g}{\Vert v\Vert}v
\end{eqnarray}
\subsubsection{Conclusion}
So based on aforementioned observations and analysis, normalization methods not changing the statistical characteristics of hidden units directly seem to be more suitable for time series forecasting models. Additionally, we would like to further discuss if it is possible for normalization methods like BN/LN to be applied to time series forecasting models. Historically, it has been found out \cite{WN} that only when all inputs of a one-layer network are whitened, the function of BN is equivalent to that of WN. However, in time series forecasting tasks, this requirement implies that any element of any prediction window with any size is whitened. Then, it can be immediately deduced that time series shall be stationary white noise, which is impossible in real-world practice. Therefore, Corollary \ref{corollary1} could be deduced.
\begin{corollary}
	\label{corollary1}
	Normalization methods invariant to data re-parameterization do not fit time series forecasting tasks unless time series are strictly stationary.\par
\end{corollary}
\subsection{Multivariate Forecasting}
\label{section2.2}
To acquire predictions of $N$ variates in a specific dataset, most of state-of-the-art time series forecasting models introduce all $N$ variates $\{x_1,..,x_N\}$ of datasets together into the certain model for multivariate forecasting \cite{scinet,TS2Vec}. Unfortunately, forecasting results of them are sometimes even much worse than mean results of separate univariate forecasting for these $N$ variates. So here comes Corollary \ref{corollary2} and we prove it by Bayesian Information Criterion (BIC) \cite{BIC}.\par
\begin{corollary}
	\label{corollary2}
	Introducing any independent variate into the forecasting of target variate could not bring benefits.\par
\end{corollary}
\par
Without loss of generality and following H\ref{h3}, we select the first variate $x_1$ as forecasting target and assume that ${x_{i_1},\ldots,x_{i_j}} (i_j \in [1,n])$ are variates independent of $x_1$. BIC scores of certain model only forecasting $x_1$ are shown in Eq. \ref{eq8} before/after introduced mentioned independent variates $(B_1/B_2)$. $m$ is the length of total training data, $k_1/k_2$ refers to the multiplication of input sequence length and the number of input variate(s) and $L_1/L_2$ represents the likelihood function. As independent variates are added into the network, the number of parameters increases, i.e. $k_1\textless k_2$. Meanwhile, because the participation of independent variates will not improve the forecasting accuracy or even may worsen it, $-2ln(L_1)\leq -2ln(L_2)$. It can be observed that BIC score of certain model combined with independent variates will increase with high probability for the target variate forecasting. Then Corollary \ref{corollary2} could be verified. Note that Corollary \ref{corollary2} is still established with AIC \cite{AIC}.\par
\begin{eqnarray}
	\label{eq8}
	B_1 =ln(m)k_1-2ln(L_1),\ B_2 =ln(m)k_2-2ln(L_2)
\end{eqnarray}

\subsubsection{Solution: Cos-Relation Matrix}
Based on above investigation and proof, introducing independent variates may even bring worse effect for target variate forecasting. Hence, it is necessary to measure the relevance among these input variates and prevent influences brought by irrelevant variates. Borrowing ideas from Attention mechanism\cite{attention2017} and contrastive learning \cite{TS2Vec,simclr}, we utilize $\cos(u,v) = \vert u^\top v\vert/\Vert u\Vert\Vert v\Vert$ (i.e. absolute cosine similarity) to evaluate the relevance between variates $u$ and $v$ where their values at each time stamp are treated as their different features. Considering $N$ forecasting variates $x_i$ of sequence length $L_{in}$, we denote by $W(x)_{N\times N}$ the relation matrix of all variates where $w_{ij}$ measures the relevance between $x_i$ and $x_j$ (Eq.\ref{eq4}). We call it Cos-Relation Matrix.\par
\begin{eqnarray}
	\label{eq4}
	W(x)_{n\times n}=Mat(w_{ij})=Mat(cos(x_i,x_j))
\end{eqnarray}
After defining the relation matrix, we post-multiply input sequence tensor of each prediction window by it. Assume that batch size is 1 and the input sequence length is $L_{in}$, then the input tensor $S_{L_{in}\times n}$ will be transformed into $S_{L_{in}\times n}'$. It maintains the dimensions, however each variate sequence is turned into the weighted sum of all variate sequences. Then they are supposed to be delivered to separate independent network to do feature extraction and prediction works. The Corollary \ref{corollary3} below shows that variates with weaker relevance with any other specific variate will also donate smaller contributions during the forecasting of the same variate based on above methods.\par
\begin{figure}[]
	\centering
	\includegraphics[width=3.2in]{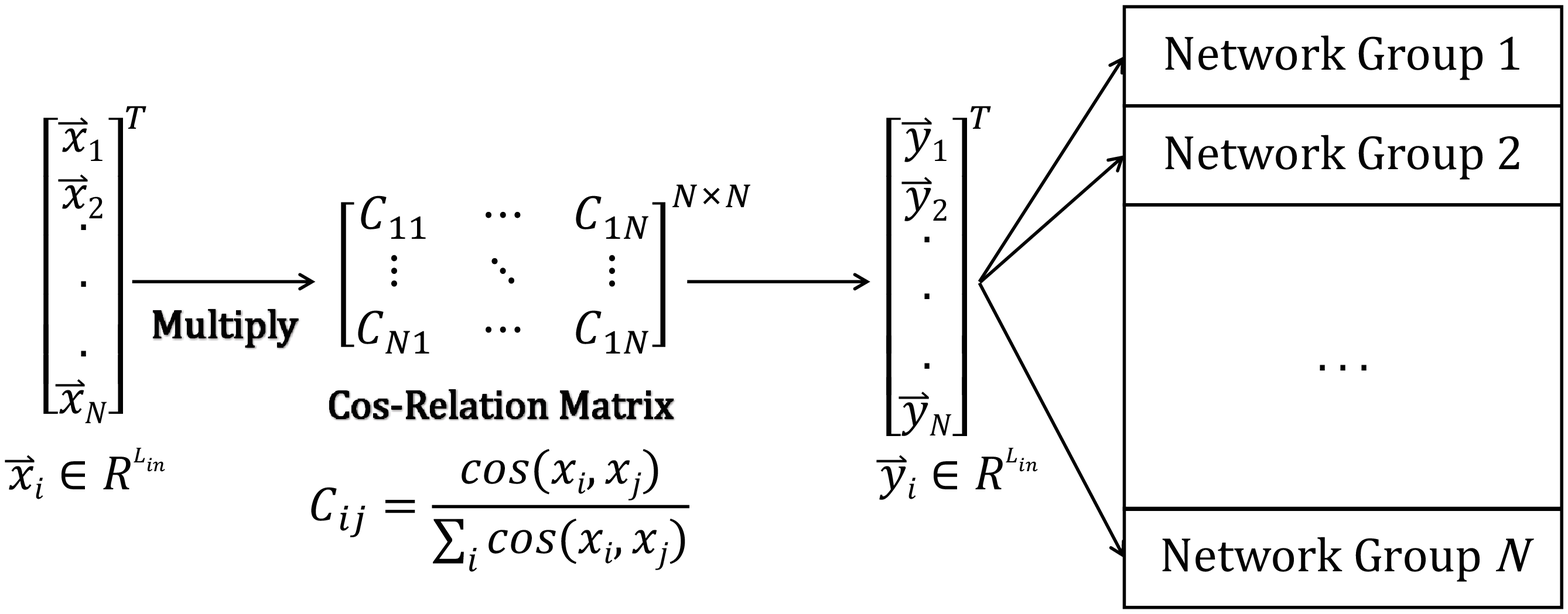}
	\caption{The usage of Cos-Relation Matrix $C_{ij}$. It multiplies input sequence tensor $x_i$ into $y_i$ which is the input of network group $i$ $(i \in [1,N])$.}
	\label{fig1}
\end{figure}
\begin{corollary}
	\label{corollary3}
	Ignoring non-linear parts of the network, the usage of relation matrix together with independent networks for each variate $W$ will make forecasting result of each variate contributed by all variates according to the relevance value in $W$.
\end{corollary}
The proof of Corollary \ref{corollary3} is evident. Ignoring the non-linear activation and getting rid of normalization methods re-parameterizing hidden units directly on the basis of Corollary \ref{corollary1}, the additivity and homogeneity of linear projection determines that each weight value $w_{ij}$ will maintain their proportions to the end. The concrete proof is shown in Appendix C3. In another view, Corollary \ref{corollary3} also illustrates the importance of selecting suitable normalization methods. The whole process mentioned above is shown in Fig. \ref{fig1}.\par
In real applications, two details are added to strengthen the function of relation matrix $W$. Firstly, after acquiring $W$ from the whole training data, we further standardize each column to avoid possible gradient explosion through dividing each $w_{ij}$ with the sum of its column, i.e. $\hat{w}_{ij}=w_{ij}/\sum_{i}w_{ij}$. In addition, a manually selected threshold $\theta$ is set to pad those relation values under $\cos\theta$ into zero so that irrelevant variates will totally not disturb the prediction of each other, i.e. $\hat{w}_{ij}=0$ if $\hat{w}_{ij} \textless \cos\theta$ else $\hat{w}_{ij}$.\par
\subsection{Input Sequence Length}
\label{section2.3}
The research of input sequence length occupies a large portion in the field of traditional forecasting methods, e.g. ARIMA \cite{box1968,box2015}, ES \cite{FES,ETSformer}. It profoundly influences the prediction results. However, hardly ever does any deep time series forecasting model pay much attention to it. In this section, we bend our efforts to answer two questions:
\begin{enumerate}
	\item Is it essential to keep input sequence length invariant throughout feature extraction?
	\item Does the input sequence length need to be longer with the prolonging of prediction length?
\end{enumerate}
\subsubsection{Invariance of Input Sequence Length as Features?}
In the embryonic stage of deep time series forecasting, RNN/LSTM based models \cite{LSTM_TKDE,deepar} dominate the research trend. RNN/LSTM models measure the connections among sequence elements by shared hidden layers between every two sequence elements, which are similar to traditional $AR(1)$ model: $x(t) = p(x(t-1))+\epsilon$. Ensuring the steadiness of input sequence length throughout the network is indispensable for such methods in that the target sequence is acquired gradually by rolling forecasting procedure. However, with the joint of CNN/Transformer based time series forecasting models\cite{scinet,TS2Vec,informer} and  one-forward forecasting procedure \cite{Multi_horizon,multi_step_ahead}, this prerequisite is no longer needed. Differences of these two procedures are shown in Appendix A. They extract latent features from the whole input sequence to seek features and then project them to the whole output sequence simultaneously. Temporarily, they show much more promising performances than RNN/LSTM based models. However, except few models, feature extractors of others are still limited by the old stereotype that the input sequence length shall be kept invariant throughout the whole network \cite{TCN,LSTNet,logtrans}. Even those which do shorten the input sequence length make some `compensations' for this behavior\cite{informer}.\par
In fact, as the above analysis of obsolete/popular deep time series forecasting models illustrates, it is time to abandon this old stereotype for it only brings redundant computation and risks of overfitting. Just like networks in other AI fields, e.g. Computer Vision \cite{ResNet,CSPNet}, the backbone of one-forward time series forecasting method shall also shrink the scale of input sequence length while enlarging the representation dimensions as the network gets deeper. Therefore, hierarchical time features could be captured with less time and space complexity. Moreover, input sequence elements are all known so there is also no risk of future information leakage. Therefore, Corollary \ref{corollary4} could be deduced. We will discuss more about this in the Section \ref{section3.1}/\ref{section3.2} as it is one of the core ideas of our proposed RTNet.\par
\begin{corollary}
	\label{corollary4}
	Within the feature extractor of auto-regressive time series forecasting network, it is feasible to shorten the dimension of input sequence length in latent space if employing one-forward forecasting procedure.\par
\end{corollary}
\subsubsection{Variance of Input Sequence Length as Hyper-parameter?}
The phrase, `long-term dependency', occurs more and more frequently in recent time series forecasting researches \cite{KSparse,FEDformer} owing to the powerful feature extractor, Transformer. Dramatically, such researches tacitly find out Conjecture \ref{conjecture1}.
\begin{conjecture}
	\label{conjecture1}
	Dealing with long sequence time series inputs, the self-attention scores of Transformer encoders fit sparse distribution.
\end{conjecture}
In fact, the sparseness of Transformer attention scores is a topic far not limited in time series forecasting \cite{sparseTKDE,Reformer, Longformer}. However, in line with the spirit of specific analysis of specific problems, we believe that the occurence of Conjecture \ref{conjecture1} in time series forecasting is not a coincidence and we are the first one trying to explain this to the best of our knowledge. According to H\ref{h4}. and concepts in traditional ARIMA model \cite{box1968,box2015}, any time series forecasting network is no more than an auto-regressive LTI system if not considering activation modules and has fixed parameters after training. So it is highly possible that time series forecasting network has similar characteristic of $AR(p)$, i.e. Theorem \ref{theorem1}.
\begin{theorem}
	\label{theorem1}
	The PACF of an $AR(p)$ model satisfies that $\phi_{kk}=0$ for $k>p$.
\end{theorem}
\par
Suppose that time series ${Y_t}$ is normally distributed and stationary. PACF refers to partial autocorrelation function and $\phi_{kk}$ is the correlation between ${Y_t}$ and ${Y_{t-k}}$ after removing the effect of the intervening variables $Y_{t-1},Y_{t-2},\ldots,Y_{t-k+1}$.\par
Theorem \ref{theorem1} shows that if the first predicted element $Y_{t+1}$ satisfies $AR(p)$, then predicted elements after it will be totally unrelated to ${Y_{t-k}}$ if $k>p$. Therefore, we could immediately come to Corollary \ref{corollary5} if any time series forecasting network is also fully auto-regressive.
\begin{corollary}
	\label{corollary5}
	Given stationary or piece-wise stationary time series, we could find a certain sequence length $P$ and a small fluctuation $\epsilon\ll P$. Then the forecasting performance of any auto-regressive network which sets input sequence length as $L_{in}$, where $\vert L_{in} - P\vert>\epsilon$, will be statistically worse than the same network setting input sequence length as $P$ due to over-fitting.\par
\end{corollary}
\par
Although based on H\ref{h4}, practical time series are rarely stationary, Corollary \ref{corollary5} is still tenable. We could reasonably assume that the time series is piece-wise stationary otherwise the time series is even not auto-regressive. So in each piece, the Corollary \ref{corollary5} is established so that there will be a series of $\{P_i\}_{i=1}^n$ where $n$ is the number of finite pieces so $P_i$ is also finite. Consequently, the required $P\in[\min{P_i}, \max{P_i}]$. We further discuss the rationality of Corollary \ref{corollary5} in Appendix C4. In fact, Corollary \ref{corollary5} could not directly calculate the best $P$. It only points out the invariance of the best input sequence length (region) with the prolonging of prediction length. However, it is still very meaningful for it proves that there is no need for tuning the input sequence length with the needs of different prediction lengths.\par
Let's turn back to the topic of sparse score distribution of Transformer. Based on Corollary \ref{corollary5}, it occurence is no more unusual. We could even give a more complete corollary:
\begin{corollary}
	\label{corollary6}
	Dealing with irrationally long sequence time series inputs, both the self-attention scores of forecasting Transformer encoders and the cross-attention scores of decoders fit sparse distribution.\par
\end{corollary}
\par
The establishment of Corollary \ref{corollary6} is very obvious. Suppose that $P(\pm \epsilon)$ is already the approximately most suitable input sequence length for a certain forecasting task, then any input sequence element is irrelevant to most of other elements based on Corollary \ref{corollary5} if given input sequence whose length is $L_{in} \gg P$. When it comes to Transformer, this fact will be converted to the phenomenon that the self-attention scores of encoders will fit long-tail distribution, i.e. Conjecture \ref{conjecture1}, in that elements are only related to $2P$ adjacent elements just as many other researches report \cite{informer,logtrans,Yformer}. Moreover, unknown elements in the prediction window are more likely to be relevant to the last $P$ elements of the input window and not related to earlier ones (The number of related elements here is $P$ rather than $2P$ owing to P\ref{p2}). Thus, cross-attention scores of decoders also fit long-tail distribution. Finally, Corollary \ref{corollary7} can be deduced.\par
\begin{corollary}
	\label{corollary7}
	Given stationary or piece-wise stationary time series, it is a better idea for forecasting Transformer to receive shorter but appropriate input sequences and employ Canonical Attention Mechanism instead of processing unnecessary long input sequences and using diverse `Sparse Attention Mechanism'.\par
\end{corollary}
\section{RTNet}
We fully integrate solutions/conclusions given out in Section \ref{section2} into our design of RTNet. In the surplus section, three core ideas of RTNet will be introduced, followed with the presentation of RTNet with end-to-end and contrastive learning based forecasting formats.\par
\subsection{Application of Respecting Time Series Properties}
\label{section3.1}
\begin{figure}[]
	\centering
	\subfloat[]{\includegraphics[width=1.64in]{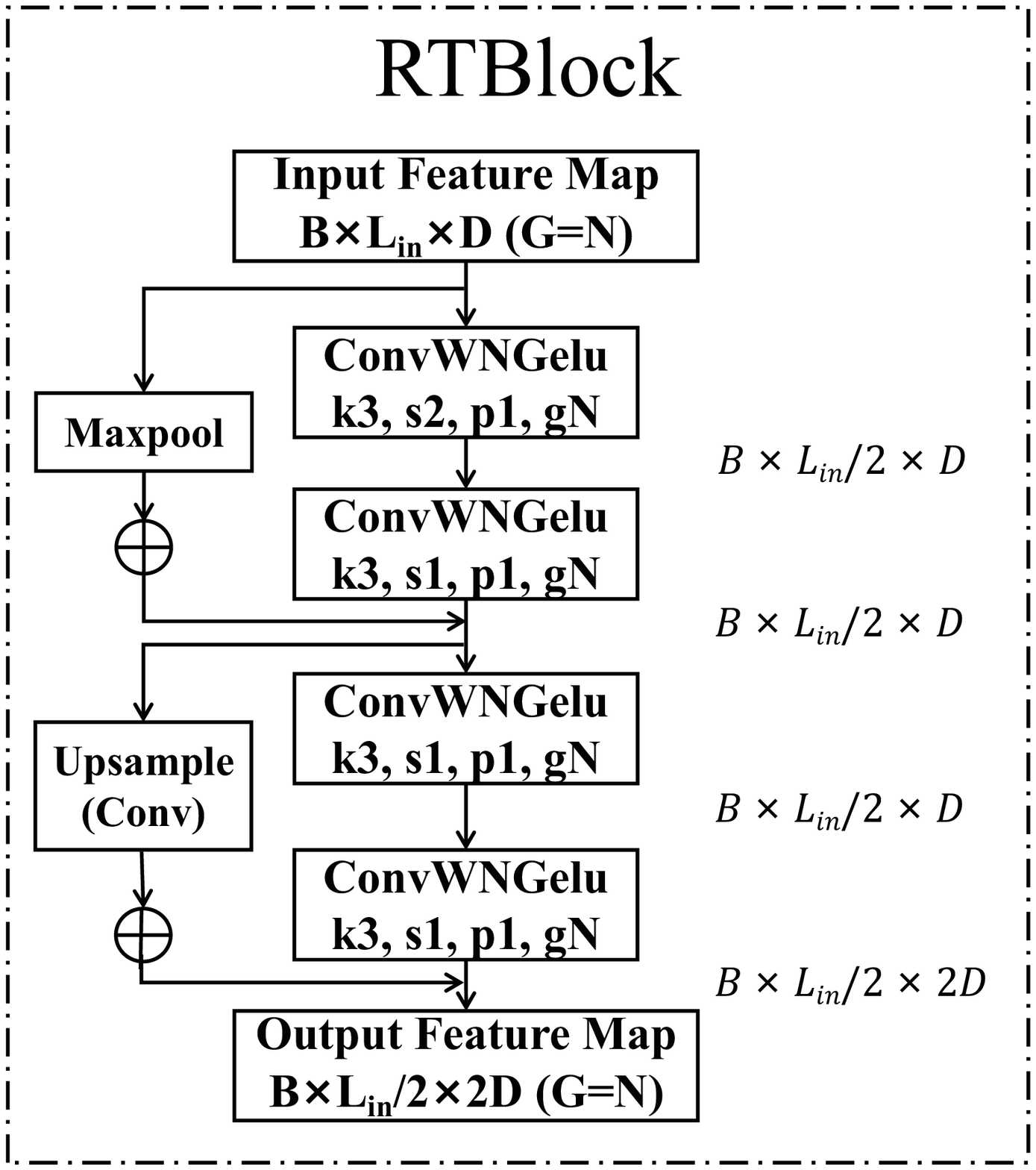}\label{fig3a}}
	\subfloat[]{\includegraphics[width=1.29in]{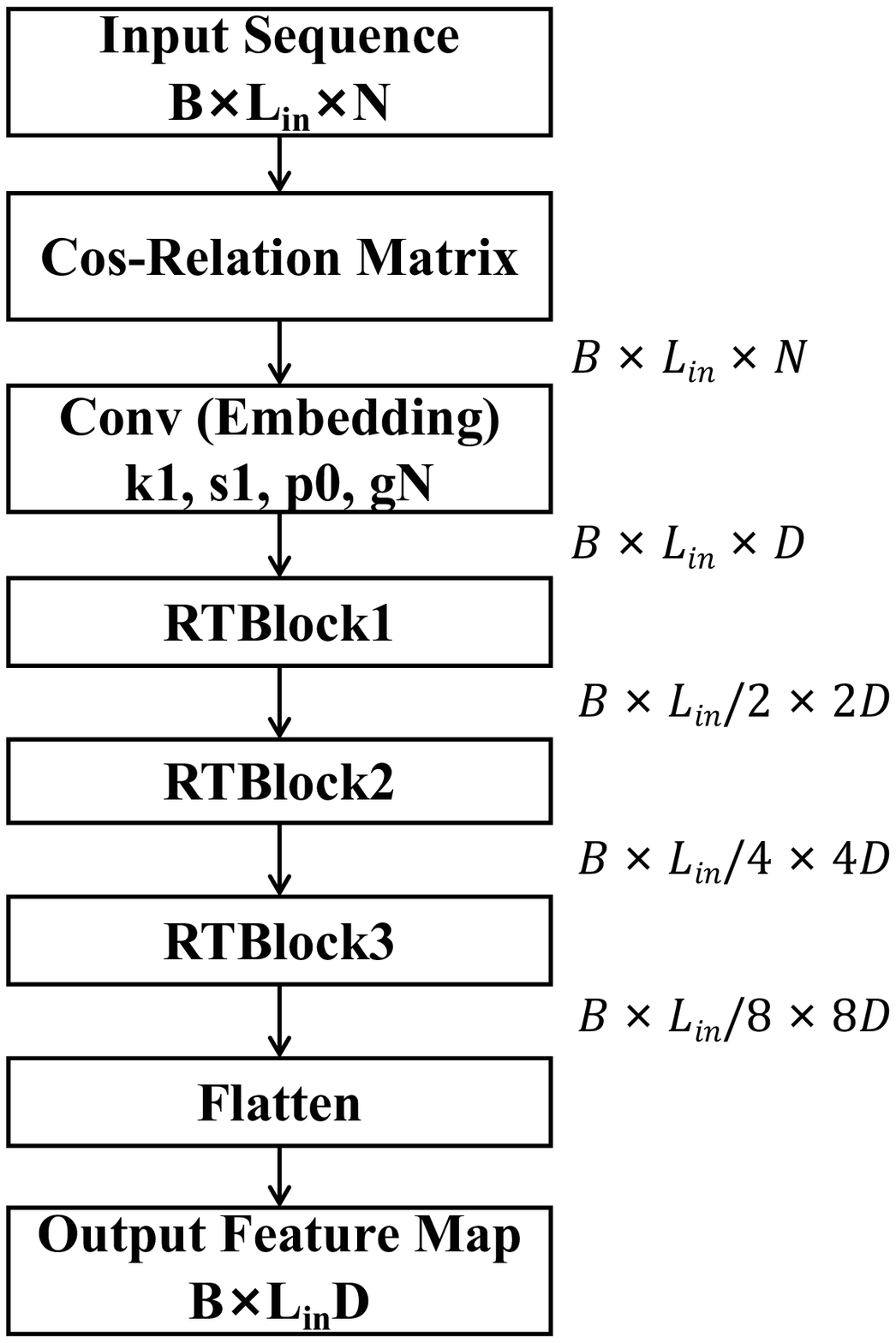}\label{fig3b}}	
	\caption{An example of RTNet backbone is shown in (b). It is mainly composed of three RTBlocks shown in (a). In each RTblock, a maxpooling layer and an upsampling layer are used to ensure identical dimensions of Res-connected tensors. The meanings of symbols are as follows and the same in following figures: $B$: batch size; $L_{in}$: input sequence length; $N$: number of variates; $G$: number of groups ; $D$: hidden units; $k$: kernel size; $s$: stride; $p$: padding and $g$: group.}
	\label{fig3}
\end{figure}

The first core idea is definitely respecting time series properties due to the theme of this paper. So the backbone, i.e. input sequence feature extractor, of RTNet presented in Fig.\ref{fig3} completely obeys corollaries drawn in Section \ref{section2}. Imitating the architecture of ResNet-18 \cite{ResNet} which is typical in CV field, RTNet is its transformed `One Dimension' version. Its dominant component, RTblock, is also transformed from the blocks used in ResNet-18. RTNet only uses an additional convolutional layer as the embedding layer. Notice that convolutional layers used in RTNet are all group convolutional layers \cite{groupConv}. Although group convolution is invented for the reduction of memory usage, it has more meaningful value here in that it ensures that the weight allocation of cos-relation matrix maintains throughout the whole network (Corollary \ref{corollary3}). The group number is supposed to be same with the number of variates.\par
Matching aforementioned three points in Section \ref{section2}, in RTNet we employ WN as the normalization method (Corollary \ref{corollary1}), apply cos-relation matrix at the beginning of network (Corollary \ref{corollary3}) and gradually shorten the second dimension of hidden units, i.e. sequence length (Corollary \ref{corollary4}). Fig.\ref{fig3b} only shows RTNet composed of three RTBlocks, more blocks could be added according to the length of input sequence. With more RTBlocks in RTNet, the sequence lengths of hidden units are gradually halved while the channels of hidden units are doubled. Therefore, the total output feature dimension is invariant if the embedding dimension is determined. It means that, not limited by the conventional element-wise extracting method, RTNet is capable of capturing more universal features without owning more parameters in final FC projection layer (Fig.\ref{fig6}) which occupies a large portion of network parameters.\par
\subsection{Causal Pyramid Network}
\label{section3.2}
In line with H\ref{h4}, time series are normally not stationary so that its statistics vary with the length of input sequences. Hence, although RTNet in its current modality is able to capture universal features of the whole input sequence, it lacks the ability to capture multi-scale hierarchical feature maps. At this moment, pyramid networks are naturally associated with this demand. Nevertheless, common pyramid networks/ideas popular in other AI fields lose efficacy here \cite{pyramid,pyramidCV2,pyramidCV3}. In fact, we have tried the RTNet combination with classical structures such like FPN \cite{FPN}, PAN \cite{PAN} in CV fields. The forecasting accuracy is out of satisfactory and even drops. We believe that it is also the time series property that causes this happening. Taking objection detection task in CV as an instance, we are unable to know which fixed local area is more useful in every frame \cite{yolo,FPN}. So we shall take every possible local reception field into account then architectures like FPN will work. However, P\ref{p2} and Corollary \ref{corollary5} illustrate that input sequence elements closer to the prediction window are statistically more related to it. It is equivalent to the fact that we have already gotten priori knowledge where we should pay more attention to in any input sequence. Consequently, we propose CPN (\textbf{\textit{C}}ausal \textbf{\textit{P}}yramid \textbf{\textit{N}}etwork) which is a more suitable pyramid network for time series forecasting.\par
\begin{figure}[!t]
	\centering
	\includegraphics[width=3in]{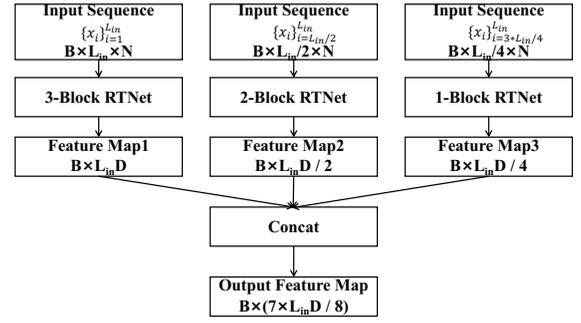}
	\caption{An example of CPN. The input element $x_i\in {R}^N$. Three feature maps outputted by three independent networks are concatenated into the final feature map. The sub-network containing most RTBlocks (the leftmost one in the figure) is defined as the dominant pyramid network/feature extractor.}
	\label{fig4}
\end{figure}

An example of CPN is shown in Fig.\ref{fig4}. In the light of the number of RTBlocks $ N $ owned by the dominant feature extractor($ N=3 $ in Fig.\ref{fig3}/\ref{fig4}), identical number of feature extractors are employed and mutually independent to maximize the number of pyramid networks. The $ i ${th} feature extractor receives the last $ 1/2^{i - 1} $ of input sequence and contains $ (N - i + 1) $ RTBlocks. The final output features are composed of all features provided by every feature extractor. By employing CPN, RTNet will pay more attention to input sequence elements which are more closed and expectedly more relevant to the prediction sequence. Therefore, later input sequence elements will be extracted more hierarchical features belonging to multi-scale representations, which is the origin of name `Causal Pyramid Network'.\par
\subsection{Decoupled Time Embedding}
\label{section3.3}
Without doubt, time series properties contain time properties. Time embedding is thereby commonly used in large amounts of time series forecasting networks \cite{logtrans,informer,KSparse,Yformer}. Most of networks treat it the same way like position embedding of Transformer \cite{attention2017} making it effective in limited occasions. They simply mix them with normal variates of input sequences as network inputs. There is no harm to treat time signals in another way. No matter which time embedding, for example, hour-of-day, is nothing but a new `variate', a objectively existing variate that we could totally know even in the future. These `time variates' distinguish themselves from normal variates for they are not causal and not auto-regressive. Conversely, they are determined and objective at a determined time stamp. Therefore, if we expect to efficiently utilize time embedding, we shall at least employ them in prediction window rather than input window. We will discuss it under two occasions.\par
\textbf{A. If time embedding is truly relative to normal variates.} This occasion is equivalent to saying that periodic term could be decomposed from signals of normal variates and its signal period is related to time embedding. If so, certainly it is more convenient to predict the decomposed periodic term of the prediction window directly through the time embedding at the same time stamp. Otherwise, the network will first find relations among normal variates and time embedding in the input window and then project them into the outputs in the prediction window, causing more computation and parameter costs.\par
\textbf{B. If time embedding is not relative to normal variates.} This occasion also contains parts of occasions A where time embedding is slightly relative or the relative time information is hard to be described by commonly used time embedings, e.g. hour-of-day. In this occasion, employing time embedding may cause over-fitting or harm the prediction of normal variates. So making the feature extraction of input sequence and time embedding mutually independent will minimize the loss. Obviously, it can be easily achieved by only using the time embedding in prediction window.\par
\begin{figure}[!t]
	\centering
	\subfloat[]{\includegraphics[width=1.58in]{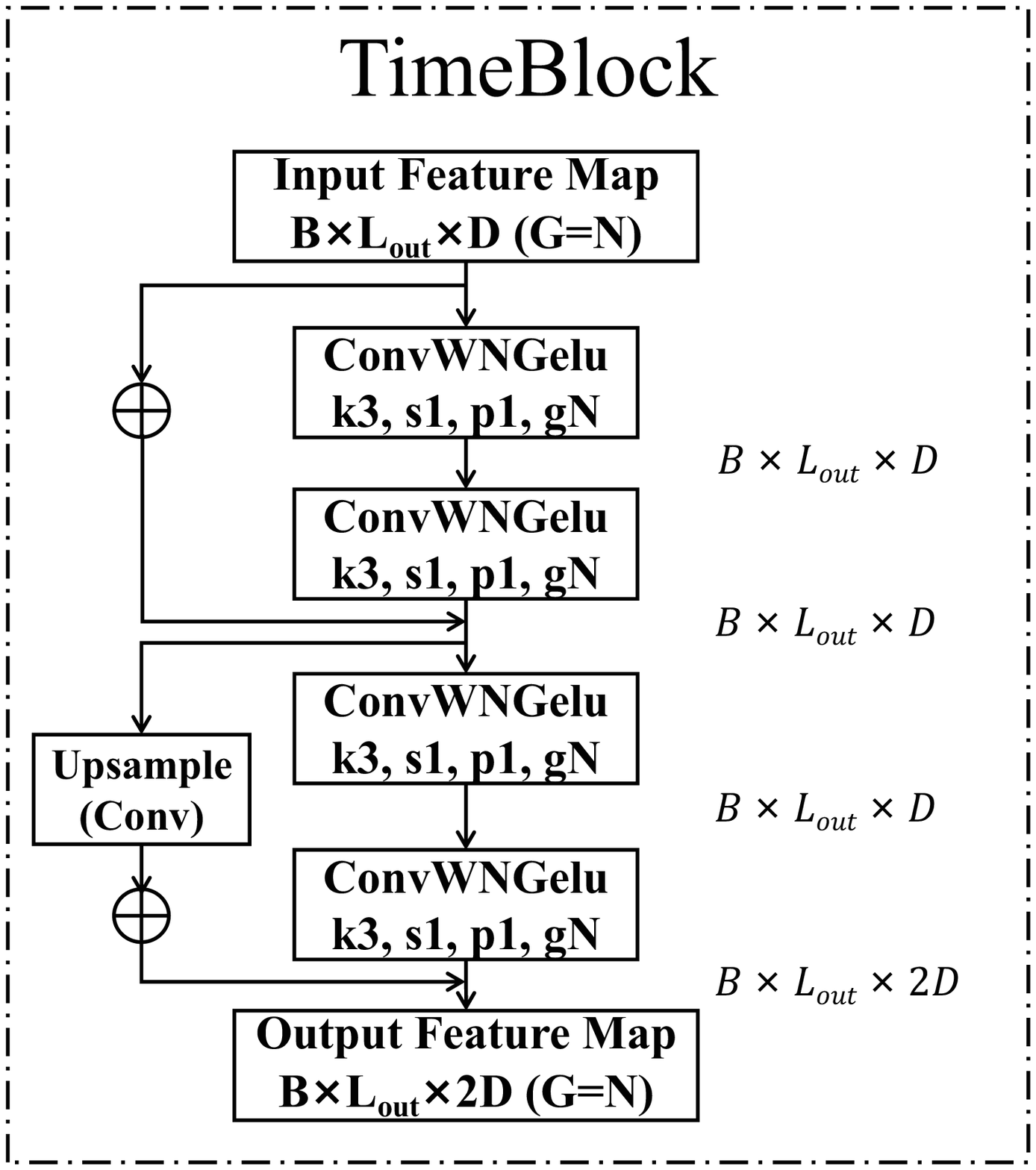}\label{fig5a}}
	\subfloat[]{\includegraphics[width=1.63in]{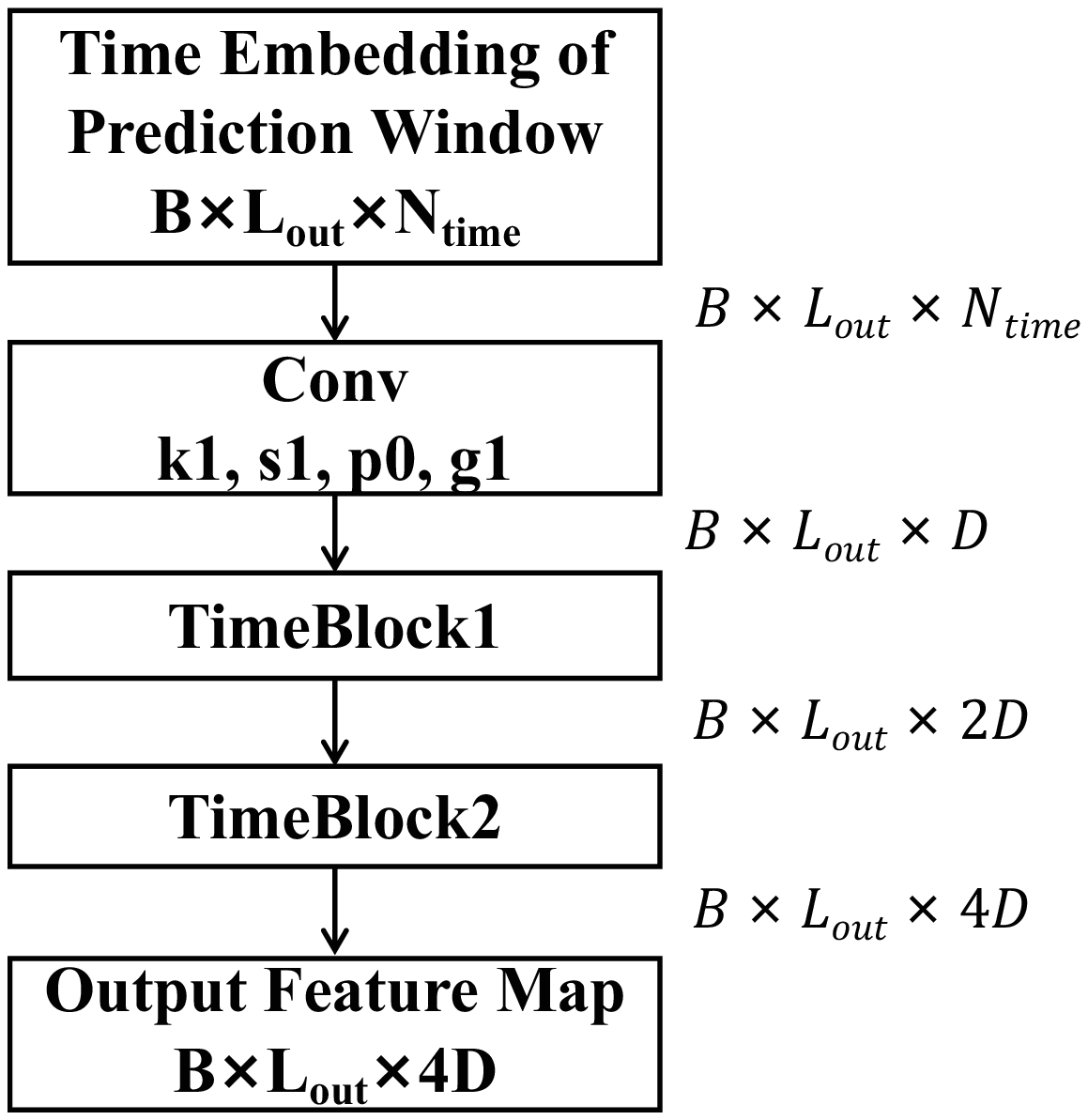}\label{fig5b}}	
	\caption{The architecture of time embedding network in (b) is mainly composed by two TimeBlocks shown in (a). $ N_{time} $ is the number of time embedding and $ L_{out} $ is the length of prediction sequence.}
	\label{fig5}
\end{figure}

Based on above analysis, we deduce that the feature extraction of time embedding should be independent from that of input sequence and propose the method of decoupled time embedding. RTNet use an additional tiny time embedding network show as Fig. \ref{fig5}. It is similar to the backbone architecture in Fig.\ref{fig3} but the stride of all convolutions is 1 for there is no need to extract global features from time-invariant periodic signals. Group convolution is also used to match the process of backbone where group number is the number of variates rather than the number of time embedding. We denote by TimeNet the decoupled time embedding network for following usage.\par
\subsection{End-to-end Forecasting Format}
\label{section3.4}
\begin{figure*}[]
	\centering
	\includegraphics[width=4.5in]{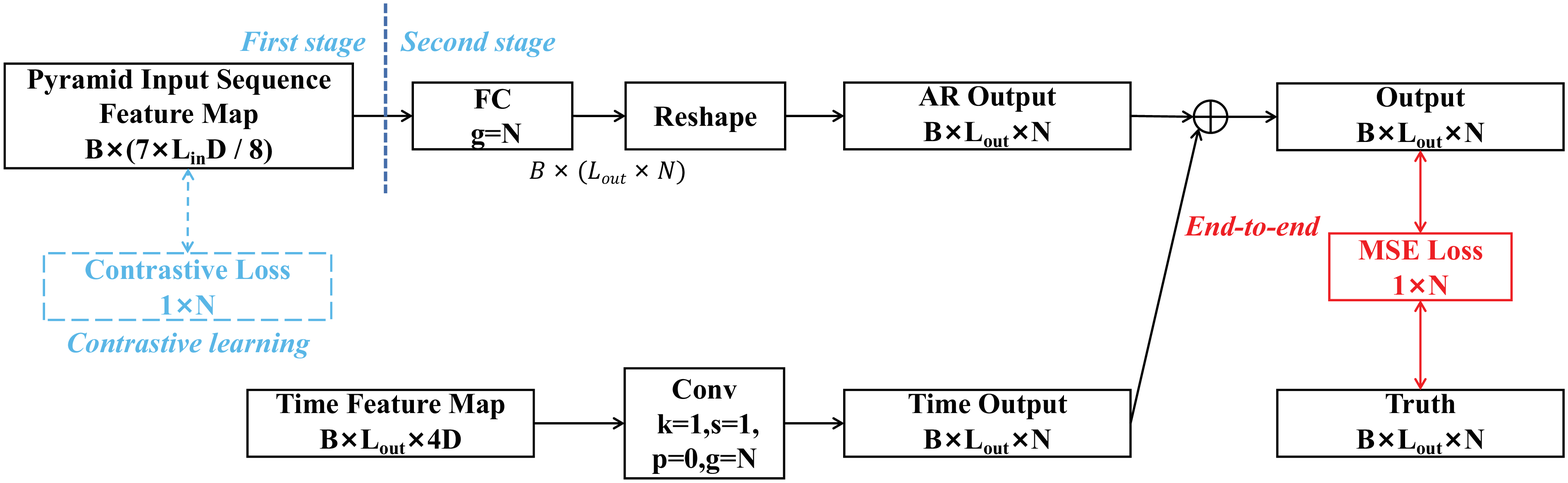}
	\caption{Overall architecture of RTNet for end-to-end and contrastive learning based format. The input sequence output (AR output) and time output are added together to calculate MSE loss with the truth (red block) in end-to-end format and the second stage of contrastive learning based format. At the first stage of contrastive learning based format, contrastive loss (blue block) is calculated by pyramid input sequence feature map.}
	\label{fig6}
\end{figure*}
Above three subsections \ref{section3.1},\ref{section3.2},\ref{section3.3} have already shown the feature extraction process of RTNet, in this subsection \ref{section3.4} and the subsection below \ref{section3.5}, we will respectively present two ways to use extracted feature map for forecasting. This subsection shows end-to-end forecasting format.\par
An overview of the whole network is shown in Fig.\ref{fig6}. The input sequence feature map, i.e. AR(\textbf{\textit{A}}uto-\textbf{\textit{R}}egressive) part which is extracted by RTNet backbone in Fig. \ref{fig4}, is projected through a group fully-connected layer and reshaped to the size of prediction window. Meanwhile, time feature map is extracted by decoupled time embedding network in Fig. \ref{fig5} and projected through a one-kernel-size group convolutional layer. Both results are added together to obtain the final output. MSE loss function is used for network training. To cooperate with cos-relation matrix, a loss vector, instead of a simple number, is back propagated. Average square errors are calculated individually with each variate to ensure their independence throughout the whole training process. If there are $N$ variates to predict, then the length of loss vector will be $N$. See Appendix D for pseudo code.\par
\subsection{Contrastive Learning Based Forecasting Format}
\label{section3.5}
We choose contrastive learning as the way to learn representations of input sequence feature map while combining RTNet with self-supervised forecasting format. Existing time series forecasting baselines with contrastive learning mainly build element-wise positive/negative samples \cite{TS2Vec,Cost}. Obviously, representations they obtain will not be more universal than our `sequence-wise' methods. RTNet now maintains its form of end-to-end forecasting (Fig.\ref{fig6}), however is trained in two stages. The first stage learns the representations of input sequences through contrastive learning while the second stage does the prediction work.\par
In the first stage, similar to SimCLR\cite{simclr}, we randomly sample a minibatch of $ B $ input sequences and define the contrastive representation task on sets of augmented instances originated from the minibatch, resulting in $ (1 + I)B $ data points. $ I $ is the number of augmented instances from each input sequence where the data augmentation method is randomly chosen among `Scaling'\footnote{Scaling : $ f(x_i)=x_i\cdot (1+\beta'),\ \beta'\sim U(-\beta,\beta)$}, `Jittering'\footnote{Jittering: $ f(x_i)=x_i+\beta', \ \  \beta' \sim U(-\beta,\beta) $} and `Entirety Scaling'\footnote{Entirety Scaling: $ f(x)=x\cdot(1+\beta'), \ \  \beta' \sim U(-\beta,\beta) $}. Moreover, input sequences in the same minibatch need to meet Condition \ref{condition} to avoid occasions when time stamps of some input sequences are mostly coincided.\par
\begin{condition}
	\label{condition}
	Whatever $ {s_i,s_j} $ of length $ L_{in} $ belonging to the same minibatch, interception $(s_i, s_j) \leq L_{in}-\frac{L_{in}}{\alpha}, \alpha\geq1 $ is a manually selected hyper-parameter.
\end{condition}
\begin{figure}[]
	\centering
	\includegraphics[width=2in]{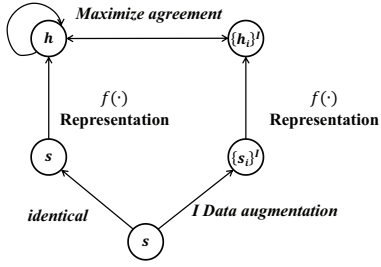}
	\caption{The framework for input sequence representation with contrastive learning. As for any initial sequence $ s $, $ \{s_i\} $ will be its augmented set where $ i \in [1, I] $. The CPN $ f(.) $ is trained to maximize agreements of the representation of $ s $ with itself and those of $\{s_i\}$.}
	\label{fig7}
\end{figure}
\begin{figure}[]
	\centering
	\includegraphics[width=2.1in]{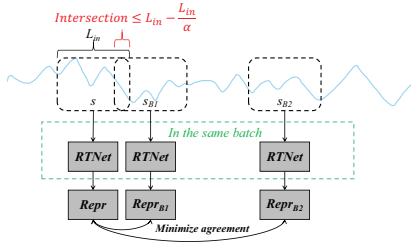}
	\caption{Selections of negative instances. Overlaps among given initial instance $s$ and its negative instances $\{s_{Bi}\}$ are not larger than $ L_{in}{-}\frac{L_{in}}{\alpha}$. }
	\label{fig8}
\end{figure}

As shown in Fig.\ref{fig7}, Positive pairs of each input sequence $ s $ include the pair $(s, s)$ and the pairs $(s,{s_i}) $. ${s_i}$ are augmented instances. Given a certain input sequence and its positive instances, we treat the rest $ (1 + I)(B - 1)$  instances as negative instances as Fig.\ref{fig8}. Cosine similarity again is used here to measure the agreement within each pair. Then the loss function for sequence $s_m(m\in [1,B])$ is defined as Eq.\ref{eq12}.\par
In Eq.\ref{eq12},$h_{j}$ denotes the representation of $j ${th} input sequence and $\{h_{ji}\}^I$ is the set of augmented instances' representations. Different from the canonical InfoNCE, we take the self-similarity into account in that we find out that only maximizing agreements among each input sequence and its augmented sets are not enough to distinguish the representation of it with those of other sequences in latent space during experiments. We believe it is the non-stationarity of time series that leads to this phenomenon. The value `$e$' in numerater is in fact $sim(h_m,h_m)$, i.e. self-similarity.\par
\begin{align}
	\label{eq12}
	&sim(u,v)= \exp(\dfrac{\vert u^\top \cdot v\vert}{\Vert u\Vert\cdot\Vert v\Vert}) \nonumber\\
	Loss_m = -\log& \dfrac{e+(\sum\limits_{i=1}^I sim(h_m,h_{mi}))}{\sum\limits_{j=1}^B(sim(h_m,h_j)+(\sum\limits_{i=1}^I sim(h_m,h_{ji})))}
\end{align}

Note that Eq.\ref{eq12} only gives the loss function of the  $m ${th} sequence of a certain variate. If there are $N$ variates, then the whole loss function will be also a loss vector taking the length of $N$ while the loss function of each variate is the average of corresponding $\{Loss_m\}^N$. See Appendix D for preudo code. The feature extraction of time embedding does not participate in the first stage as periodic signals do not own enough complicated latent features worth doing it.\par
Once the first stage finishes training, network parameters before `FC' in Fig.\ref{fig6} (including backbone presented in Fig.\ref{fig3}/\ref{fig4}) are determined and steady. During the second stage, the rest of RTNet is trained to predict the final output sequence which is alike process in subsection \ref{section3.4}. The only difference is that there will not exist any back propagation in the sub-network which has already finished training during the first stage.  Unlike other baselines \cite{TS2Vec,Cost} which use regressors such as Ridge regressor, SVM, etc. in the second stage, we use a group convolutional layer to obtain the prediction sequence for two reasons. Firstly, the whole architecture of RTNet will be maintained in both two forecasting formats. Moreover, using group convolutional layer will be more convenient to keep the independence of projection processes of different variates. Similar to the final process of RTNet in end-to-end forecasting format, the projection results of representations outputed by the first stage and additional projection results of the feature map of time embedding are added together in the second stage for final prediction results. MSE loss function is utilized.\par
\section{Experiment}
\label{section4}
\subsection{Dataset Introduction and Analysis}
\label{section4.1}
Above all, it is necessary to introduce and analyze properties of employed datasets for respecting time series properties. Unfortunately, most of deep time series forecasting papers just introduce them without analyzing differences of their properties. So we would like to fulfill this blank here.\par
We choose ETT, WTH and ECL\footnote{ These three datasets were available at: \url{ https://drive.google.com/drive/folders/1ohGYWWohJlOlb2gsGTeEq3Wii2egnEPR?usp=sharing}} datasets for their different properties according to H\ref{h2}-H\ref{h4}. We present global/local distributions of three datasets target variates in Fig.\ref{fig9}/\ref{fig10} for analysis. Their introductions are shown in Appendix E1.\par
\textbf{Properties of different datasets:}
 \par
\begin{figure*}[]
	\centering
	\subfloat[]{\includegraphics[height=1.3in]{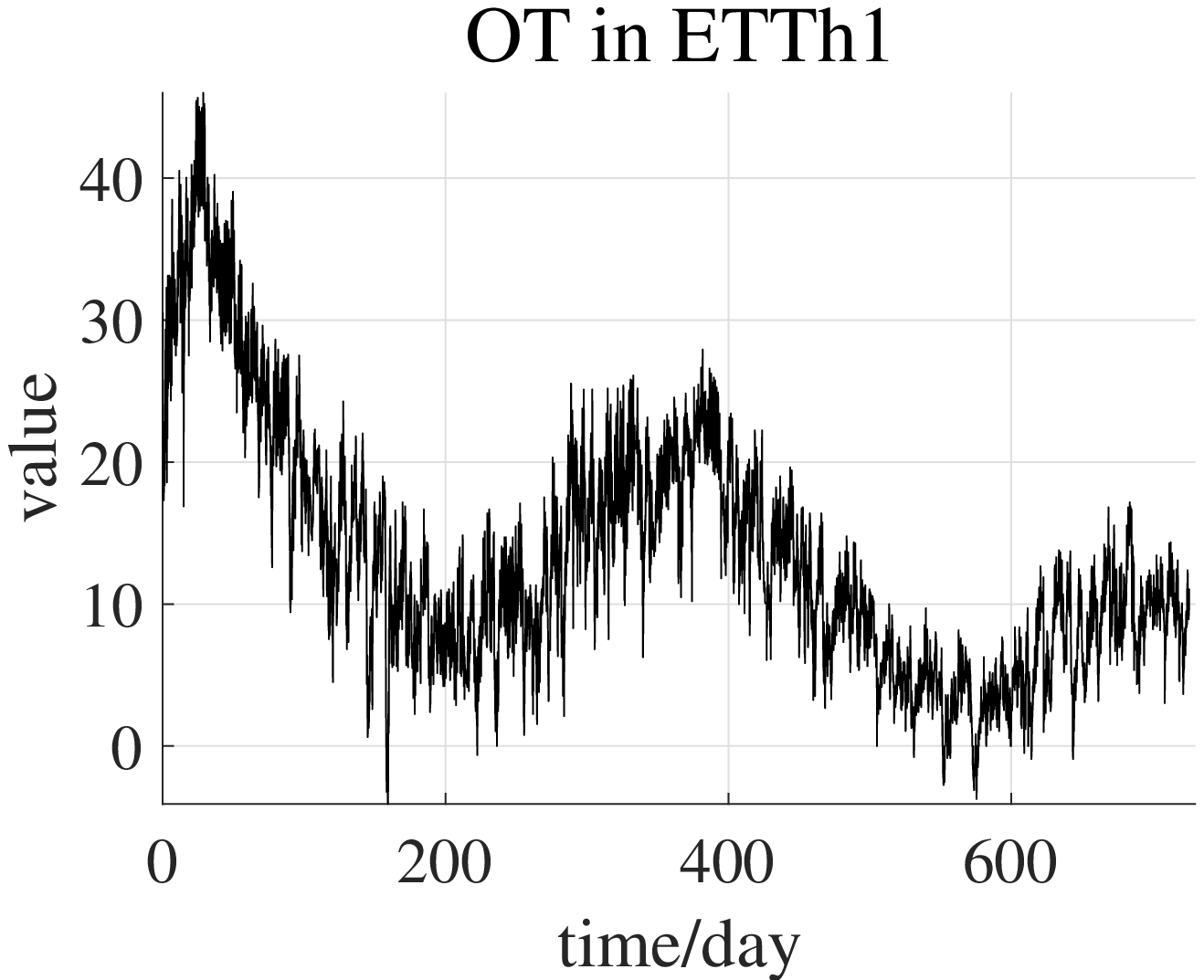}\label{fig9a}}
	\subfloat[]{\includegraphics[height=1.3in]{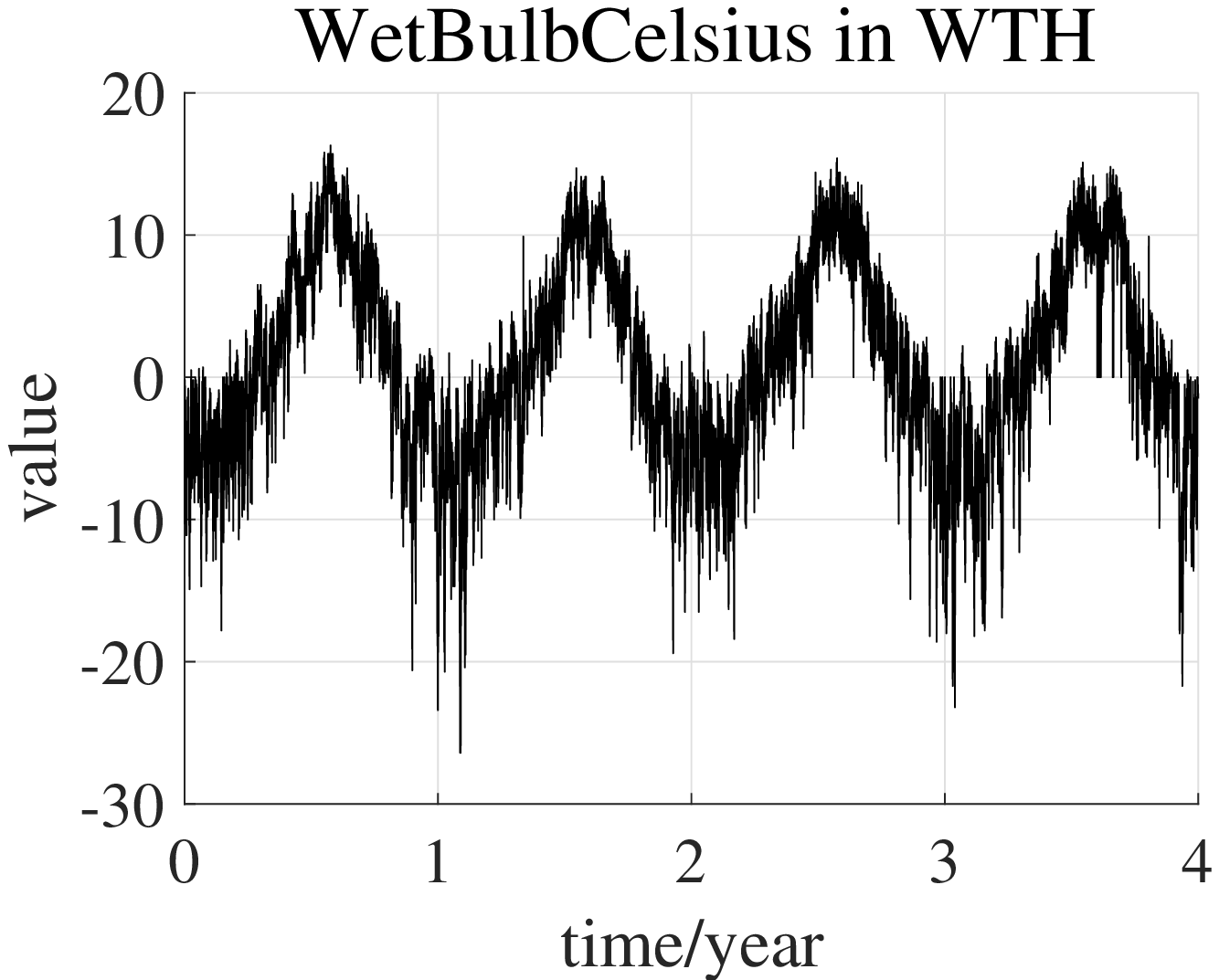}\label{fig9b}}	
	\subfloat[]{\includegraphics[height=1.3in]{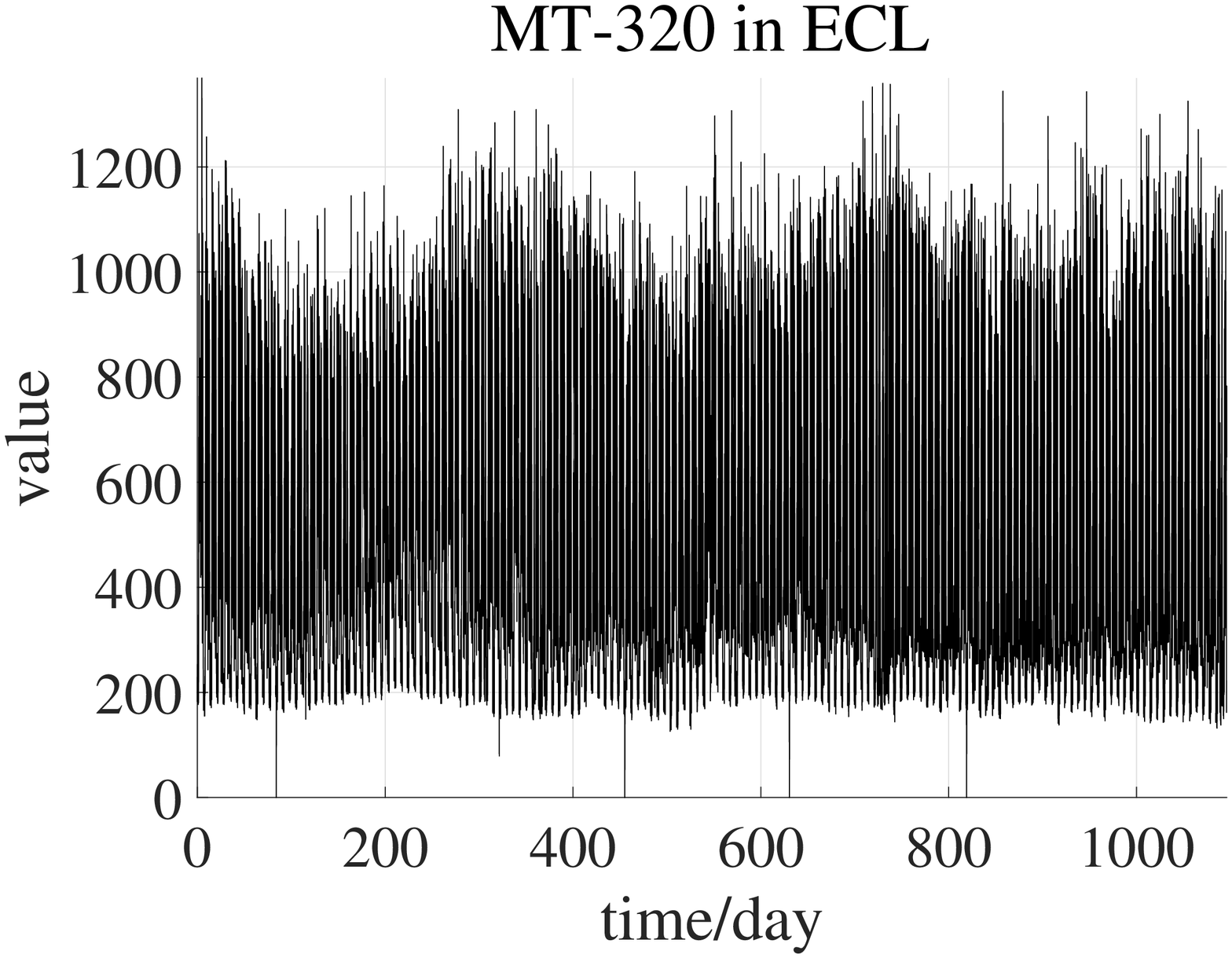}\label{fig9c}}	
	\caption{Global windows of target variables in ETTh$ _{1} $(a)/WTH(b)/ECL(c).}
	\label{fig9}
\end{figure*}
\begin{figure*}[]
	\centering
	\subfloat[]{\includegraphics[height=1.3in]{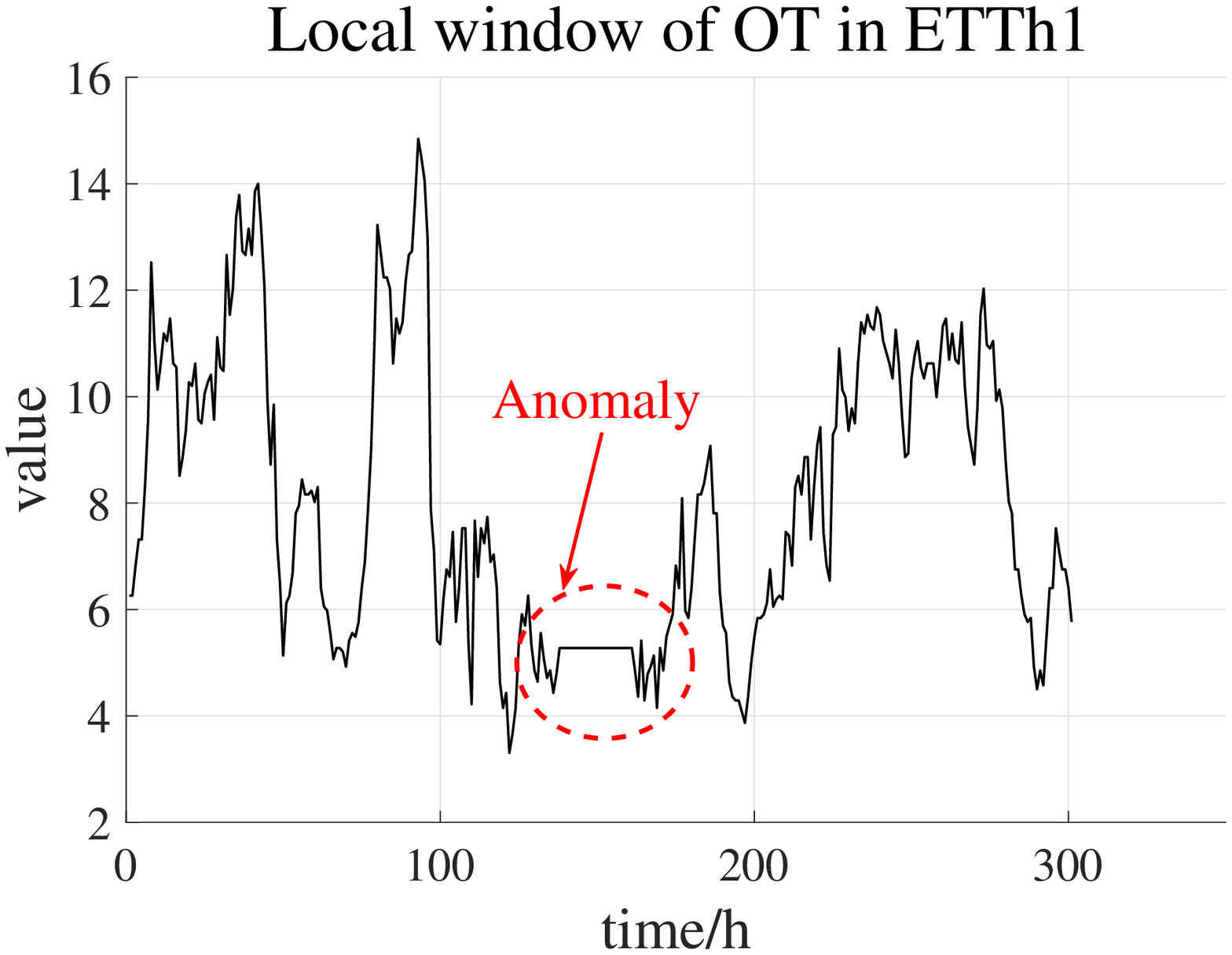}\label{fig10a}}
	\subfloat[]{\includegraphics[height=1.3in]{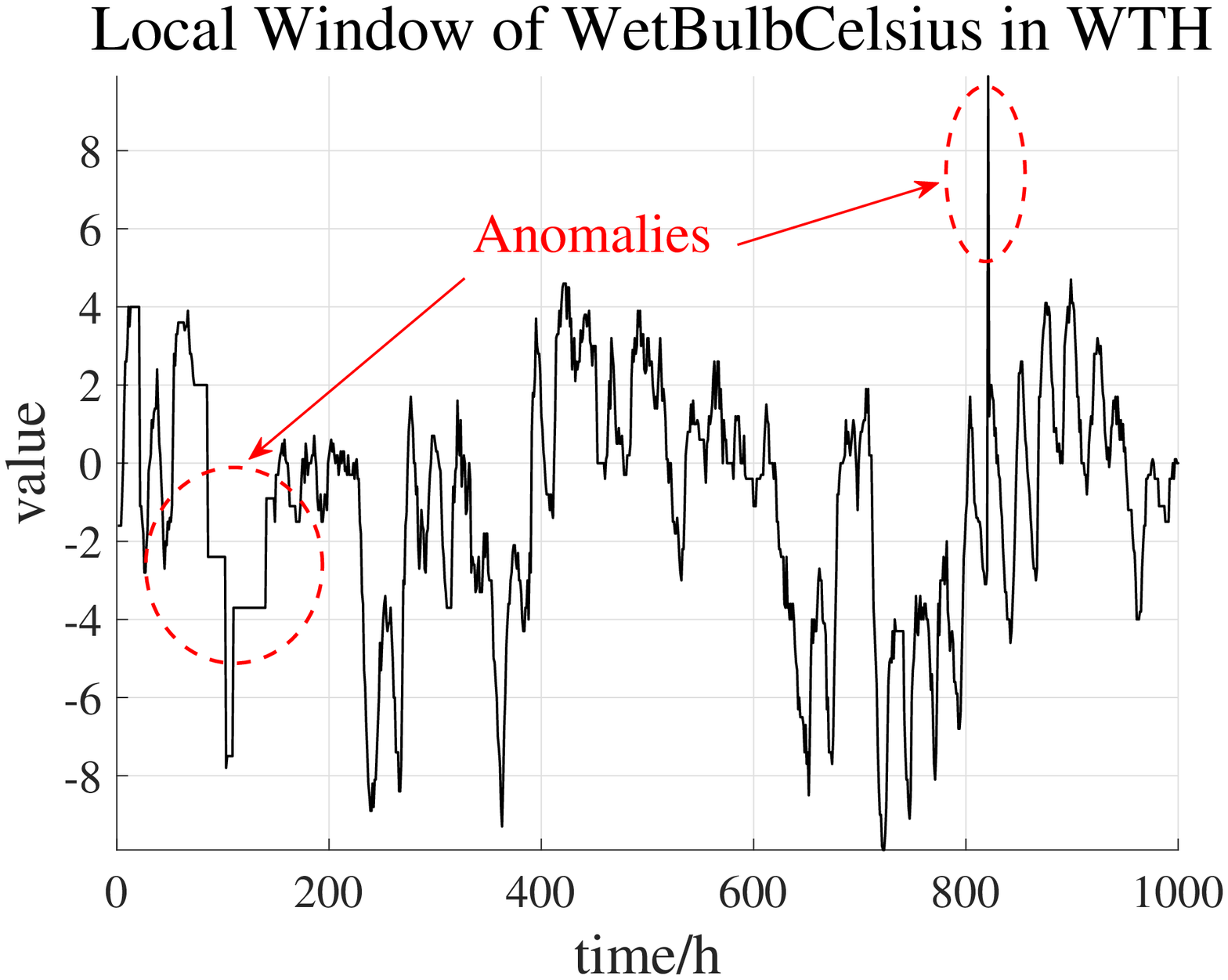}\label{fig10b}}	
	\subfloat[]{\includegraphics[height=1.31in]{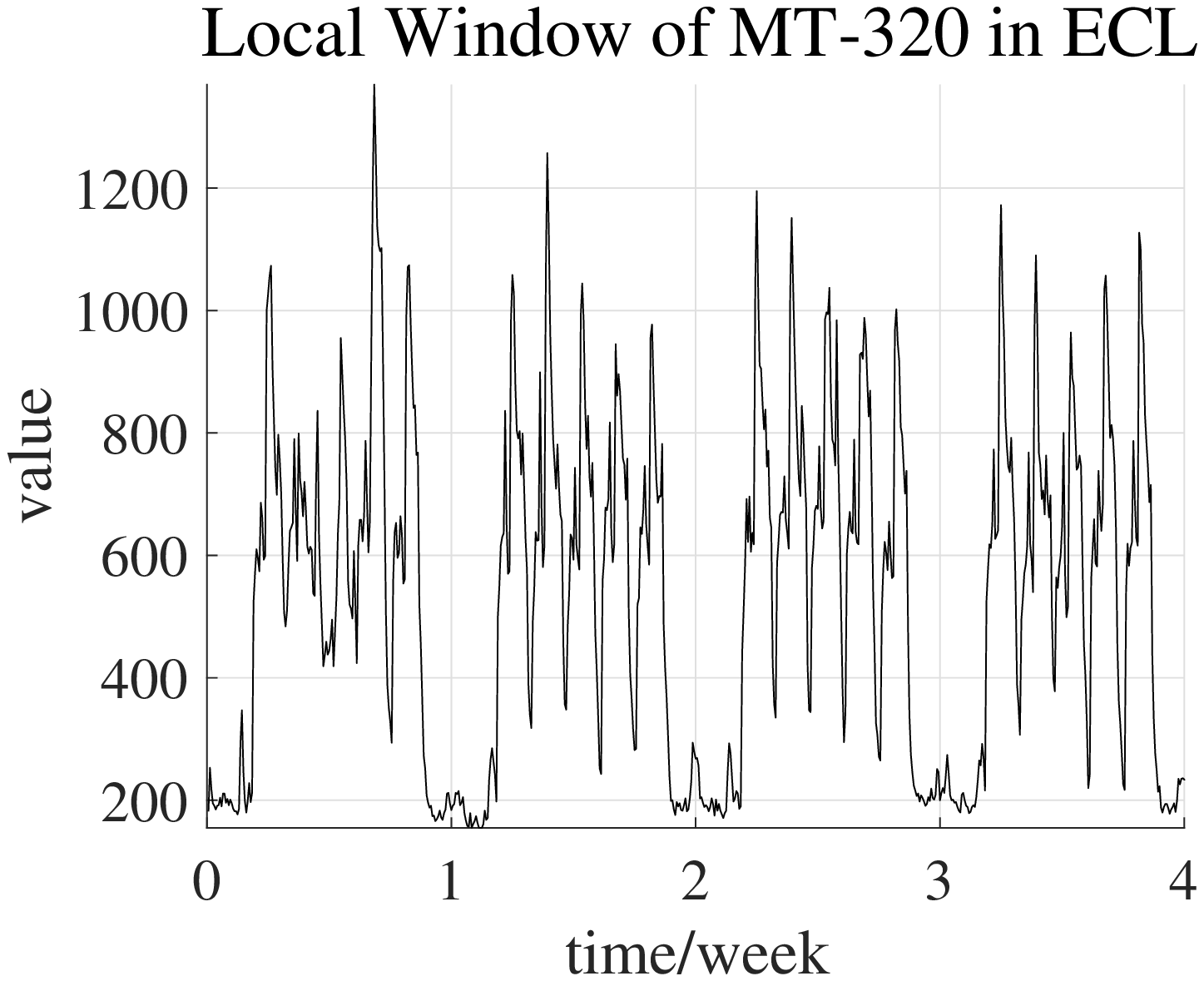}\label{fig10c}}	
	\caption{Local windows of target variables in ETTh$ _{1} $(a)/WTH(b)/ECL(c). Anomalies in these windows are circled in red.}
	\label{fig10}
\end{figure*}
(1) Time-related property:\par
We define that a dataset is time-related only when it contains periodic terms whose periods belong to commonly used time units, e.g. 1 hour, 1 day=24 hours, 1 month=720 hours. We make this definition in that deep time series forecasting networks could choose to employ time embedding in these occasions. Then we analyze whether the target variates of three datasets are time-related.\par
In ETT, we take sub-dataset ETTh$ _{1} $ as an example. Fig. \ref{fig9a}/\ref{fig10a} shows that the target value of `OT' fluctuates heavily either in general or in local. So it is highly possible that within the data distribution of `OT', there does not exist periodic terms. As a result, it is expected that models dealing with `OT' do not need time embedding parts.\par
In WTH, things get different. It is obvious that in Fig. \ref{fig9b} the target value `WetBulbCelsius' has a 1-year-period periodic term. However, there still exists non-negligible turbulence from the local point of view in Fig. \ref{fig10b}. Objectively speaking, as the length of input windows cannot be too long, it is hard for auto-regressive models to learn the regularity of 1-year period. Therefore, models dealing with `WetBulbCelsius' require time embedding parts.\par
Situations in ECL vary again. Fig. \ref{fig9c}/\ref{fig10c} illustrates that the target value `MT\_320' takes 1 week as one of the periods. It is stable both globally and locally. The stability of the data distribution and the short period determine that adding time embedding parts or not to models when dealing with `MT\_320' will not affect forecasting results distinctly.\par
(2) Stationarity and anomalies:\par 
Similar with former analysis, from Fig. \ref{fig9a}/\ref{fig10a} it can be known that the data distribution of `OT' is not stationary globally and there exist anomalies in local input windows. Moreover, it can be deduced from Fig. \ref{fig9b} that the data distribution of `WetBulbCelsius' is stationary globally. Nevertheless, the statistical characteristics of `WetBulbCelsius' change in different local windows like Fig. \ref{fig10b}. Meanwhile, anomalies also exist in the data distribution of `WetBulbCelsius'. Different from others, the data distribution of `MT\_320' is stationary both globally/locally and it almost has no anomaly as Fig.\ref{fig9c}/\ref{fig10c} shows.\par
\begin{figure}[]
	\centering
	\includegraphics[width=2in]{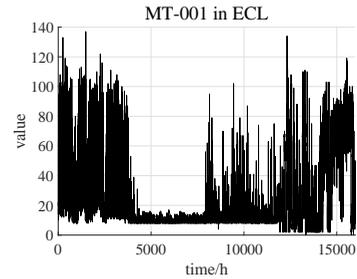}
	\caption{An example of non-autoregressive variate. It can be observed that data distributions of `MT\_001' before/after 5000 h are totally different. Meanwhile, anomalies seem even more frequent than normal data points. These phenomena show that `MT\_001' is non-autoregressive.}
	\label{fig11}
\end{figure}
(3) Multivariate relationship:\par
Based on priori knowledge, variates in WTH\cite{atmospheric} are closely relevant to each other and variates in ECL\cite{ECL} are in fact instances so they are independent to each other. However, relations of variates in ETT \cite{informer} are unknown.\par
(4) Auto-regression analysis:\par
Most of variates in these three dataset are auto-regressive like target values, however, there also exist some non-autoregressive variates especially in ECL, e.g. `MT\_001' shown in Fig.\ref{fig11}. These non-autoregressive variates are impossible to be precisely predicted through auto-regressive models owing to almost irrelevant data distributions of train/val/test and great amount of anomalies. Therefore, there is no need to perform multivariate forecasting of ECL.\par
From the above analysis, we can conclude that three datasets we employed have totally different properties in diverse views. Therefore, they are worth performing experiments on and different strategies shall be utilized to do forecasting with them to achieve perfect performances, which could fully highlight the theme of this paper, i.e. the significance of respecting time series properties.\par
\subsection{Experiment Settings and Hyper-parameters}
All settings and hyper-parameters are thoughtfully selected according to time series properties and corresponding analysis mentioned in Section \ref{section4.1}.\par
We select the prediction length identical to other baselines \cite{informer,scinet,TS2Vec,Cost}. It is chosen within \{24, 48, 96, 288, 672\} for ETTm$ _{1} $ and within \{24, 48, 168, 336, 720\} for other datasets. WN is chosen as normalization method abide by analysis in Section \ref{section2.1}. As variates in WTH are closely relevant to each other according to Section \ref{section4.1}, cos-relation matrix and groups convolutions are not applied to multivariate forecasting of WTH. When performing multivariate forecasting with ETT, the threshold $ \theta $ of cos-relation matrix is set to $ 45\degree/30\degree $. We select proper input sequence lengths for different datasets through numerous experiments on the basis of analysis in Section \ref{section2.3}. In following experiments, input sequence length is set to 168/336/48/336 for time series forecasting with ETTh$_{\{1,2\}}$/ETTm$ _1 $/WTH/ECL. \par
The number of RTBlock is chosen within \{3, 4\} for different datasets and the number of pyramid networks is identical. The number of TimeBlock is 2 as Fig. \ref{fig5} shows when forecasting with WTH and time embedding is not employed when forecasting with ETT/ECL in the light of Section \ref{section4.1}. RTNet employs 6 kinds of time embedding information if needed, including \{hour-of-day, day-of-week, day-of-month, day-of-year, week-of-year and month-of-year\}. The number of augmented examples generated from each input sequence are 3. The amplitude of data augmentation is 0.2 as default. $ \alpha $ mentioned in Condition \ref{condition} which controls the overlapping degree of adjacent input sequences is set to 4.\par
Each experiment is run in 20 rounds with different random seeds to obtain statistical performances. Average results are shown in this section. MSE (${\sum_{i=1}^n{(x_i-\hat{x}_i)}^2}/{n}$) and MAE (${\sum_{i=1}^n{|x_i-\hat{x}_i|}}/{n}$) are used for forecasting accuracy criteria. Due to space limitation, we only present MSE in the whole Section \ref{section4} and put MAE results together with error bars in Appendix E .\par
Other hyper-parameters are presented in Appendix F. Some of them vary along with the conditions of different datasets. Note that we did not intentionally tune these hyper-parameters for ostensibly better results and most of them keep steady throughout whole experiments.\par
\subsection{Validation Experiments}
\begin{table*}[]
	\renewcommand{\arraystretch}{1.1}
	\caption{MSE of Validation Experiment on Normalization Methods under ETTh$_{1}$}
	\label{tab1}
	\centering
	\setlength\tabcolsep{4pt}
	\begin{threeparttable}
		\begin{tabular}{ccccccccccc}
			\toprule[1.5pt]
			Baseline & \multicolumn{3}{c}{RTNet(E)}  &\multicolumn{3}{c}{RTNet(C)}& \multicolumn{2}{c}{Informer(E)}  & \multicolumn{2}{c}{TS2Vec(C)}  \\
			\cmidrule(lr){2-4}
			\cmidrule(lr){5-7}
			\cmidrule(lr){8-9}
			\cmidrule(lr){10-11}
			Pred\textbackslash{}Norm     & WN     & BN     & LN     & WN     & BN     & LN     & LN       & None   & BN     & None  \\
			\midrule[1pt]
			24       &  \textit{\textbf{0.029}}  & 0.031  & 0.040   &  \textit{\textbf{0.037}}  & 0.045  & 0.048  & 0.098    &  \textit{\textbf{0.095}}  & 0.048  &  \textit{\textbf{0.039}} \\
			48      &  \textit{\textbf{0.045}} & 0.059  & 0.064  &  \textit{\textbf{0.071}}  & 0.075  & 0.083  & 0.158    &  \textit{\textbf{0.147}}  & 0.078  &  \textit{\textbf{0.062}} \\
			168      &  \textit{\textbf{0.072}} & 0.098  & 0.086  &  \textit{\textbf{0.088}}  & 0.094  & 0.178  & 0.183    &  \textit{\textbf{0.164}}  & 0.157  &  \textit{\textbf{0.142}} \\
			336     &  \textit{\textbf{0.083}}  & 0.112  & 0.179  &  \textit{\textbf{0.110}}   & 0.131  & 0.210   & 0.222    &  \textit{\textbf{0.173}}  & 0.188  &  \textit{\textbf{0.160}}  \\
			720      &  \textit{\textbf{0.119}}  & 0.130   & 0.159  &  \textit{\textbf{0.133}}  & 0.181  & 0.229  & 0.269    &  \textit{\textbf{0.145}}  & 0.215  &  \textit{\textbf{0.179}} \\
			\bottomrule[1.5pt] 
		\end{tabular}
	\end{threeparttable}
\end{table*}
\subsubsection{Baselines}
To further validate our proposed hypotheses and corollaries in Section \ref{section2}, we do validation experiments on not only RTNet with end-to-end/contrastive learning based format but also other two typical baselines Informer\cite{informer}/TS2Vec\cite{TS2Vec} which are separately end-to-end/contrastive learning based models. Their initial results are taken from their reports. For different independent validation experiments, different datasets are utilized to illustrate the generality of our proposed hypotheses and corollaries. The MSE results of all experiments are shown and the best MSE result of each situation is highlighted in bold and italicized. In addition, `E' refers to baseline with end-to-end format and `C' denotes baseline with contrastive learning based format.\par
\subsubsection{Normalization}
To validate Corollary \ref{corollary1}, we respectively combine RTNet with WN(default)/LN/BN to do univariate forecasting on ETTh$ _{1} $. As to Informer/TS2Vec, Informer attaches LN as default and TS2Vec does not employ any normalization method \cite{informer,TS2Vec}. Therefore, we remove LN from Informer and add BN into TS2Vec to compare forecasting differences with/without improper normalization methods.\par
As LN helps attention scores converge, we remove LN from decoders of Informer which more heavily affect predictions and keep LN in encoders. Meanwhile, the network of TS2Vec in the second prediction stage is a ridge regression, so we replace the ridge regression with a single linear projection layer in order to employ BN. Forecasting results of total four methods are shown in Tab.\ref{tab1}.\par
It can be observed from Tab.\ref{tab1} that RTNet with WN outperforms itself with BN/LN. Moreover, the performances of RTNet with LN rank third either in end-to-end or contrastive learning based format. These phenomena are identical to the analysis in Section \ref{section2.1}, illustrating that normalization methods directly re-parameterizing hidden units like BN/LN will only do harm to time series forecasting networks. Forecasting results of Informer without LN and TS2Vec with BN further confirm Corollary \ref{corollary1}. \par
The performances of Informer increase after removing LN. In other words, the benefits LN brings fail to compensate the harms. At the same time, the performances of TS2Vec decrease after added BN. During the first stage of TS2Vec, cosine similarity used in contrastive loss is determined only by the direction of the variates but not the magnitude, which is shown as Eq.\ref{eq13}. That means that BN will not cause appreciable influences during the first stage of TS2Vec. Even so, BN still does harm to the forecasting.\par
\begin{eqnarray}
	\label{eq13}
	\cos(a,b) = \dfrac{\vert a^\top b\vert}{\Vert a\Vert\Vert b\Vert} = \dfrac{\vert a^\top kb\vert}{\Vert a\Vert\Vert kb\Vert} = \cos(a,kb)
\end{eqnarray}

Thereupon, the validity and generality of Corollary \ref{corollary1} are verified.\par
\subsubsection{Multivariate Forecasting}
\begin{table}[]
	\renewcommand{\arraystretch}{1.1}
	\caption{The Cos-relation Matrix of ETTh$_1$/ETTm$ _{1} $}
	\label{tab2}
	\centering
	\setlength\tabcolsep{2pt}
	\begin{threeparttable}
		\begin{tabular}{cccccccc}
			\toprule[1.5pt]
			Cosine & OT & HUFL  & HULL  & MUFL  & MULL  & LUFL & LULL \\
			\midrule[1pt]
			OT          & 1  & 0     & 0     & 0     & 0     & 0    & 0    \\
			HUFL        & 0  & 1     & 0     & 0.984 & 0     & 0    & 0    \\
			HULL        & 0  & 0     & 1     & 0     & 0.926 & 0    & 0    \\
			MUFL        & 0  & 0.984 & 0     & 1     & 0     & 0    & 0    \\
			MULL        & 0  & 0     & 0.926 & 0     & 1     & 0    & 0    \\
			LUFL        & 0  & 0     & 0     & 0     & 0     & 1    & 0    \\
			LULL        & 0  & 0     & 0     & 0     & 0     & 0    & 1  \\
			\bottomrule[1.5pt]  
		\end{tabular}
	\end{threeparttable}
\end{table}
\begin{table}[]
	\renewcommand{\arraystretch}{1.1}
	\caption{MSE of Validation Experiment on Cos-relation Matrix under ETTh$_1$}
	\label{tab3}
	\centering
	\setlength\tabcolsep{1.8pt}
	\begin{threeparttable}
		\begin{tabular}{ccccccccc}
			\toprule[1.5pt]
			Baseline & \multicolumn{2}{c}{RTNet(E)}          & \multicolumn{2}{c}{RTNet(C)}          & \multicolumn{2}{c}{Informer}          & \multicolumn{2}{c}{TS2Vec}          \\
			\cmidrule(lr){2-3}
			\cmidrule(lr){4-5}
			\cmidrule(lr){6-7}
			\cmidrule(lr){8-9}
			Pred  & With     & Without & With   & Without & With     & Without & With   & Without \\
			\midrule[1pt]
			24  &  \textit{\textbf{0.340}}    & 0.449   &  \textit{\textbf{0.329}}  & 0.484   &  \textit{\textbf{0.411}}    & 0.577   &  \textit{\textbf{0.493}}  & 0.590   \\
			48   &  \textit{\textbf{0.373}}    & 0.583   &  \textit{\textbf{0.378}}  & 0.536   &  \textit{\textbf{0.478}}    & 0.685   &  \textit{\textbf{0.540}}  & 0.624   \\168 &  \textit{\textbf{0.448}}    & 0.882   &  \textit{\textbf{0.457}}  & 0.865   &  \textit{\textbf{0.599}}    & 0.931   &  \textit{\textbf{0.653}}  & 0.762   \\336 &  \textit{\textbf{0.503}}    & 1.192   &  \textit{\textbf{0.514}}  & 1.085   &  \textit{\textbf{0.657}}    & 1.128   &  \textit{\textbf{0.701}} & 0.931   \\720 &  \textit{\textbf{0.543}}    & 1.148   &  \textit{\textbf{0.550}}  & 1.288   &  \textit{\textbf{0.809}}    & 1.215   &  \textit{\textbf{0.705}}  & 1.063   \\
			\bottomrule[1.5pt]    
		\end{tabular}
	\end{threeparttable}
\end{table}
\begin{table}[t]
	\renewcommand{\arraystretch}{1.1}
	\caption{MSE of Validation Experiment on Cos-relation Matrix under ETTm$_1$}
	\label{tab4}
	\centering
	\setlength\tabcolsep{1.8pt}
	\begin{threeparttable}
		\begin{tabular}{ccccccccc}
			\toprule[1.5pt]
			Baseline & \multicolumn{2}{c}{RTNet(E)}  & \multicolumn{2}{c}{RTNet(C)}  & \multicolumn{2}{c}{Informer} & \multicolumn{2}{c}{TS2Vec}   \\
			\cmidrule(lr){2-3}
			\cmidrule(lr){4-5}
			\cmidrule(lr){6-7}
			\cmidrule(lr){8-9}
			Pred     & With       & Without     & With       & Without    & With       & Without & With        & Without  \\
			\midrule[1pt]
			24       &  \textit{\textbf{0.207}}      & 0.312       &  \textit{\textbf{0.216}}      & 0.263      &  \textit{\textbf{0.271}}       & 0.323   &  \textit{\textbf{0.402}}       & 0.453     \\48     &  \textit{\textbf{0.268}}      & 0.491       &  \textit{\textbf{0.275}}     & 0.359      &  \textit{\textbf{0.365}}      & 0.494   &  \textit{\textbf{0.489}}       & 0.592     \\96    &  \textit{\textbf{0.305}}      & 0.685       & \textit{\textbf{0.300}}      & 0.438      &  \textit{\textbf{0.490}}      & 0.678   &  \textit{\textbf{0.494}}       & 0.635     \\288  &  \textit{\textbf{0.370}}     & 1.094       &  \textit{\textbf{0.365}}      & 0.614      &  \textit{\textbf{0.603}}     & 1.056   &  \textit{\textbf{0.573}}       & 0.693   \\672   &  \textit{\textbf{0.434}}      & 1.204       & \textit{\textbf{0.436}}      & 0.766      &  \textit{\textbf{0.703}}      & 1.192   &  \textit{\textbf{0.626}}       & 0.782     \\
			\bottomrule[1.5pt]
		\end{tabular}
	\end{threeparttable}
\end{table}
\begin{table*}[]
	\renewcommand{\arraystretch}{1.1}
	\caption{MSE of Validation Experiment on Input Sequence Lengths under WTH}
	\label{tab6}
	\centering
	\setlength\tabcolsep{2.3pt}
	\begin{threeparttable}
		\begin{tabular}{ccccccccccccccccccccc}
			\toprule[1.5pt]
			Baseline  & \multicolumn{5}{c}{RTNet(E)}           & \multicolumn{5}{c}{RTNet(C)}& \multicolumn{5}{c}{Informer(E)}           & \multicolumn{5}{c}{TS2Vec(C)}  \\
			\cmidrule(lr){2-6}
			\cmidrule(lr){7-11} 
			\cmidrule(lr){12-16}\cmidrule(lr){17-21}
			Pred$\backslash$Input & 32                                             & 48                                             & 96    & 192   & 384   & 32                                             & 48                                             & 96    & 192   & 384   & 32                                             & 48                                             & 96    & 192   & 384   & 32                                             & 48                                             & 96    & 192   & 384   \\ \midrule[1pt]
			24    & 0.090                                          &  \textit{\textbf{0.089}} & 0.091 & 0.094 & 0.098 & 0.094                                          &  \textit{\textbf{0.093}} & 0.098 & 0.097 & 0.099 &  \textit{\textbf{0.101}} & 0.106                                          & 0.112 & 0.137 & 0.152 &  \textit{\textbf{0.097}} & 0.097                                          & 0.099 & 0.097 & 0.102 \\
			48                                                         & 0.134                                          &  \textit{\textbf{0.131}} & 0.134 & 0.139 & 0.144 & 0.141                                          &  \textit{\textbf{0.139}} & 0.142 & 0.144 & 0.144 & 0.150                                          &  \textit{\textbf{0.149}} & 0.177 & 0.181 & 0.200 &  \textit{\textbf{0.139}} & 0.140                                          & 0.140 & 0.140 & 0.142 \\
			168                                                        & 0.193                                          &  \textit{\textbf{0.192}} & 0.195 & 0.207 & 0.236 &  \textit{\textbf{0.191}} & 0.194                                          & 0.198 & 0.202 & 0.202 & 0.250                                          &  \textit{\textbf{0.236}} & 0.255 & 0.256 & 0.282 & 0.203                                          &  \textit{\textbf{0.200}} & 0.202 & 0.206 & 0.208 \\
			336                                                        &  \textit{\textbf{0.206}} & 0.207                                          & 0.214 & 0.223 & 0.233 &  \textit{\textbf{0.208}} & 0.212                                          & 0.218 & 0.222 & 0.227 &  \textit{\textbf{0.250}} & 0.260                                          & 0.268 & 0.275 & 0.303 & 0.224                                          &  \textit{\textbf{0.223}} & 0.226 & 0.231 & 0.230 \\
			720                                                        & 0.202                                          &  \textit{\textbf{0.199}} & 0.201 & 0.203 & 0.209 &  \textit{\textbf{0.203}} & 0.208                                          & 0.212 & 0.213 & 0.220 &  \textit{\textbf{0.257}} & 0.265                                          & 0.267 & 0.276 & 0.283 & 0.234                                          &  \textit{\textbf{0.230}} & 0.233 & 0.233 & 0.237\\
			\bottomrule[1.5pt]
		\end{tabular}
	\end{threeparttable}
\end{table*}
\begin{table}[]
	\renewcommand{\arraystretch}{1.1}
	\caption{Ranks of All Input Sequence Lengths in Tab.\ref{tab6}}
	\label{tab7}
	\centering
	\begin{threeparttable}
		\begin{tabular}{cccccc}
			\toprule[1.5pt]
			Input & 48 & 32 & 96 & 192 & 384 \\
			\midrule[1pt]
			Times of first &  \textit{\textbf{11}} & 9 & 0 & 0 & 0 \\
			Rank & 1 & 2 & 3 & 3 & 3 \\
			\bottomrule[1.5pt]
		\end{tabular}
	\end{threeparttable}
\end{table}

As Corollary \ref{corollary3} has already been proved in Appendix C3 rigorously, we only verify Corollary \ref{corollary2} and the function of cos-relation matrix. Multivariate forecasting conditions of ETTh$ _{1} $ and ETTm$ _{1} $ are used in this experiment. The threshold $ \theta $ of the matrix is set to 45$ \degree $. For the sake of applying cos-relation matrix and the concept of group convolution to Informer and TS2Vec. The multi-head attention of Informer is replaced with multi-group attention (number of heads = number of groups) to maintain the network independence of different variates. Ridge regression of TS2Vec is also replaced with a single group convolutional layer. The cos-relation matrix of two datasets are shown in Tab.\ref{tab2} and forecasting results of four methods are shown in Tab.\ref{tab3}/\ref{tab4}.\par
Note that the cos-relation matrix of ETTh$ _{1} $/ETTm$ _{1} $ is the same even to three decimal places so there is only one table for cos-relation matrix. This is not surprising for ETTm$ _{1} $ is in fact the down sampling of ETTh$ _{1} $. This phenomenon implicitly demonstrates that cos-relation matrix could somehow represent the general relationship among the variates. Tab.\ref{tab2} also shows that variates in ETTh$ _{1} $/ETTm$ _{1} $ have weak connections with each other. Only the pairs of MUFL-HUFL and MULL-HULL are related to each other in Tab.\ref{tab2}. This is identical to reality in that they are simultaneous useful load (UFL) or useless load (ULL). Coupled with forecasting results in Tab.\ref{tab3}/\ref{tab4} where baselines with cos-relation matrix greatly outperform those without it, the validity and generality of Corollary \ref{corollary2} are verified.\par
\subsubsection{Input sequence length}
\begin{figure*}[b]
	\centering
	\includegraphics[width=5.5in]{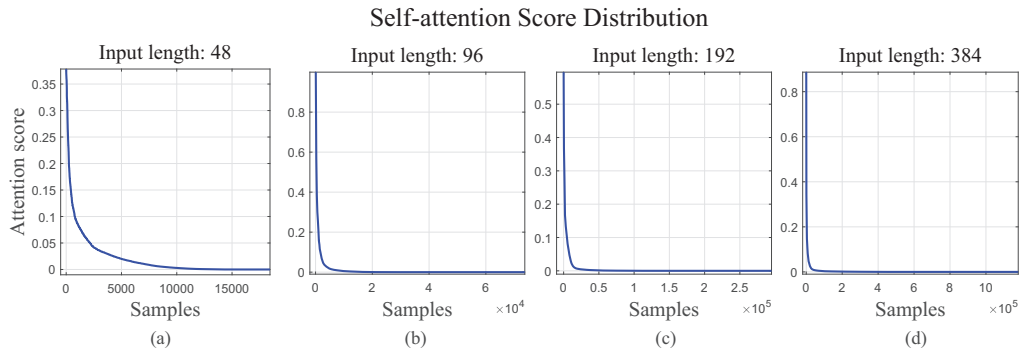}
	\caption{Self-attention score distribution of encoder with input length in \{48, 96, 192, 384\}. The sparsity of the score distributions is getting more apparent with the prolonging of input length. This indicates that elements are likely to be only related to adjacent neighbors.}
	\label{fig12}
\end{figure*}
\begin{figure*}[]
	\centering
	\includegraphics[width=5.5in]{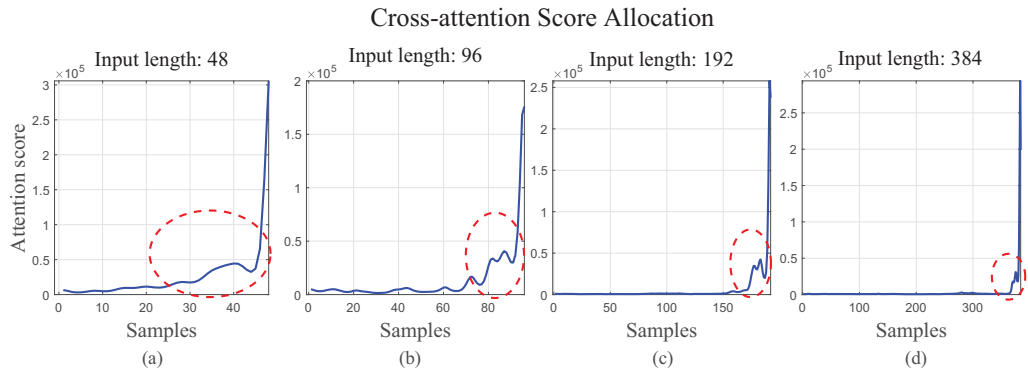}
	\caption{Cross-attention score allocation of decoder with input length in \{48, 96, 192, 384\}. The shapes of the curves are similar if we only concentrate on the last dozens of elements (red circle). It is also distinct that the prolonged earlier elements make very limited contributions to the cross-attention scores in all sub-figures, i.e. there exists no long-term dependency.}
	\label{fig13}
\end{figure*}
\begin{table*}[]
	\renewcommand{\arraystretch}{1.1}
	\caption{Average MSE Results of Prediction Groups in Comparison Experiment}
	\label{tab8}
	\centering
	\setlength\tabcolsep{2pt}
	\begin{threeparttable}
		\begin{tabular}{ccccccccccccccccccc}
			\toprule[1.5pt]
			Task & \multicolumn{10}{c}{S} & \multicolumn{8}{c}{M}\\
			\cmidrule(lr){2-11}
			\cmidrule(lr){12-19}
			Baseline\textbackslash{}Dataset & ETTh$ _{1} $ & Rank & ETTh$ _{2} $ & Rank & ETTm$ _{1} $ & Rank & WTH   & Rank & ECL   & Rank & ETTh$ _{1} $ & Rank & ETTh$ _{2} $ & Rank & ETTm$ _{1} $ & Rank & WTH   & Rank\\
			\midrule[1pt]
			RTNet(E)    & 0.070        & 1    & 0.140        & 1    & 0.034        & 1    & 0.163 & 1    & 0.116 & 1    & 0.442        & 1    & 1.106        & 3    & 0.317        & 1    & 0.426 & 2    \\
			RTNet(C)    & 0.088        & 2    & 0.150        & 2    & 0.057        & 3    & 0.169 & 2    & 0.128 & 2    & 0.445        & 2    & 0.965        & 2    & 0.318        & 2    & 0.422 & 1    \\
			CoST(C)     & 0.091        & 3    & 0.161        & 3    & 0.054        & 2    & 0.183 & 7    & 0.478 & 7    & 0.650        & 4    & 1.283        & 4    & 0.409        & 3    & 0.430 & 3    \\
			TSFresh     & 0.104        & 4    & 0.255        & 15   & 0.069        & 7    & 0.292 & 13   & ---     & ---    & 3.441        & 19   & 15.020       & 19   & 0.767        & 14   & 2.336 & 19   \\
			MoCo(C)     & 0.110        & 5    & 0.168        & 5    & 0.057        & 4    & 0.176 & 4    & 0.478 & 6    & 0.846        & 7    & 1.503        & 5    & 0.639        & 6    & 0.444 & 5    \\
			TS2Vec(C)   & 0.116        & 6    & 0.165        & 4    & 0.065        & 6    & 0.181 & 6    & 0.486 & 8    & 0.794        & 6    & 1.543        & 6    & 0.631        & 5    & 0.451 & 6    \\
			TCC(C)      & 0.119        & 7    & 0.192        & 8    & 0.108        & 10   & 0.179 & 5    & 0.507 & 9    & 0.988        & 12   & 2.516        & 13   & 0.687        & 8    & 0.457 & 7    \\
			SCINet(E)   & 0.137        & 8    & 0.249        & 14   & 0.059        & 5    & 0.239 & 9    & 0.345 & 4    & 0.565        & 3    & 0.893        & 1    & 0.477        & 4    & 0.500 & 10   \\
			CPC(C)      & 0.147        & 9    & 0.197        & 9    & 0.081        & 9    & 0.187 & 8    & 0.516 & 10   & 0.926        & 11   & 1.856        & 8    & 0.697        & 9    & 0.462 & 8    \\
			TNC(C)      & 0.150        & 10   & 0.168        & 6    & 0.069        & 8    & 0.175 & 3    & 0.474 & 5    & 0.904        & 8    & 1.869        & 9    & 0.740        & 12   & 0.441 & 4    \\
			Triplet(C)  & 0.162        & 11   & 0.208        & 11   & 0.108        & 11   & 0.240 & 10   & 0.555 & 11   & 1.104        & 16   & 1.816        & 7    & 0.782        & 15   & 0.562 & 13   \\
			Informer(E) & 0.186        & 12   & 0.204        & 10   & 0.241        & 15   & 0.243 & 11   & 0.607 & 13   & 0.907        & 9    & 2.371        & 12   & 0.749        & 13   & 0.574 & 14   \\
			LogTrans(E) & 0.196        & 13   & 0.217        & 12   & 0.270        & 16   & 0.280 & 12   & 0.795 & 16   & 1.043        & 15   & 2.898        & 16   & 0.965        & 16   & 0.645 & 15   \\
			N-BEATS(E)  & 0.218        & 14   & 0.326        & 18   & 0.162        & 13   & ---     & ---    & 0.891 & 17   & ---            & ---    & ---            & ---    & ---            & ---    & ---     & ---    \\
			TCN(E)      & 0.263        & 15   & 0.180        & 7    & 0.183        & 14   & 0.391 & 17   & 0.673 & 15   & 1.021        & 14   & 2.564        & 14   & 0.719        & 10   & 0.558 & 12   \\
			DeepAR(E)   & 0.322        & 16   & 0.264        & 16   & 0.777        & 22   & 0.342 & 15   & 0.337 & 3    & ---            & ---    & ---            & ---    & ---            & ---    & ---     & ---    \\
			ARIMA(E)    & 0.325        & 17   & 3.003        & 21   & 0.292        & 17   & 0.518 & 19   & ---     & ---    & ---            & ---    & ---            & ---    & ---            & ---    & ---     & ---    \\
			LSTMa(E)    & 0.334        & 18   & 0.264        & 17   & 0.422        & 19   & 0.353 & 16   & ---     & ---    & 1.009        & 13   & 2.671        & 15   & 1.423        & 18   & 1.000 & 17   \\
			LSTNet(E)   & 0.363        & 19   & 3.035        & 22   & 0.328        & 18   & 0.397 & 18   & 0.672 & 14   & 1.909        & 18   & 3.343        & 18   & 1.981        & 19   & 0.731 & 16   \\
			TST(C)      & 0.413        & 20   & 0.223        & 13   & 0.122        & 12   & 0.318 & 14   & 0.573 & 12   & 0.911        & 10   & 2.054        & 11   & 0.667        & 7    & 0.488 & 9    \\
			Reformer(E) & 1.107        & 21   & 0.349        & 19   & 0.741        & 21   & 0.927 & 20   &       &      & 1.566        & 17   & 3.298        & 17   & 1.278        & 17   & 1.352 & 18   \\
			Prophet(E)  & 1.126        & 22   & 1.583        & 20   & 0.690        & 20   & 1.760 & 21   & ---     & ---    & ---            & ---    & ---            & ---    & ---            & ---    & ---     & ---    \\
			StemGNN(E)  & ---            & ---    & ---            & ---    & ---            & ---    & ---     & ---    & ---     & ---    & 0.738        & 5    & 1.940        & 10   & 0.729        & 11   & 0.503 & 11  \\  
			\bottomrule[1.5pt]
		\end{tabular}
		\begin{tablenotes}
			\footnotesize
			\item[1] The result is the average MSE of prediction length group \{24, 48, 96, 288, 672\} with ETTm$ _{1} $ or \{24, 48, 168, 336, 720\} with other dataset.
			\item[2] RTNet results of each prediction length group are average results run in 20 rounds.
		\end{tablenotes}
	\end{threeparttable}
\end{table*}

We prolong the input sequence length within \{32, 48, 96, 192, 384\} to do univariate forecasting on WTH for further verifying Corollary \ref{corollary5}. Informer and TS2Vec are in their intact formats. Forecasting results and ranks of all input sequence lengths according to forecasting performances are shown in Tab.\ref{tab6} and \ref{tab7} separately. As to the verification of Corollary \ref{corollary6}/\ref{corollary7}, we utilize a canonical Transformer architecture composed of one encoder and one decoder to examine the exact sparse attention score distributions under the same settings. We fix the prediction length as 24 and prolong the input sequence length within \{48, 96, 192, 384\} to study the attention score distribution. Fig.\ref{fig12} (a)-(d) show the total self-attention score distributions of the encoder sampled from all prediction windows. Fig.\ref{fig13} (a)-(d) show the cross-attention score allocations of all input sequence elements which are the sum of corresponding decoder cross-attention scores sampled from all prediction windows.\par
As Tab.\ref{tab6}/\ref{tab7} shows, all of baselines acquire better performances when input sequence length is 32/48 for all prediction lengths in this experiment. It indicates that prolonging the input sequence length exaggeratedly to `satisfy' long-term forecasting may even cause the drop of forecasting accuracy. Meanwhile, it is very likely that the most appropriate input sequence length is within a small area around 48. So Corollary \ref{corollary5} is verified empirically and we choose 48 as the input sequence length to do forecasting on WTH.\par
Fig.\ref{fig12} and \ref{fig13} illustrate that with prolonging the input sequence length, the attention score matrices are getting more and more sparser. It is obvious that both self-attention and cross-attention scores form long-tail distributions. Then Corollary \ref{corollary6} is established empirically. Additionally, as to cross-attention score allocations in Fig.\ref{fig13}, it can be observed that only the last few elements dominate total cross-attention scores. Therefore, the behavior of prolonging input sequence length to acquire long-term dependency and simultaneously constructing sparse attention to reduce computation complexity are contradictory to each other and will only burden the forecasting, i.e. Corollary \ref{corollary7} is established empirically. It also enlightens us to cautiously examine the existence of long-term dependency before trying to extracting it from a certain dataset.\par
Conclusively, all corollaries are verified theoretically and empirically on the basis of hypothesis in Section \ref{section2}.\par
\subsection{Comparison Experiment}
We perform comparison experiments on univariate (S) and multivariate (M) forecasting to examine the performances of RTNet in both end-to-end and contrastive learning based formats. Baselines we select contain end-to-end models, contrastive learning based models (CoST\cite{Cost}, TS2Vec\cite{TS2Vec}, TNC\cite{TNC}, MoCo\cite{moco}, Triplet\cite{T_loss}, CPC\cite{cpc}, TST\cite{TST}, TCC\cite{TS_TCC}) and a feature engineered model (TSFresh package). End-to-end models include traditional time series forecasting models (ARIMA\cite{box1968,box2015}, Prophet\cite{Prophet}, N-BEATS\cite{N_BEATS}), CNN (SCINet\cite{scinet}, TCN\cite{TCN}), RNN (LSTNet\cite{LSTNet}, DeepAR\cite{deepar}, LSTMa\cite{LSTMa}), Transformer (LogTrans\cite{logtrans}, Informer\cite{informer}, Reformer\cite{Reformer}) and GNN (StemGNN\cite{StemGNN}). Most of the results are taken from other papers \cite{informer,TS2Vec,Cost} and the rest of them are compensated by us using unified settings for fair comparison. Note that not all models are available for all forecasting conditions owing to their own limits, e.g. ARIMA\cite{box1968,box2015} is unable to do multivariate forecasting. Average MSE results of prediction length group (\{24, 48, 168, 336, 720\}/\{24, 48, 96, 288, 672\}) with each dataset are shown in Tab.\ref{tab8}, together with the accuracy rank of each model, and the whole results are shown in Appendix E3. We also put additional experiments results there, including abaltion study, parameter sensitivity, etc.\par
It could be observed that RTNet in end-to-end and contrastive learning based formats outperform other models in nearly all conditions. Large amounts of forecasting accuracy records are greatly broken through by RTNet even though only considering average performances. For instance, the MSE of RTNet with the prediction length of 24 in ETTh$ _{1} $ is the first and only one that is lower than 0.03 (Watch Tab.15-23 in Appendix E3.2).\par
\begin{figure}[]
	\centering
	\subfloat[]{\includegraphics[height=1.37in]{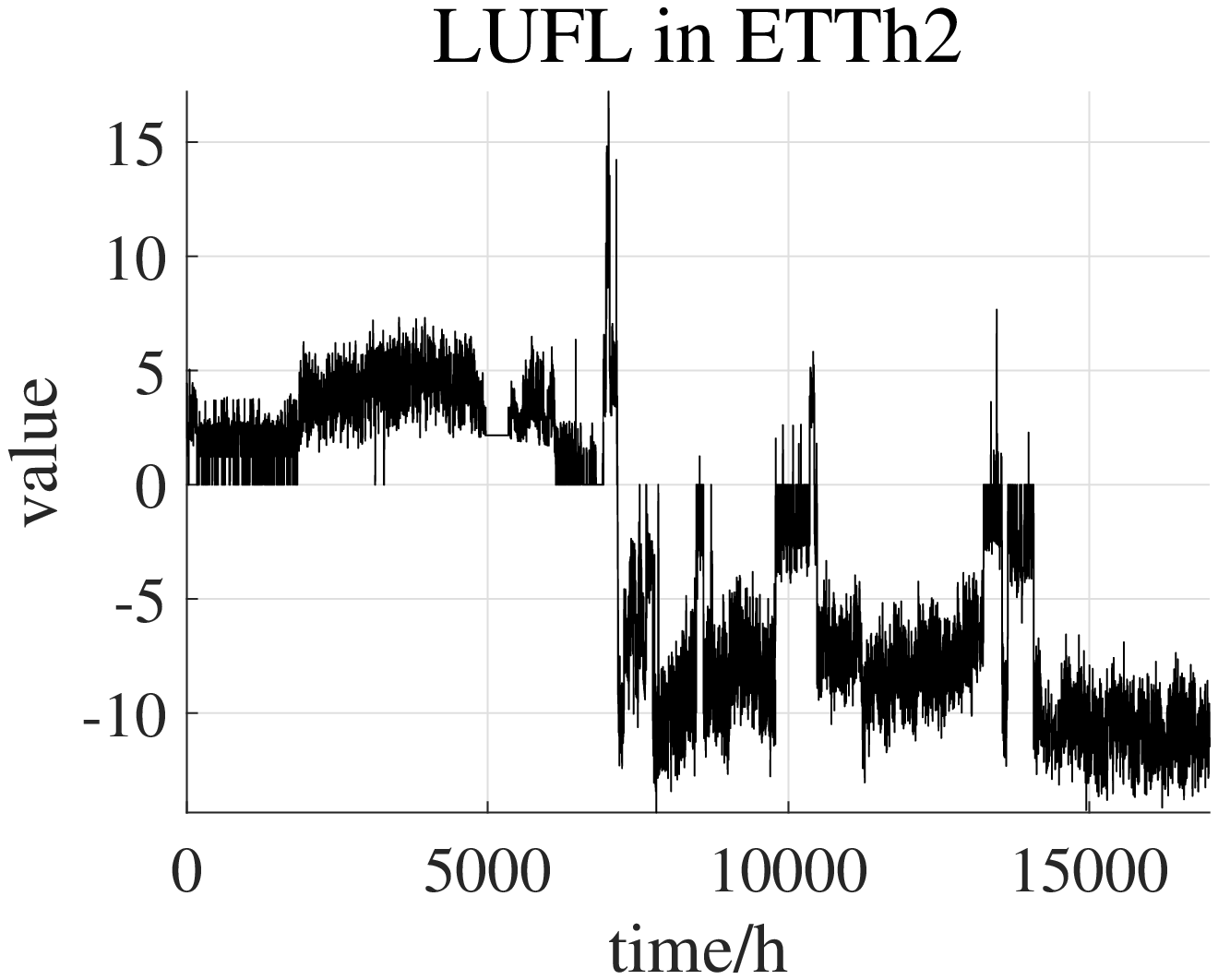}\label{fig16a}}
	\subfloat[]{\includegraphics[height=1.37in]{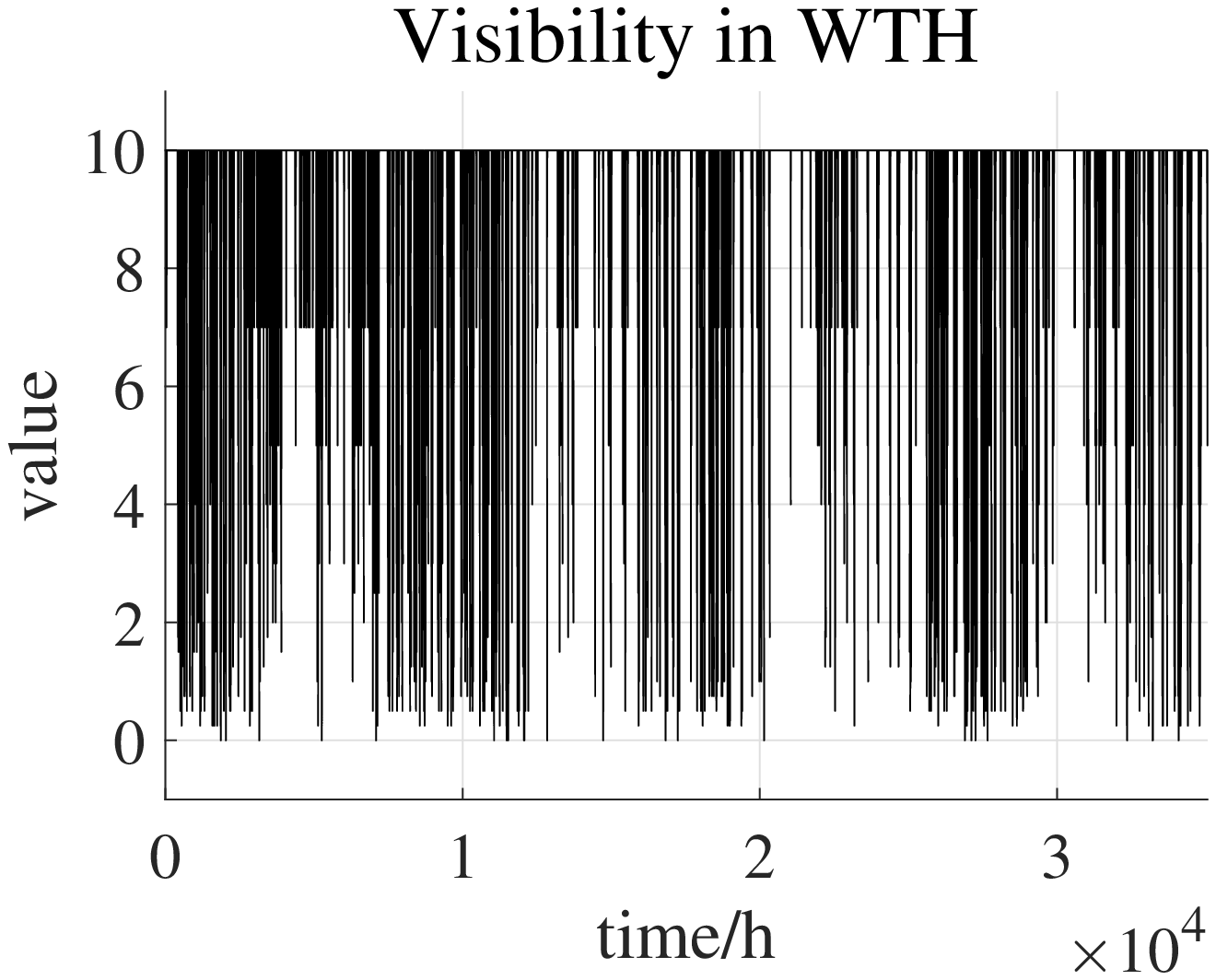}\label{fig16b}}
	\caption{Examples of variates owning non-autoregressive parts or frequent anomalies, including `LUFL' in ETTh$ _{2} $ and `Visibility' in WTH.}
	\label{fig16}
\end{figure}

Besides, RTNet in end-to-end format surpasses itself in contrastive learning based format except doing multivariate forecasting with ETTh$ _{2} $ and WTH. As contrastive learning based method aims to learn more universal features and ignores subtle details, the performance of RTNet in contrastive learning based format fails to challenge the performance of itself in end-to-end format in most of situations where both universal and detailed features need to be learned. However, when handling forecasting situations which have frequent anomaly or are not strictly auto-regressive, paying much attention to details will guide the model to lose the capability of extracting universal features and cause severe overfitting problem. Taking a look at the data distribution of `LUFL' of ETTh$ _{2} $ and `Visibility' of WTH as Fig.\ref{fig16}, it can be observed that multivariate forecasting with ETTh$ _{2} $ and WTH are such conditions so contrastive learning based forecasting format is more suitable for them.\par

\subsection{Computation Efficiency}
We choose SCINet and TS2Vec (linear projection layer at the second stage) as baselines to examine the computation efficiency of RTNet. They are relatively fast respectively among methods in end-to-end/costrastive learning based forecasting formats. The total training and inference time comparison under the univariate forecasting of ETTh$ _{1} $ is shown in Tab.\ref{taba18}. It can be observed that RTNet is much more faster than SCINet and TS2Vec in all occasions, especially RTNet with contrastive learning based format, thanks to the mechanism of shrinking input sequence lengths during feature extraction and fixed input sequence lengths. Furthermore, the training and inference time of RTNet/TS2Vec decreases when prolonging prediction length within \{336, 720\}. It is caused by the increasing risk of overfitting which leads to fewer participated epoches during training phase due to the mechanism of early stopping. Meanwhile, the training and inference time of SCINet still increases even with fewer epochs, illustrating that element-wise feature extraction will bring great amount of extra computation, especially with those end-to-end forecasting methods.\par
\begin{table}[]
	\renewcommand{\arraystretch}{1.2}
	\caption{The Running Time Comparison under ETTh$_1$}
	\label{taba18}
	\centering
	\setlength\tabcolsep{1.8pt}
	\renewcommand{\multirowsetup}{\centering}	
	\begin{threeparttable}
		\begin{tabular}{ccccc}
			\toprule[1.5pt]
			Pred\textbackslash{}Baselines & RTNet(E) & RTNet(C) & SCINet   & TS2Vec  \\
			\midrule[1pt]
			24        & 21s      &  \textit{\textbf{11s}}     & 4min13s  & 1min3s  \\
			48        & 24s      &  \textit{\textbf{11s}}      & 4min50s  & 1min14s \\
			168       & 35s      &  \textit{\textbf{11s}}     & 7min55s  & 1min12s \\
			336       & 20s      &  \textit{\textbf{11s}}     & 9min21s  & 1min09s \\
			720       & 18s      &  \textit{\textbf{11s}}     & 11min08s & 1min08s\\
			\bottomrule[1.5pt]
		\end{tabular}
	\end{threeparttable}
\end{table}

\section{Discussion}
Unique and complex real-world time series determine that time series forecasting networks need to respect times series properties for perfect performances and some state-of-the-art methods from other AI fields may not be suitable for them. All corollaries, analysis, methods and experiments proposed in this paper are meant to prove this idea and their results are in expectations. Corollaries are verified theoretically and empirically while methods (mainly RTNet) break through amount of forecasting records. The forecasting capability of other state-of-the-art baselines will also be greatly improved when considering our corollaries. In fact, methods we provided are just some of feasible ways to verify the rationality and effectiveness of respecting time series properties. Any component/theory basis of RTNet has limitations to some extents on the basis of above analysis/experiments. That means that RTNet owns significant potentials and could be combined with more practical methods/theories to achieve better performances as long as they respect time series properties. This is also our future research direction.\par
\section{Conclusion}
In this paper, we first propose the idea of respecting time series properties and point out three common problems of time series forecasting networks brought by blindly borrowing state-of-the-art methods from other AI fields or not taking time series properties into account. Then we give out corresponding corollaries, analysis and solutions of these problems. Based on aforementioned corollaries, we propose RTNet which can do time series forecasting flexibly in both end-to-end and contrastive learning based formats. Extensive validation experiments demonstrate the validity of respecting time series properties for time series forecasting networks while comparison experiments illustrate the promising forecasting ability of RTNet.\par
\appendices
\section{Preliminary}
We present the definition of time series forecasting, the form of contrastive learning based forecasting and two commonly-used forecasting strategies here.\par
\textbf{Time series forecasting:} Given input time series sequence $\{x_i^j\}, i\in[1,t], j\in[1,N]$, time series forecasting task is to predict the following forecasting sequence $\{x_i^j\}, i\in[t+1,T], j\in[1,N]$. $N$ is the number of forecasting variates, $t$ is the length of input window and ($T-t$) is the length of output window.\par
\textbf{Contrastive learning based forecasting:} Contrastive learning based time series forecasting framework contains two stages. The first stage of model is designed to map input sequences into a latent space and extract representations of the input sequences through contrastive loss function \cite{simclr,Cost,TNC,T_loss}. The second stage of model is trained to obtain the prediction sequence from representations via MLP \cite{simclr}, ridge regression \cite{TS2Vec}, SVM \cite{Cost}, etc. Grads of representations of the first stage are detached during the second stage training.\par
\textbf{Multi-step rolling forecasting strategy:} Rolling forecasting strategy\cite{deepar,multi_stepTKDE,Multistep} uses ($T-t$) steps to forecast the whole prediction window. $k$th step forecasts the elements at time stamp ($t+k$) with input sequence window $\{x_i^j\}, i\in[k, t+k-1], j\in[1,N]$. It could be seen that input sequence varies with the processing of the forecasting.\par
\textbf{One-step one-forward forecasting strategy:} One-forward forecasting strategy uses only one step to forecast the whole prediction window \cite{informer,scinet,TS2Vec}. The whole prediction sequence is simultaneously obtained in one-forward procedure through the initial input sequence window.\par
\section{Related Works}
With the growing demand of long-term time series forecasting accuracy in various domains \cite{GraphAttention_TKDE,multi_stepTKDE,trafficTKDE,CRESPOCUARESMA200487, kim2003financial, li2017diffusion, zeroual2020deep, zhao2021empirical, cao2021spectral}, traditional forecasting models, e.g. ARIMA \cite{box1968,box2015}, SES \cite{SES}, are no longer able to deal with more and more complicated forecasting situations. Thus, the research of deep time series forecasting models becomes prevalent. Starting from RNN \cite{LSTM,GRU,LSTM_TKDE,deepar,LSTNet,LSTMa,TNC}, popular networks which are successful in other research fields are successively applied to time series forecasting, like CNN \cite{scinet,TCN,TS2Vec,Cost,T_loss}, GNN \cite{StemGNN,stfgnn,MTGNN} and Transformer \cite{logtrans,informer,Autoformer,FEDformer,Yformer,KSparse,ETSformer}. They are mainly built upon the hypothesis that time series are causal, auto-regressive and stationary. The causality of time series is with no doubt but the other two are simultaneously established only in ideal occasions. There are some other researches \cite{Cost, FEDformer, MSTL,seasonal,STR,STL} trying to decouple seasonal and trend terms of time series which is also not realistic if the time series is not simultaneously auto-regressive and stationary. Sarcastically, most of real-world datasets used in their experiments are coincidentally not stationary or even not auto-regressive. Without respecting this certain truth and other time series properties, their performances may advance owing to more complicated network architectures but are far from perfection. As the auto-regression assumption of time series seems impossible to neglect in any forecasting task, we mainly discuss realistic forecasting conditions when time series are auto-regressive but may be not stationary in this paper.\par

\section{Supplymentary of Corollaries and Analysis}
\subsection{Examples to Illustrate Problems Brought by BN/LN}
Without loss of generality, we use a very simple example to further illustrate why BN is not suitable for time series forecasting networks.\par
For an auto-regressive time series $x(t + 1) =f(x(t)) = 2x(t)$, we take $B_1$ and $B_2$ as different batches, $I_t$ refers to instance in these two batches where each instance is only an element at time $t$ and $t{=}1,2,3,4$ as Eq.\ref{eq2}. The $B_1/ B_2$ will be transformed to $B_1'/ B_2'$ (Eq.\ref{eq3}) if applying BN. Obviously, as an injective model, a single neural network can never describe such multi-valued mapping. Note that standardization preprocessing of raw data will not solve this problem. Because the standardization is performed on every element in the dataset rather than each prediction window of one mini-batch, it does not fundamentally eliminate the scale disparity among prediction windows though may alleviate it.\par
\begin{align}
	\label{eq2}
	B_1&=\{I_1, I_2\}, I_1=1\ \ f(I_1)=2, I_2=2\ \ f(I_2)=4 \notag\\
	B_2&=\{I_3, I_4\}, I_3=4\ \ f(I_3)=8, I_4=8\ \ f(I_4)=16
\end{align}
\begin{align}
	\label{eq3}
	B_1'&=\{I_1', I_2'\}, I_1'=-1\ f(B_1')=2, I_2'=1\ f(I_2')=4 \notag \\
	B_2'&=\{I_3', I_4'\}, I_3'=-1\ f(I_3')=8, I_4'=1\ f(I_4')=16
\end{align}

We also give another example for further depiction of problems brought by LN during time series forecasting.\par
Another simple auto-regressive time series is shown as Eq.\ref{eq4}, where $B_j (j=\{1,2\})$ are different mini-batches each composed of only one instance $L_j$ (sequence) as Eq.\ref{eq5}. $e_{t}$ refers to element at time-step $t$ of the sequence and $t$ is from 1 to 6. The $B_1/B_2$ will be converted into $B_1'/B_2'$ as Eq.\ref{eq6} after applied LN. Evidently, it is also a multi-valued mapping which a single neural network can never describe.\par

\begin{eqnarray}
	\label{eq4}
	x(t + 2) =f(x(t)) = x(t) + x(t + 1)
\end{eqnarray}
\begin{align}
	\label{eq5}
	&B_1=\{L_{1}\} = \{\{e_{1}, e_{2}\}\}\notag\\ 
	&e_{1}\ =1 \ \  e_{3}=f(e_{1})=e_{1} +e_{2}=3 \notag\\ 
	&e_{2}\ =2 \ \  e_{4}=f(e_{2})=e_{2} + e_{3}=5\notag\\
	&B_2=\{L_{2}\}=\{\{e_{3}, e_{4}\}\}\notag\\ 
	&e_{3}\ =3\ \  e_{5}=f(e_{3})=e_{3} + e_{4}=8 \notag\\  
	&e_{4}\ =5\ \  e_{6}=f(e_{4})=e_{4}+ e_{5}=13
\end{align}
\begin{align}
	\label{eq6}
	&B_1'=\{L_{1}'\} = \{\{e_{1}', e_{2}'\}\}\notag\\ 
	&e_{1}'\ =-1 \ \  e_{3}'=f(e_{1}')=e_{1} +e_{2}=3 \notag\\ 
	&e_{2}'\ =1 \ \ \ \ \   e_{4}'=f(e_{2}')=e_{2}' + e_{3}'=5\notag\\
	&B_2'=\{L_{2}'\}=\{\{e_{3}', e_{4}'\}\}\notag\\ 
	&e_{3}'\ =-1\ \  e_{5}'=f(e_{3}')=e_{3}' + e_{4}'=8 \notag\\  
	&e_{4}'\ =1\ \ \ \ \  e_{6}'=f(e_{4}')=e_{4}'+ e_{5}'=13
\end{align}

Additionally, these two examples also illustrate that time series are not scale invariant for forecasting tasks, which proves the rationality of P.1 proposed in Section 2 of the main text.\par
\subsection{Example to Illustrate the Deficiency of Maintaining Input Sequence Length in Latent Space}
\begin{figure}[!t]
	\centering
	\includegraphics[width=3in]{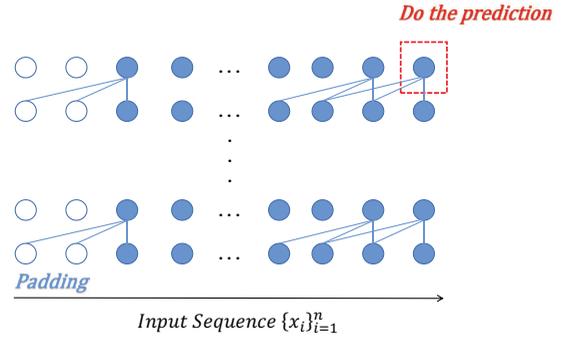}
	\caption{Rolling forecasting format of causal convolution. Only the last element framed by red dotted line is used to do the prediction. However, other elements in the same layer do not participate the prediction, causing computation waste. Besides, it will lead to weak capability of capturing long-term dependency.}
	\label{fig2}
\end{figure}

Typical causal convolution architecture is used here (Fig.\ref{fig2}) to show the deficiency of maintaining input sequence length in latent space. Obviously, it will bring unnecessary computation and make network unable to capture amply global features of input sequence.\par
\subsection{Proof of Corollary 3}
We will prove Corollary 3 by induction. Suppose sequence length is 1 and the number of variates is $ N $. The input sequence is a $ 1\times N $ matrix $ x $ as Eq.\ref{eqa1}. Cos-relation matrix is a $ N\times N $ matrix $ C $ as Eq.\ref{eqa2}. Let $ y=x\cdot C $ be Eq.\ref{eqa3}. To simplify the further discussion, we only concern linear components and ignore bias. Without loss of generality, we utilize group convolution to denote independent networks for each variate just as our application in RTNet and take hidden units in the first group as an example for further discussion owing to the independence and equivalence of each group. Let $ W(k) $ be the weight matrix of the first group in the $ k$th layer as Eq.\ref{eqa4}, $ m_{k} $ is the number of the output hidden dimension of the first group in the $ k$th layer and $ f_{k}(\cdot) $ denotes the $ k$th linear projection of the first group as Eq.\ref{eqa5}. $ O_{k} $ refers to the output of the first group in the $ k$th layer.\par
\begin{eqnarray}
	\label{eqa1}
	x=[x_1,x_2,\dots,x_N]
\end{eqnarray}
\begin{eqnarray}
	\label{eqa2}
	C=
	\begin{pmatrix}  
		c_{11} & \cdots & 	c_{1N} \\  
		\vdots & \ddots & \vdots \\  
		c_{N1} & \cdots & c_{NN}  
	\end{pmatrix} 
\end{eqnarray}
\begin{eqnarray}
	\label{eqa3}
	y{=}\!
	\begin{pmatrix}  
		x_1c_{11}+\dots+x_Nc_{N1} \\  
		x_1c_{12}+\dots+x_Nc_{N2} \\ 
		\cdots \\  
		x_1c_{1N}+\dots+x_Nc_{NN} 
	\end{pmatrix}^\top\!=\!\begin{pmatrix}  
		\sum_{i\in[1,N]}x_ic_{i1} \\  
		\sum_{i\in[1,N]}x_ic_{i2} \\ 
		\cdots \\  
		\sum_{i\in[1,N]}x_ic_{iN} 
	\end{pmatrix}^\top
\end{eqnarray}
\begin{eqnarray}
	\label{eqa4}
	W(k)=
	\begin{pmatrix}  
		w(k)_{11} & \cdots & 	w(k)_{1m_{k-1}} \\  
		\vdots & \ddots & \vdots \\  
		w(k)_{m_{k}1} & \cdots & w(k)_{m_{k}m_{k-1}}  
	\end{pmatrix} 
\end{eqnarray}
\begin{eqnarray}
	\label{eqa5}
	f_k(x)=W(k)\cdot x 
\end{eqnarray}

For the basic case, we need to show that the Corollary 3 is true when $ k=1 $ and $ m_0=1 $. The first group here is $ \sum_{i\in[1,N]}x_ic_{i1} $ so the output of it is shown as Eq.\ref{eqa6}.\par
\begin{eqnarray}
	\label{eqa6}
	O_1=f_1(y_1)=W_1\cdot y_1 = \begin{pmatrix}  
		w(1)_{11}\sum_{i\in[1,N]}x_ic_{i1}  \\  	w(1)_{21}\sum_{i\in[1,N]}x_ic_{i1}  \\  
		\cdots\\  
		w(1)_{m_{1}1}\sum_{i\in[1,N]}x_ic_{i1}    
	\end{pmatrix}
\end{eqnarray}

It could be easily observed that 
$c_{i1}/\sum c_{i1}=w(1)_{j1}c_{i1}/\sum w(1)_{j1}c_{i1}\ (\forall i\in[i,N], j\in[1,m_1])$ owing to homogeneity of linear projection. Then Corollary 3 is true when $k=1$. \par
For the induction step, assume that the Corollary 3 is true when $ k=l,\ l \in N$. It is equivalent to $c_{i1}/\sum c_{i1}=a_jc_{i1}/\sum a_jc_{i1}\ (\forall i\in[i,N], j\in[1,m_l])$ for the first group where $ a_j $ is a certain vector. Then $ O_l $ will be:
\begin{eqnarray}
	\label{eqa7}
	O_l=f_l(\dots(f_1(y_1)))= \begin{pmatrix}  
		a_1\sum_{i\in[1,N]}x_ic_{i1}  \\  	a_2\sum_{i\in[1,N]}x_ic_{i1}  \\  
		\cdots\\  
		a_{m_l}\sum_{i\in[1,N]}x_ic_{i1}    
	\end{pmatrix}
\end{eqnarray}

Consider the next $ (l+1)$th layer. The output $O_{i+1}$ is shown as Eq.\ref{eqa8}.

\newcounter{MYtempeqncnt}
\begin{figure*}[]
	\normalsize	\setcounter{MYtempeqncnt}{\value{equation}}
	\begin{equation}
		\label{eqa8}
		\begin{aligned}
			O_{l+1}&=f_{l+1}(O_l)\\&= \begin{pmatrix}  
				w(l+1)_{11}a_1\sum_{i\in[1,N]}x_ic_{i1} +w(l+1)_{21}a_2\sum_{i\in[1,N]}x_ic_{i1} +\dots +w(l+1)_{{m_l}1}a_{m_l}\sum_{i\in[1,N]}x_ic_{i1}  \\  	w(l+1)_{12}a_1\sum_{i\in[1,N]}x_ic_{i1} +w(l+1)_{22}a_2\sum_{i\in[1,N]}x_ic_{i1} +\dots +w(l+1)_{{m_l}2}a_{m_l}\sum_{i\in[1,N]}x_ic_{i1} \\  
				\cdots\\  
				w(l+1)_{1{m_{l+1}}}a_1\sum_{i\in[1,N]}x_ic_{i1} +w(l+1)_{2{m_{l+1}}}a_2\sum_{i\in[1,N]}x_ic_{i1} +\dots +w(l+1)_{{m_{l}}{m_{l+1}}}a_{m_l}\sum_{i\in[1,N]}x_ic_{i1}    
			\end{pmatrix} \\&= \begin{pmatrix}  
				b_1\sum_{i\in[1,N]}x_ic_{i1}  \\  	b_2\sum_{i\in[1,N]}x_ic_{i1}  \\  
				\cdots\\  
				b_{l+1}\sum_{i\in[1,N]}x_ic_{i1}    
			\end{pmatrix} ,\qquad b_j=\sum\nolimits_{i\in[1,m_{l+1}]}a_iw(l+1)_{ij},\; j\in[1,l+1] 
		\end{aligned} 
	\end{equation}
\end{figure*}

It is easy to deduce that in the first group $c_{i1}/\sum c_{i1}=b_jc_{i1}/\sum b_jc_{i1}$. Hence, Corollary 3 holds when $k=l+1$.\par
Therefore, Corollary 3 is true, completing the induction.\par
QED
\subsection{Rationality of Corollary 5}
We use BIC scores to prove the rationality of Corollary 5. Assume that we have already had an appropriate input sequence length $ P $. BIC scores of a certain forecasting model are shown in Eq.\ref{eqa9}/\ref{eqa11} respectively before/after extending length of input sequence ($ B_1/B_2 $). $ m $ is the length of total training data, $ k_1/k_2 $ is the number of parameters, $ P/P' $ denotes the length of input sequence, $ N $ denotes the number of variates and $ L_1/L_2 $ represents the likelihood function as Eq. \ref{eqa10}/\ref{eqa12}. $ k_1/k_2 $ can be defined as multiplication of $ P/P' $ and $ N $.\par
\begin{eqnarray}
	\label{eqa9}
	B_1 =ln(m)k_1-2ln(L_1)
\end{eqnarray}
\begin{eqnarray}
	\label{eqa10}
	k_1=P\cdot N,\; L_1=\frac{1}{m}\sum\nolimits_{i\in [1,m]} \log f(x_i|\theta)
\end{eqnarray}
\begin{eqnarray}
	\label{eqa11}
	B_2 =ln(m)k_2-2ln(L_2)
\end{eqnarray}
\begin{eqnarray}
	\label{eqa12}
	k_2=P'\cdot N,\; L_2=\frac{1}{m}\sum\nolimits_{i\in [1,m]} \log f(x_i|\theta)
\end{eqnarray}
On account of extending the length of input sequence, $ k_2>k_1 $. As shown in our validation experments, the MSE loss of a certain model increases when extending the length of input sequence on the basis of $P$. It means that the likelihood function will decrease in these conditions. Therefore, BIC scores of models are likely to increase if extending the length of input sequence on the basis of $ P $, i.e. it is more possible for models to get overfitting. In summery, Corollary 5 is rational.\par
\section{Concrete Details of RTNet Architecture}
\label{B}
The concrete details of RTNet with end-to-end and contrastive learning based forecasting format are shown in Algorithm 1 and 2 separately in the next page.\par
\begin{table*}[]
	\renewcommand{\arraystretch}{1.5}
	\label{algorithm1}
	\centering
	\begin{tabular}{llll}
		\toprule[1.5pt]
		\multicolumn{4}{l}{\textbf{Algorithm 1 } End-to-end Forecasting Format of RTNet}            \\ \midrule[1pt]
		\multicolumn{2}{l}{AR (Auto-Regressive) network }                                                        & \multicolumn{2}{l}{Time embedding network}                     \\
		\textbf{Input:} Input sequence:  $\{s_i\}_{i=1}^{L_{in}}$            & $B\times L_{in}\times N$                                & Time embedding:  $\{t_i\}_{i=1}^{L_{out}}$                   & $B\times L_{out}\times N$  \\
		1: $\begin{pmatrix} 
			F_1\\F_2\\F_3
		\end{pmatrix}=CPN
		\begin{pmatrix}
			\{s_i\}_{i=1}^{L_{in}} \\ \{s_i\}_{i= L_{in}/2}^{L_{in}} \\ \{s_i\}_{i= 3L_{in}/4}^{L_{in}}
		\end{pmatrix}$ &
		\makecell[l]{$B\times L_{in}D$\\ $B\times L_{in}D/2$\\ $B\times L_{in}D/4$}  & \multirow{2}{*}{$T$=TimeNet ($\{t_i\}_{i=1}^{L_{out}}$)} & \multirow{2}{*}{$B\times L_{out}\times 4D$} \\
		2: $CPN_{out}$ = concat($ F_1,F_2,F_3 $)                                                                              & $B\times 7L_{in}D/8$                                    &                                             &                            \\
		3: $FC_{out}$ = FC($CPN_{out}$)  & $B\times L_{out}N$  & \multirow{2}{*}{$T_{out} = 1\times 1$ Conv($ T $)   }    & \multirow{2}{*}{$B\times L_{out}\times N$ } \\
		4: $AR_{out}$ = Reshape($FC_{out}$)     & $B\times L_{out}\times N$                               &          &                            \\
		5: Output = $AR_{out}+ T_{out}$                                                                                       && &         \\
		\textbf{Ensure:} loss = MSE (Output, Truth)                   & \multicolumn{3}{l}{$ 1 \times 3 $ (The number of pyramid networks)}  \\ 
		\bottomrule[1.5pt]
	\end{tabular}
\end{table*}
\begin{table*}[]
	\renewcommand{\arraystretch}{1.5}
	\label{algorithm2}
	\centering
	\begin{tabular}{llll}
		\toprule[1.5pt]
		\multicolumn{4}{l}{\textbf{Algorithm 2 } Contrastive Learning Based Forecasting Format of RTNet }                                                                     \\
		\midrule[1pt]
		\multicolumn{2}{l}{AR (Auto-Regressive) network }                                                        & \multicolumn{2}{l}{Time embedding network}                     \\
		\textbf{Input:} Input sequence:  $\{s_i\}_{i=1}^{L_{in}}$            & $B\times L_{in}\times N$                                & Time embedding:  $\{t_i\}_{i=1}^{L_{out}}$                   & $B\times L_{out}\times N$  \\
		\textbf{The first stage:}&&&\\
		1: $\begin{pmatrix} 
			F_1\\F_2\\F_3
		\end{pmatrix}=CPN
		\begin{pmatrix}
			\{s_i\}_{i=1}^{L_{in}} \\ \{s_i\}_{i= L_{in}/2}^{L_{in}} \\ \{s_i\}_{i= 3L_{in}/4}^{L_{in}}
		\end{pmatrix}$ &
		\makecell[l]{$B\times L_{in}D$\\ $B\times L_{in}D/2$\\ $B\times L_{in}D/4$}  & \multirow{2}{*}{$T$=TimeNet ($\{t_i\}_{i=1}^{L_{out}}$)} & \multirow{2}{*}{$B\times L_{out}\times 4D$}\\
		2: $CPN_{out}$ = concat($ F_1,F_2,F_3 $)                                                                              & $B\times 7L_{in}D/8$  &&  \\
		\textbf{Ensure:} $ loss_1 $ = Contrastive loss($CPN_{out}$)  & \multicolumn{3}{l}{$ 1 \times 3 $ (The number of pyramid networks)}  \\
		\textbf{The second stage:} &&&\\
		3: $AR_{out}$ = $1\times 1$ Conv ($CPN_{out}$) (detached)   & $B\times L_{out}\times N$                             & $T_{out}$ = $1\times 1$ Conv ($ T $)       & $B\times L_{out}\times N$  \\
		4: Output = $AR_{out}$ + $ T_{out} $                                                                                     &&  &    \\
		\textbf{Ensure:} $ loss_2 $ = MSE (Output, Truth)& \multicolumn{3}{l}{$ 1 \times 3 $ (The number of pyramid networks)}    \\ \bottomrule[1.5pt]                   
	\end{tabular}
\end{table*}
\section{ Supplementary Experiments}
\subsection{Dataset Introduction}
We perform experiments on three real-world benchmark datasets (ETT, WTH and ECL).\par
\textbf{ETT} (Electricity Transformer Temperature) dataset consists of 2-years data of two electric stations, including 1-hour-level datasets (ETTh$ _1 $, ETTh$ _{2} $) and 15-minute-level datasets (ETTm$ _{1} $, ETTm$ _{2} $). Each data point is composed of the target value `OT' (Oil Temperature) and other 6 power load features. We choose ETTh$ _{1} $, ETTh$ _{2} $ and ETTm$ _{1} $ to do experiments. The train/val/test are 12/4/4 months.\par
\textbf{WTH} (Weather) dataset is a 1-hour-level dataset spanning 4 years from 2010 to 2013, including 12 climatological features from nearly 1,600 locations in the U.S. `WetBulbCelsius' is chosen as the target value. The train/val/test is 60\%/20\%/20\%.\par
\textbf{ECL} (Electricity Consuming Load) dataset is composed of the electricity consumption (Kwh) of 321 clients. We follow the setting of Zhou. \cite{informer} who converted the dataset into 1-hour-level of 2 years and set `MT\_320' as the target value. The train/val/test is 60\%/20\%/20\%.\par
\subsection{Validation Experiments}
\subsubsection{Supplement of Multivariate Forecasting}
We further validate the function of cos-relation matrix on WTH where variates are closely related due to priori knowledge. We perform multivariate experiment on RTNet with/without cos-relation matrix and group convolution. MSE results are shown in Tab.\ref{tab5}.\par
It could be seen that the performances of RTNet drop heavily with cos-relation matrix when it comes to short-term forecasting of WTH. However, during long-term forecasting, forecasting performances of RTNet with/without cos-relation matrix are similar. These phenomena illustrate that the application of cos-relation matrix and group convolution will pay the price of worse forecasting performances under the situation of short-term fully-related multivariate forecasting but is still beneficial in long-term situations.\par
\begin{table}[]
	\renewcommand{\arraystretch}{1.1}
	\caption{MSE of Validation Experiment on Cos-relation Matrix Under WTH}
	\label{tab5}
	\centering
	\begin{threeparttable}
		\begin{tabular}[]{ccccc}
			\toprule[1.5pt]
			Baseline& \multicolumn{2}{c}{RTNet(E)}& \multicolumn{2}{c}{RTNet(C)}                \\
			\cmidrule(lr){2-3}	
			\cmidrule(lr){4-5}			
			Pred&Without&With&Without&With  \\
			\midrule[1pt]
			\multirow{2}{*}{24}&		\textit{\textbf{0.298}} &0.358 &  \textit{\textbf{0.296}}&0.363   \\
			&1.1\% & 0.7\% & 0.7\% & 0.9\% \\
			\multirow{2}{*}{48}& \textit{\textbf{0.361}} &0.411 &  \textit{\textbf{0.363}} &0.430  \\
			& 2.6\% & 0.6\% & 1.2\% & 1.5\% \\
			\multirow{2}{*}{168}& \textit{\textbf{0.469}} &0.487 & \textit{\textbf{0.462}} &0.501 \\
			& 4.4\% & 0.7\% & 0.6\% & 0.9\% \\
			\multirow{2}{*}{336}& \textit{\textbf{0.497}} &0.505 & \textit{\textbf{0.490}} &0.521 \\
			& 1.5\% & 0.9\% & 0.7\% & 2.4\% \\
			\multirow{2}{*}{720}&\textit{\textbf{0.505}} &0.507 &  \textit{\textbf{0.500}}&0.529  \\
			& 1.5\% & 1.5\% & 0.6\% & 2.6\%\\
			\bottomrule[1.5pt]
		\end{tabular}
	\end{threeparttable}   
\end{table}
\begin{figure}[]
	\centering
	\includegraphics[width=2.2in]{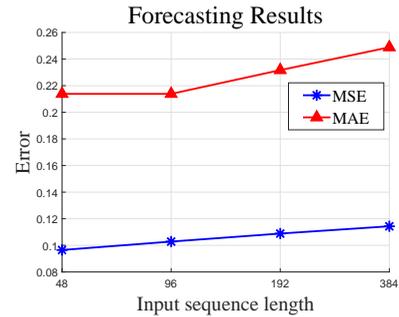}
	\caption{The forecasting errors MSE (blue) and MAE (red) with input length in \{48, 96, 192, 384\}. They increase distinctly when prolonging the input sequence. It means that prolonging input sequence may even lead to over-fitting instead of acquiring long-term dependency.}
	\label{fig15}
\end{figure}
\begin{figure*}[]
	\centering
	\subfloat[]{\includegraphics[height=2in]{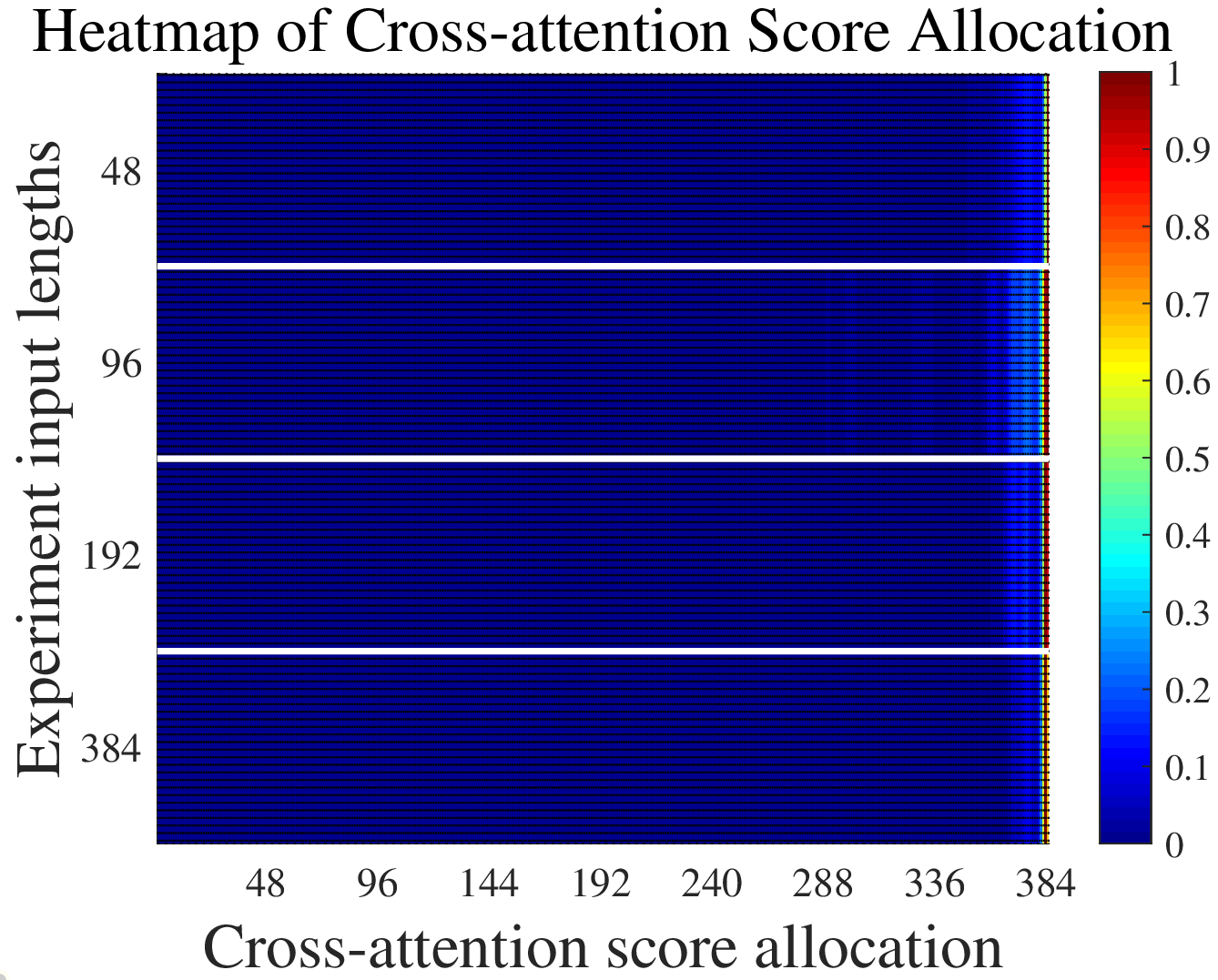}\label{fig14a}}
	\subfloat[]{\includegraphics[height=2in]{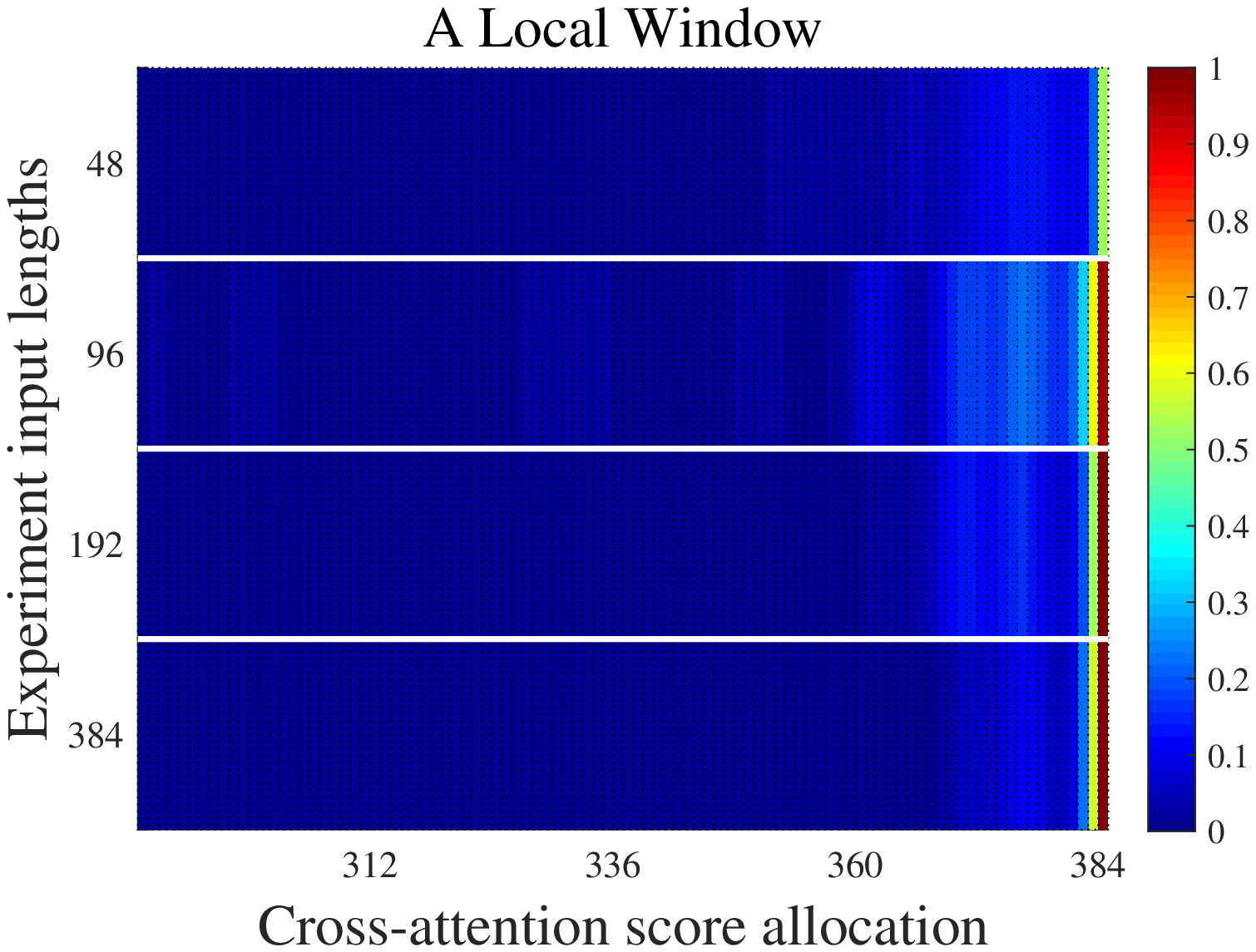}\label{fig14b}}
	\caption{The heatmap of cross-attention score allocation. (a) is the whole heatmap and (b) is a local window which contains scores of last 96 input elements. Owing to different lengths of all input sequences, some non-existing scores at early time stamps of relatively shorter input sequences are padded to zero for better illustration. Obviously, only last few input elements occupy most of attention scores and the number of them seems unchanged with input sequences of different lengths.}
	\label{fig14}
\end{figure*}

\subsubsection{Attention Score Distribution}
We present the forecasting errors of the canonical Transformer model mentioned in Section 4.3.4 of the main text with prolonging the input sequence length in Fig.\ref{fig15}. Moreover, heatmaps of cross-attention score allocation shown in Fig.\ref{fig14} more clearly indicate that the number and position of domination elements will not change with the input sequence length.\par
\begin{table*}[b]
	\renewcommand{\arraystretch}{1.2}
	\caption{MAE of Validation Experiment on Normalization Methods}
	\label{taba1}
	\centering
	\begin{threeparttable}
		\begin{tabular}{ccccccccccc}
			\toprule[1.5pt]
			Baseline & \multicolumn{6}{c}{RTNet}  & \multicolumn{2}{c}{Informer}  & \multicolumn{2}{c}{TS2Vec}  \\
			\cmidrule(lr){2-7}
			\cmidrule(lr){8-9}
			\cmidrule(lr){10-11}
			Format& \multicolumn{3}{c}{E}   & \multicolumn{3}{c}{C}      & \multicolumn{2}{c}{E}  & \multicolumn{2}{c}{C}     \\
			\cmidrule(lr){2-4}
			\cmidrule(lr){5-7}
			\cmidrule(lr){8-9}
			\cmidrule(lr){10-11}
			Pred $\backslash$ Norm     & WN     & BN     & LN     & WN     & BN     & LN     & LN       & None   & BN     & None  \\
			\midrule[1pt]
			\multirow{2}{*}{24} & \textit{\textbf{0.132}} & 0.139  & 0.154  & \textit{\textbf{0.153}} & 0.165  & 0.168  & \textit{\textbf{0.247}}    & 0.253  & 0.166  & \textit{\textbf{0.151}} \\
			&$\pm$2.8\% &$\pm$3.3\%  &$\pm$13.6\% &$\pm$4.9\%  &$\pm$9.7\%  &$\pm$7.1\%  &         &$\pm$12.9\% &$\pm$9.8\%  &      \\
			\multirow{2}{*}{48}  & \textit{\textbf{0.164}} & 0.193  & 0.193  & \textit{\textbf{0.207}}  & 0.212  & 0.225  & 0.319    & \textit{\textbf{0.313}} & 0.213  & \textit{\textbf{0.189}} \\
			&$\pm$3.9\% &$\pm$14.1\% &$\pm$16.5\% &$\pm$7.8\%  &$\pm$9.4\%  &$\pm$12.7\% &          &$\pm$15.6\% &$\pm$13.4\% &       \\
			\multirow{2}{*}{168} & \textit{\textbf{0.204}} & 0.252  & 0.222  & \textit{\textbf{0.229}}& 0.238  & 0.333  & 0.346    & \textit{\textbf{0.328}}  & 0.312  & \textit{\textbf{0.291}} \\
			&$\pm$5.9\% &$\pm$19.3\% &$\pm$13.5\% &$\pm$8.5\%  &$\pm$6.5\%  &$\pm$22.6\% &          &$\pm$11.7\% &$\pm$18.9\% &       \\
			\multirow{2}{*}{336} & \textit{\textbf{0.224}} & 0.272  & 0.341  & \textit{\textbf{0.258}}  & 0.278  & 0.362  & 0.387    & \textit{\textbf{0.340}}  & 0.348  &\textit{\textbf{0.316}} \\
			&$\pm$3.6\% &$\pm$17.0\% &$\pm$6.1\%  &$\pm$11.8\% &$\pm$12.3\% &$\pm$23.0\% &          &$\pm$9.7\%  &$\pm$7.0\%  & \\
			\multirow{2}{*}{720} & \textit{\textbf{0.276}} & 0.286  & 0.321  & \textit{\textbf{0.288}} & 0.341  & 0.391  & 0.435    & \textit{\textbf{0.305}}  & 0.387  & \textit{\textbf{0.345}} \\
			&$\pm$9.9\% &$\pm$11.4\% &$\pm$13.0\% &$\pm$9.5\%  &$\pm$11.1\% &$\pm$16.8\% &          &$\pm$6.7\%  &$\pm$15.0\% &     \\
			\bottomrule[1.5pt] 
		\end{tabular}
	\end{threeparttable}
\end{table*}
\begin{table*}[]
	\renewcommand{\arraystretch}{1.2}
	\caption{MAE of Validation Experiment on Cos-relation Matrix Under ETTh$_1$}
	\label{taba2}
	\centering
	\begin{threeparttable}
		\begin{tabular}{ccccccccc}
			\toprule[1.5pt]
			Baseline & \multicolumn{2}{c}{RTNet(E)}          & \multicolumn{2}{c}{RTNet(C)}          & \multicolumn{2}{c}{Informer(E)}          & \multicolumn{2}{c}{TS2Vec(C)}          \\
			\cmidrule(lr){2-3}
			\cmidrule(lr){4-5}
			\cmidrule(lr){6-7}
			\cmidrule(lr){8-9}
			Pred  & With     & Without & With   & Without & With     & Without & With   & Without \\
			\midrule[1pt]
			\multirow{2}{*}{24} & \textit{\textbf{0.276}}      & 0.478   & \textit{\textbf{0.378}}     & 0.504   & \textit{\textbf{0.447}}     & 0.549   & \textit{\textbf{0.466}}      & 0.531   \\
			&$\pm$9.9\%      &$\pm$6.1\%   &$\pm$0.8\%      &$\pm$7.9\%   &$\pm$6.7\%      &        &$\pm$4.4\%      &         \\
			\multirow{2}{*}{48}    & \textit{\textbf{0.404}}     & 0.567   & \textit{\textbf{0.408}}     & 0.529   & \textit{\textbf{0.483}}      & 0.625   & \textit{\textbf{0.497}}      & 0.555   \\
			&$\pm$0.5\%      &$\pm$13.5\%  &$\pm$1.1\%      &$\pm$4.2\%   &$\pm$5.0\%      &         &$\pm$4.7\%      &         \\
			\multirow{2}{*}{168}   & \textit{\textbf{0.448}}     & 0.738   & \textit{\textbf{0.452}}      & 0.710   & \textit{\textbf{0.551}}     & 0.752   & \textit{\textbf{0.562}}    & 0.639   \\
			&$\pm$1.3\%      &$\pm$7.2\%   &$\pm$1.3\%      &$\pm$6.0\%   &$\pm$7.0\%      &         &$\pm$6.1\%      &         \\
			\multirow{2}{*}{336}   & \textit{\textbf{0.474}}    & 0.875   & \textit{\textbf{0.479}}    & 0.804   & \textit{\textbf{0.578}}   & 0.873   & \textit{\textbf{0.587}}    & 0.728   \\
			&$\pm$2.3\%      &$\pm$14.4\%  &$\pm$1.0\%      &$\pm$7.0\%   &$\pm$8.2\%      &         &$\pm$6.5\%      &         \\
			\multirow{2}{*}{720}   & \textit{\textbf{0.517}}   & 0.868   & \textit{\textbf{0.520}}    & 0.896   & \textit{\textbf{0.694}}  & 0.896   & \textit{\textbf{0.599}}     & 0.799   \\
			&$\pm$1.5\%      &$\pm$5.1\%   &$\pm$0.9\%      &$\pm$5.9\%   &$\pm$7.6\%      &         &$\pm$4.4\%      &       \\
			\bottomrule[1.5pt] 
		\end{tabular}
	\end{threeparttable}
\end{table*}
\begin{table*}[]
	\renewcommand{\arraystretch}{1.2}
	\caption{MAE of Validation Experiment on Cos-relation Matrix Under ETTm$_1$}
	\label{taba3}
	\centering
	\begin{threeparttable}
		\begin{tabular}{ccccccccc}
			\toprule[1.5pt]
			Baseline & \multicolumn{2}{c}{RTNet(E)}  & \multicolumn{2}{c}{RTNet(C)}  & \multicolumn{2}{c}{Informer(E)} & \multicolumn{2}{c}{TS2Vec(C)}    \\
			\cmidrule(lr){2-3}
			\cmidrule(lr){4-5}
			\cmidrule(lr){6-7}
			\cmidrule(lr){8-9}
			Pred     & With       & Without     & With       & Without    & With       & Without & With        & Without   \\
			\midrule[1pt]
			\multirow{2}{*}{24}  & \textit{\textbf{0.291}}     & 0.382   & \textit{\textbf{0.299}}    & 0.346   & \textit{\textbf{0.339}}     & 0.369   & \textit{\textbf{0.401}}    & 0.444   \\
			&$\pm$1.0\%      &$\pm$3.6\%   &$\pm$2.4\%      &$\pm$1.7\%   &$\pm$7.1\%      &         &$\pm$5.4\%      &         \\
			\multirow{2}{*}{48}    & \textit{\textbf{0.331}}    & 0.494   & \textit{\textbf{0.336}}   & 0.411   & \textit{\textbf{0.410}}   & 0.503   & \textit{\textbf{0.457}}  & 0.521   \\
			&$\pm$1.4\%      &$\pm$0.9\%   &$\pm$0.9\%      &$\pm$3.2\%   &$\pm$9.0\%      &         &$\pm$4.8\%      &         \\
			\multirow{2}{*}{96}   & \textit{\textbf{0.356}}    & 0.596   & \textit{\textbf{0.352}}    & 0.470   & \textit{\textbf{0.492}}     & 0.614   & \textit{\textbf{0.462}}     & 0.554   \\
			&$\pm$1.0\%      &$\pm$6.8\%   &$\pm$0.6\%      &$\pm$4.5\%   &$\pm$5.0\%      &         &$\pm$4.9\%      &         \\
			\multirow{2}{*}{288}   & \textit{\textbf{0.398}}    & 0.767   &\textit{\textbf{0.390}}  & 0.583   & \textit{\textbf{0.564}}     & 0.786   & \textit{\textbf{0.512}}  & 0.597   \\
			&$\pm$1.8\%      &$\pm$8.5\%   &$\pm$0.1\%      &$\pm$5.1\%   &$\pm$5.0\%      &         &$\pm$4.6\%      &         \\
			\multirow{2}{*}{672}   & \textit{\textbf{0.436}}  & 0.845   & \textit{\textbf{0.433}} & 0.661   & \textit{\textbf{0.615}}   & 0.926   & \textit{\textbf{0.544}} & 0.653   \\
			&$\pm$1.7\%      &$\pm$6.9\%   &$\pm$1.6\%      &$\pm$0.7\%   &$\pm$4.0\%      &         &$\pm$4.9\%      &    \\
			\bottomrule[1.5pt] 
		\end{tabular}
	\end{threeparttable}
\end{table*}
\begin{table*}[]
	\renewcommand{\arraystretch}{1.2}
	\caption{MAE of Validation Experiment on Cos-relation Matrix Under WTH}
	\label{taba4}
	\centering
	\begin{threeparttable}
		\begin{tabular}[]{ccccc}
			\toprule[1.5pt]
			Baseline& \multicolumn{2}{c}{RTNet(E)}& \multicolumn{2}{c}{RTNet(C)} \\
			\cmidrule(lr){2-3}	
			\cmidrule(lr){4-5}			
			Pred&Without&With&Without&With  \\
			\midrule[1pt]
			\multirow{2}{*}{24} & \textit{\textbf{0.355}}     & 0.410   & \textit{\textbf{0.356}}     & 0.415   \\
			&$\pm$0.5\%      &$\pm$0.4\%   &$\pm$0.6\%      &$\pm$0.7\%   \\
			\multirow{2}{*}{48}  & \textit{\textbf{0.399}}      & 0.449   & \textit{\textbf{0.414}}     & 0.466   \\
			&$\pm$3.0\%      &$\pm$0.4\%   &$\pm$0.9\%      &$\pm$0.6\%   \\
			\multirow{2}{*}{168} &\textit{\textbf{0.481}}   & 0.503   & \textit{\textbf{0.486}}   & 0.515   \\
			&$\pm$2.9\%      &$\pm$0.6\%   &$\pm$0.5\%      &$\pm$0.7\%   \\
			\multirow{2}{*}{336} & \textit{\textbf{0.511}}     & 0.515   & \textit{\textbf{0.503}}     & 0.527   \\
			&$\pm$0.4\%      &$\pm$0.5\%   &$\pm$0.1\%      &$\pm$1.6\%   \\
			\multirow{2}{*}{720} & 0.518      &\textit{\textbf{0.516}}  & \textit{\textbf{0.512}}     & 0.534   \\
			& 0.5\%      & 1.0\%   & 0.2\%      & 1.7\%  \\
			\bottomrule[1.5pt]
		\end{tabular}
	\end{threeparttable}
\end{table*}
\begin{table*}[]
	\renewcommand{\arraystretch}{1.2}
	\caption{MAE of Validation Experiment on Input Sequence Lengths Under WTH}
	\label{taba5}
	\centering
	\begin{threeparttable}
		\begin{tabular}{ccccccccccc}
			\toprule[1.5pt]
			Baseline  & \multicolumn{5}{c}{RTNet(E)}           & \multicolumn{5}{c}{RTNet(C)}  \\
			\cmidrule(lr){2-6}
			\cmidrule(lr){7-11}
			Pred$\backslash$Input & 32       & 48    & 96     & 192   & 384   & 32     & 48    & 96    & 192   & 384   \\
			\midrule[1pt]
			\multirow{2}{*}{24}  & 0.207    & \textit{\textbf{0.207}}& 0.208  & 0.213 & 0.218 & \textit{\textbf{0.212}} & 0.213 & 0.217 & 0.215 & 0.217 \\
			&$\pm$0.7\%    &$\pm$0.3\% &$\pm$0.5\%  &$\pm$2.2\% &$\pm$1.0\% &$\pm$1.6\%  &$\pm$1.0\% &$\pm$2.4\% &$\pm$1.1\% &$\pm$1.6\% \\
			\multirow{2}{*}{48}  & \textit{\textbf{0.253}}   & 0.253 & 0.254  & 0.259 & 0.266 & 0.262  & \textit{\textbf{0.261}} & 0.265 & 0.267 & 0.265 \\
			&$\pm$0.3\%    &$\pm$0.5\% &$\pm$0.0\%  &$\pm$1.4\% &$\pm$1.1\% &$\pm$1.7\%  &$\pm$1.8\% &$\pm$1.0\% &$\pm$1.1\% &$\pm$0.8\% \\
			\multirow{2}{*}{168} & \textit{\textbf{0.320}}    & 0.321 & 0.326  & 0.332 & 0.353 & \textit{\textbf{0.316}}& 0.321 & 0.323 & 0.328 & 0.326 \\
			&$\pm$0.1\%    &$\pm$2.0\% &$\pm$0.5\%  &$\pm$2.6\% &$\pm$3.8\% &$\pm$0.7\%  &$\pm$1.4\% &$\pm$0.8\% &$\pm$0.9\% &$\pm$0.9\% \\
			\multirow{2}{*}{336} & 0.338    & \textit{\textbf{0.336}} & 0.337  & 0.347 & 0.347 & \textit{\textbf{0.332}}  & 0.336 & 0.347 & 0.351 & 0.352 \\
			&$\pm$1.8\%    &$\pm$1.0\% &$\pm$1.1\%  &$\pm$0.1\% &$\pm$0.3\% &$\pm$0.2\%  &$\pm$0.0\% &$\pm$1.2\% &$\pm$2.2\% &$\pm$1.9\% \\
			\multirow{2}{*}{720} & 0.333    & \textit{\textbf{0.333}} & 0.326  & 0.335 & 0.337 & \textit{\textbf{0.331}}  & 0.339 & 0.342 & 0.342 & 0.347 \\
			&$\pm$1.8\%    &$\pm$1.3\% &$\pm$0.4\%  &$\pm$2.8\% &$\pm$1.4\% &$\pm$0.1\%  &$\pm$2.8\% &$\pm$4.0\% &$\pm$3.7\% &$\pm$4.9\% \\
			\hline
			\hline
			Baseline & \multicolumn{5}{c}{Informer(E)} & \multicolumn{5}{c}{TS2Vec(C)} \\
			\cmidrule(lr){2-6}
			\cmidrule(lr){7-11}
			Pred$\backslash$Input& 32       & 48    & 96     & 192   & 384   & 32     & 48    & 96    & 192   & 384   \\
			\midrule[1pt]
			\multirow{2}{*}{24}  & \textit{\textbf{0.230}}    & 0.234 & 0.244  & 0.277 & 0.289 & \textit{\textbf{0.213}} & 0.213 & 0.215 & 0.215 & 0.217 \\
			&$\pm$5.3\%    &$\pm$6.9\% &$\pm$5.7\%  &$\pm$3.2\% &$\pm$4.9\% &$\pm$0.6\%  &$\pm$0.9\% &$\pm$1.1\% &$\pm$0.5\% &$\pm$1.5\% \\
			\multirow{2}{*}{48}  & 0.278    & \textit{\textbf{0.275}} & 0.309  & 0.316 & 0.329 & \textit{\textbf{0.261}}  & 0.263 & 0.262 & 0.262 & 0.264 \\
			&$\pm$5.8\%    &$\pm$1.4\% &$\pm$11.4\% &$\pm$1.9\% &$\pm$4.6\% &$\pm$2.2\%  &$\pm$2.2\% &$\pm$1.7\% &$\pm$1.7\% &$\pm$1.2\% \\
			\multirow{2}{*}{168} & 0.376    & \textit{\textbf{0.362}} & 0.377  & 0.376 & 0.399 & 0.330  & \textit{\textbf{0.326}} & 0.330 & 0.333 & 0.335 \\
			&$\pm$6.1\%    &$\pm$4.3\% &$\pm$0.1\%  &$\pm$1.8\% &$\pm$3.2\% &$\pm$2.6\%  &$\pm$1.5\% &$\pm$1.6\% &$\pm$2.4\% &$\pm$2.9\% \\
			\multirow{2}{*}{336} & \textit{\textbf{0.375}}   & 0.390 & 0.391  & 0.395 & 0.417 & 0.353  &\textit{\textbf{0.351}} & 0.356 & 0.360 & 0.358 \\
			&$\pm$2.7\%    &$\pm$2.6\% &$\pm$3.5\%  &$\pm$2.8\% &$\pm$2.8\% &$\pm$1.9\%  &$\pm$1.9\% &$\pm$2.4\% &$\pm$2.0\% &$\pm$2.6\% \\
			\multirow{2}{*}{720} & \textit{\textbf{0.390}}    & 0.400 & 0.398  & 0.410 & 0.406 & 0.365  & \textit{\textbf{0.361}} & 0.364 & 0.363 & 0.366 \\
			&$\pm$2.5\%    &$\pm$3.4\% &$\pm$2.4\%  &$\pm$5.8\% &$\pm$1.7\% &$\pm$1.7\%  &$\pm$3.2\% &$\pm$2.5\% &$\pm$2.3\% &$\pm$2.4\%\\
			\bottomrule[1.5pt]
		\end{tabular}
	\end{threeparttable}
\end{table*}
\subsubsection{MAE of Validation Experiments}
Tab.\ref{taba1}-\ref{taba5} are MAE results of three validation experiments. We present average results,  error bars of RTNet and the best MAE result of each condition is highlighted in bold and italicized, which are the same as following experiments.\par
\subsection{Comparison Experiment}
\subsubsection{Multivariate Forecasting with ECL}
Though it has been mentioned before that doing multivariate forecasting on ECL is meaningless, we still perform simple experiments with the prediction length 24 to illustrate the robustness and adaptivity of RTNet and the result is shown in Tab.\ref{tab9}.\par
\begin{table*}[]
	\renewcommand{\arraystretch}{1.2}
	\caption{Average Results of Multivariate Forecasting Experiment in ECL with Prediction Length of 24}
	\label{tab9}
	\centering
	\setlength\tabcolsep{1.8pt}
	\begin{threeparttable}
\begin{tabular}{ccccccccccccccccc}
			\toprule[1.5pt]
			Baseline & RTNet& RTNet& CoST & TS2Vec & TNC & MoCo & TCC & CPC & Triplet & TST & SCINet& Informer & LogTrans& TCN & LSTNet & StemGNN \\
			\cmidrule(lr){2-17}
			Format & E & C & C & C & C&C&C&C&C&E & E&E&E&E&E&E \\
			\midrule[1pt]
			MSE   & \textit{\textbf{0.123}}   & 0.130   & 0.136   & 0.287     & 0.354  & 0.288   & 0.345  & 0.403  & 0.564      & 0.311  & 0.147     & 0.312       & 0.297       & 0.305  & 0.356     & 0.439      \\
			MAE   & \textit{\textbf{0.232}}   & 0.240   & 0.242   & 0.375     & 0.423  & 0.374   & 0.425  & 0.459  & 0.578      & 0.396  & 0.260      & 0.387       & 0.374       & 0.384  & 0.419     & 0.388  \\
			\bottomrule[1.5pt]
		\end{tabular}
	\end{threeparttable}
\end{table*}
\begin{table*}[]
	\renewcommand{\arraystretch}{1.2}
	\caption{MSE of Comparison Experiment on Feature-based Baselines During Univariate Forecasting}
	\label{taba6}
	\centering
	\setlength\tabcolsep{2.5pt}
	\begin{threeparttable}
		\begin{tabular}{ccccccccccccc}
			\toprule[1.5pt]
			\multicolumn{2}{c}{\multirow{2}{*}{Dataset\textbackslash{}Baseline}}&\multirow{2}{*}{RTNet(E)}       & \multirow{2}{*}{RTNet(C)}     &\multicolumn{8}{c}{Contrastive Learning} & Feature Engineered \\
			\cmidrule(lr){5-12}
			\cmidrule(lr){13-13}
			& & &  & CoST  & TS2Vec & MoCo  & TCC   & CPC   & TNC   & Triplet & TST   & TSFresh \\
			\midrule[1pt]
			\multirow{5}{*}{ETTh$_1$} & 24  & \textit{\textbf{0.028(+12.7\%)}}& 0.035(+16.4\%) & 0.040  & 0.039  & 0.040  & 0.053 & 0.076 & 0.057 & 0.130    & 0.127 & 0.080    \\
			& 48  & \textit{\textbf{0.040(+25.0\%)}} & 0.059(+39.4\%) & 0.060  & 0.062  & 0.063 & 0.074 & 0.104 & 0.094 & 0.145   & 0.202 & 0.092   \\
			& 168 & \textit{\textbf{0.062(+32.5\%)}} & 0.074(+36.9\%) & 0.097 & 0.142  & 0.122 & 0.133 & 0.162 & 0.171 & 0.173   & 0.491 & 0.097   \\
			& 336 & \textit{\textbf{0.076(+19.0\%)}} & 0.086(+56.2\%) & 0.112 & 0.160   & 0.144 & 0.161 & 0.183 & 0.192 & 0.167   & 0.526 & 0.109   \\
			& 720 & \textit{\textbf{0.097(+46.5\%)}} & 0.109(+44.2\%) & 0.148 & 0.179  & 0.183 & 0.176 & 0.212 & 0.235 & 0.195   & 0.717 & 0.142   \\
			\hline
			\multirow{5}{*}{ETTh$_2$} & 24  & 0.071(+2.2\%)  & \textit{\textbf{0.067(+11.5\%)}}  & 0.079 & 0.091  & 0.095 & 0.111 & 0.109 & 0.097 & 0.160    & 0.134 & 0.176   \\
\end{tabular}
	\end{threeparttable}
\end{table*}
\begin{table*}[]
	\renewcommand{\arraystretch}{1.2}
	\caption{MSE of Comparison Experiment on End-to-end Baselines During Univariate Forecasting}
	\label{taba7}
	\centering
	\setlength\tabcolsep{1.8pt}
	\begin{threeparttable}
		\begin{tabular}{ccccccccccccccc}
			\toprule[1.5pt]
			\multicolumn{2}{c}{Dataset\textbackslash{}Baseline} & RTNet(E) & RTNet(C) & SCINet & Informer & LogTrans & N-BEATS & TCN   & DeepAR & ARIMA & LSTMa & LSTNet & Reformer & Prophet \\
			\midrule[1pt]
			\multirow{5}{*}{ETTh$_1$} & 24  & \textit{\textbf{0.028}}    & 0.035    & 0.097  & 0.098    & 0.103    & 0.094   & 0.075 & 0.107  & 0.086 & 0.094 & 0.114  & 0.172    & 0.093   \\
			& 48  & \textit{\textbf{0.040}}    & 0.059    & 0.096  & 0.158    & 0.167    & 0.210   & 0.227 & 0.162  & 0.133 & 0.175 & 0.193  & 0.228    & 0.150   \\
			& 168 & \textit{\textbf{0.062}}   & 0.074    & 0.110  & 0.183    & 0.207    & 0.232   & 0.316 & 0.239  & 0.364 & 0.210 & 0.236  & 1.460    & 1.194   \\
			& 336 & \textit{\textbf{0.076}}  & 0.086    & 0.147  & 0.222    & 0.230    & 0.232   & 0.306 & 0.445  & 0.428 & 0.556 & 0.590  & 1.728    & 1.509   \\
			& 720 & \textit{\textbf{0.097}}  & 0.109    & 0.233  & 0.269    & 0.273    & 0.322   & 0.390 & 0.658  & 0.613 & 0.635 & 0.683  & 1.948    & 2.685   \\
			\hline
			\multirow{5}{*}{ETTh$_2$} & 24  & 0.071    & \textit{\textbf{0.067}} & 0.236  & 0.093    & 0.102    & 0.198   & 0.109 & 0.080  & 3.538 & 0.080 & 3.554  & 0.135    & 0.179   \\
			& 48  & \textit{\textbf{0.092}} & 0.095    & 0.195  & 0.155    & 0.169    & 0.234   & 0.147 & 0.125  & 3.168 & 0.125 & 3.190  & 0.172    & 0.284   \\
\end{tabular}
	\end{threeparttable}
\end{table*}
\begin{table*}[]
	\renewcommand{\arraystretch}{1.2}
	\caption{MSE of Comparison Experiment on Feature-based Baselines During Multivariate Forecasting}
	\label{taba8}
	\centering
	\setlength\tabcolsep{2.5pt}
	\begin{threeparttable}
		\begin{tabular}{ccccccccccccc}
			\toprule[1.5pt]
			\multicolumn{2}{c}{\multirow{2}{*}{Dataset\textbackslash{}Baseline}}&\multirow{2}{*}{RTNet(E)}       & \multirow{2}{*}{RTNet(C)}     &\multicolumn{8}{c}{Contrastive Learning} & Feature Engineered \\
			\cmidrule(lr){5-12}
			\cmidrule(lr){13-13}
			& & &  & CoST  & TS2Vec & MoCo  & TCC   & CPC   & TNC   & Triplet & TST   & TSFresh \\
			\midrule[1pt]
			\multirow{5}{*}{ETTh$_1$} & 24  & 0.340(+0.2\%)  & \textit{\textbf{0.326(+1.8\%)}}  & 0.386 & 0.590  & 0.623 & 0.766 & 0.728 & 0.708 & 0.942   & 0.735 & 3.858   \\
			& 48  & \textit{\textbf{0.371(+1.0\%)}}  & 0.372(+2.8\%)  & 0.437 & 0.624  & 0.669 & 0.825 & 0.774 & 0.749 & 0.975   & 0.800 & 4.246   \\
			& 168 & \textit{\textbf{0.440(+3.6\%)}}   & 0.451(+2.5\%)  & 0.643 & 0.762  & 0.820 & 0.982 & 0.920 & 0.884 & 1.135   & 0.973 & 3.527   \\
			& 336 & \textit{\textbf{0.485(+7.8\%)}}   & 0.505(+3.6\%)  & 0.812 & 0.931  & 0.981 & 1.099 & 1.050 & 1.020 & 1.187   & 1.029 & 2.905   \\
			& 720 & \textit{\textbf{0.530(+5.1\%)}}  & 0.539(+3.9\%)  & 0.970 & 1.063  & 1.138 & 1.267 & 1.160 & 1.157 & 1.283   & 1.020 & 2.667   \\
			\hline
			\multirow{5}{*}{ETTh$_2$} & 24  & 0.269(+41.2\%) & \textit{\textbf{0.219(+82.9\%) }} & 0.447 & 0.423  & 0.444 & 1.154 & 0.551 & 0.612 & 1.285   & 0.994 & 8.720   \\
\end{tabular}
	\end{threeparttable}
\end{table*}
\begin{table*}[]
	\renewcommand{\arraystretch}{1.2}
	\caption{MSE of Comparison Experiment on End-to-end Baselines During Multivariate Forecasting}
	\label{taba9}
	\centering
	\setlength\tabcolsep{2.5pt}
	\begin{threeparttable}
		\begin{tabular}{cccccccccccc}
			\toprule[1.5pt]
			\multicolumn{2}{c}{Dataset\textbackslash{}Baseline} & RTNet(E) & RTNet(C) & SCINet & Informer & LogTrans & TCN   & LSTMa & LSTNet & Reformer & StemGNN \\
			\midrule[1pt]
			\multirow{5}{*}{ETTh$_1$} & 24  & 0.340    & \textit{\textbf{0.326}}     & 0.505  & 0.577    & 0.686    & 0.767 & 0.536 & 1.293  & 0.887    & 0.614   \\
			& 48  & \textit{\textbf{0.371}}    & 0.372    & 0.459  & 0.685    & 0.766    & 0.713 & 0.616 & 1.456  & 1.159    & 0.748   \\
			& 168 & \textit{\textbf{0.440}}   & 0.451    & 0.510  & 0.931    & 1.002    & 0.995 & 1.058 & 1.997  & 1.686    & 0.663   \\
			& 336 & \textit{\textbf{0.485}}  & 0.505    & 0.607  & 1.128    & 1.362    & 1.175 & 1.152 & 2.655  & 1.919    & 0.927   \\
			& 720 & \textit{\textbf{0.530}}     & 0.539    & 0.746  & 1.215    & 1.397    & 1.453 & 1.682 & 2.143  & 2.177    &---      \\
			\hline
			\multirow{5}{*}{ETTh$_2$} & 24  & 0.269    & \textit{\textbf{0.219}}    & 0.510  & 0.720    & 0.828    & 0.935 & 1.049 & 2.742  & 1.381    & 1.292   \\
			& 48  & 0.481    & \textit{\textbf{0.461}}     & 0.518  & 1.457    & 1.806    & 1.300 & 1.331 & 3.564  & 1.715    & 1.099   \\
\end{tabular}
	\end{threeparttable}
\end{table*}
\begin{table*}[]
	\renewcommand{\arraystretch}{1.2}
	\caption{MAE of Comparison Experiment on Feature-based Baselines During Univariate Forecasting}
	\label{taba10}
	\centering
	\setlength\tabcolsep{2pt}
	\begin{threeparttable}
		\begin{tabular}{ccccccccccccc}
			\toprule[1.5pt]
			\multicolumn{2}{c}{\multirow{2}{*}{Dataset\textbackslash{}Baseline}}&\multirow{2}{*}{RTNet(E)}       & \multirow{2}{*}{RTNet(C)}     &\multicolumn{8}{c}{Contrastive Learning} & Feature Engineered \\
			\cmidrule(lr){5-12}
			\cmidrule(lr){13-13}
			& & &  & CoST  & TS2Vec & MoCo  & TCC   & CPC   & TNC   & Triplet & TST   & TSFresh \\
			\midrule[1pt]
			\multirow{5}{*}{ETTh$_1$} & 24  & \textit{\textbf{0.128(+5.7\%)}}  & 0.146(+10.4\%) & 0.152 & 0.151  & 0.151 & 0.175 & 0.217 & 0.184 & 0.289   & 0.284 & 0.224   \\
			& 48  & \textit{\textbf{0.157(+8.2\%)}}   & 0.191(+17.0\%) & 0.186 & 0.189  & 0.191 & 0.209 & 0.259 & 0.239 & 0.306   & 0.362 & 0.242   \\
			& 168 & \textit{\textbf{0.192(+12.5\%)}} & 0.209(+18.6\%) & 0.236 & 0.291  & 0.268 & 0.284 & 0.326 & 0.329 & 0.336   & 0.596 & 0.253   \\
			& 336 & \textit{\textbf{0.216(+7.4\%)}}  & 0.228(+26.8\%) & 0.258 & 0.316  & 0.297 & 0.320 & 0.351 & 0.357 & 0.333   & 0.618 & 0.263   \\
			& 720 & \textit{\textbf{0.248(+22.1\%) }} & 0.260(+21.1\%) & 0.306 & 0.345  & 0.347 & 0.343 & 0.387 & 0.408 & 0.368   & 0.760 & 0.302   \\
			\hline
			\multirow{5}{*}{ETTh$_2$} & 24  & 0.202(+3.0\%)  & \textit{\textbf{0.194(+7.8\%) }}  & 0.207 & 0.230  & 0.234 & 0.255 & 0.251 & 0.238 & 0.316   & 0.281 & 0.331   \\
\end{tabular}
	\end{threeparttable} 
\end{table*}
\begin{table*}[]
	\renewcommand{\arraystretch}{1.2}
	\caption{MAE of Comparison Experiment on End-to-end Baselines During Univariate Forecasting}
	\label{taba11}
	\centering
	\setlength\tabcolsep{1.8pt}
	\begin{threeparttable}
		\begin{tabular}{ccccccccccccccc}
			\toprule[1.5pt]
			\multicolumn{2}{c}{Dataset\textbackslash{}Baseline} & RTNet(E) & RTNet(C) & SCINet & Informer & LogTrans & N-BEATS & TCN   & DeepAR & ARIMA & LSTMa & LSTNet & Reformer & Prophet \\
			\midrule[1pt]
			\multirow{5}{*}{ETTh$_1$} & 24  & \textit{\textbf{0.128 }}    & 0.146    & 0.253  & 0.247    & 0.259    & 0.238   & 0.210 & 0.280  & 0.190 & 0.232 & 0.272  & 0.319    & 0.241   \\
			& 48  & \textit{\textbf{0.157 }} & 0.191    & 0.245  & 0.319    & 0.328    & 0.367   & 0.402 & 0.327  & 0.242 & 0.322 & 0.358  & 0.395    & 0.300   \\
			& 168 & \textit{\textbf{0.192 }}  & 0.209    & 0.257  & 0.346    & 0.375    & 0.391   & 0.493 & 0.422  & 0.456 & 0.352 & 0.392  & 1.089    & 0.721   \\
			& 336 & \textit{\textbf{0.216 }}   & 0.228    & 0.309  & 0.387    & 0.398    & 0.388   & 0.495 & 0.552  & 0.537 & 0.644 & 0.698  & 0.978    & 1.766   \\
			& 720 & \textit{\textbf{0.248 }} & 0.260    & 0.405  & 0.435    & 0.463    & 0.490   & 0.557 & 0.707  & 0.684 & 0.704 & 0.768  & 1.226    & 3.155   \\
			\hline
			\multirow{5}{*}{ETTh$_2$} & 24  & 0.202    & \textit{\textbf{0.194 }}  & 0.381  & 0.240    & 0.255    & 0.345   & 0.251 & 0.229  & 0.407 & 0.275 & 0.445  & 0.369    & 0.345   \\
			& 48  &\textit{\textbf{0.233 }}  & 0.237    & 0.350  & 0.314    & 0.348    & 0.386   & 0.302 & 0.283  & 0.440 & 0.318 & 0.474  & 0.505    & 0.428   \\
\end{tabular}
	\end{threeparttable}
\end{table*}
\begin{table*}[]
	\renewcommand{\arraystretch}{1.2}
	\caption{MAE of Comparison Experiment on Feature-based Baselines During Multivariate Forecasting}
	\label{taba12}
	\centering
	\setlength\tabcolsep{2pt}
	\begin{threeparttable}
		\begin{tabular}{ccccccccccccc}
			\toprule[1.5pt]
			\multicolumn{2}{c}{\multirow{2}{*}{Dataset\textbackslash{}Baseline}}&\multirow{2}{*}{RTNet(E)}       & \multirow{2}{*}{RTNet(C)}     &\multicolumn{8}{c}{Contrastive Learning} & Feature Engineered \\
			\cmidrule(lr){5-12}
			\cmidrule(lr){13-13}
			& & &  & CoST  & TS2Vec & MoCo  & TCC   & CPC   & TNC   & Triplet & TST   & TSFresh \\
			\midrule[1pt]
			\multirow{5}{*}{ETTh$_1$} & 24  & 0.386(+0.4\%)  & \textit{\textbf{0.375(+1.6\%) }}  & 0.429 & 0.531  & 0.555 & 0.629 & 0.600 & 0.592 & 0.729   & 0.633 & 1.574   \\
			& 48  & \textit{\textbf{0.402(+1.0\%) }}   & 0.403(+2.3\%)  & 0.464 & 0.555  & 0.586 & 0.657 & 0.629 & 0.619 & 0.746   & 0.671 & 1.674   \\
			& 168 & \textit{\textbf{0.442(+2.6\%) }}  & 0.446(+2.6\%)  & 0.582 & 0.639  & 0.674 & 0.731 & 0.714 & 0.699 & 0.825   & 0.768 & 1.500   \\
			& 336 & \textit{\textbf{0.463(+4.6\%) }}  & 0.474(+2.0\%)  & 0.679 & 0.728  & 0.755 & 0.786 & 0.779 & 0.768 & 0.859   & 0.797 & 1.329   \\
			& 720 & \textit{\textbf{0.509(+3.1\%)  }}  & 0.515(+1.8\%)  & 0.771 & 0.799  & 0.831 & 0.859 & 0.835 & 0.830 & 0.916   & 0.798 & 1.283   \\
			\hline
			\multirow{5}{*}{ETTh$_2$} & 24  & 0.330(+14.3\%) & \textit{\textbf{0.307(+36.1\%) }} & 0.502 & 0.489  & 0.495 & 0.838 & 0.572 & 0.595 & 0.911   & 0.779 & 2.311   \\
\end{tabular}
	\end{threeparttable}
\end{table*}
\begin{table*}[]	
	\renewcommand{\arraystretch}{1.2}
	\caption{MAE of Comparison Experiment on End-to-end Baselines During Multivariate Forecasting}
	\label{taba13}
	\centering
	\setlength\tabcolsep{2pt}
	\begin{threeparttable}
		\begin{tabular}{cccccccccccc}
			\toprule[1.5pt]
			\multicolumn{2}{c}{Dataset\textbackslash{}Baseline} & RTNet(E) & RTNet(C) & SCINet & Informer & LogTrans & TCN   & LSTMa & LSTNet & Reformer & StemGNN \\
			\midrule[1pt]
			\multirow{5}{*}{ETTh$_1$} & 24  & 0.386    & \textit{\textbf{0.375 }}     & 0.489  & 0.549    & 0.604    & 0.612 & 0.528 & 0.901  & 0.630    & 0.571   \\
			& 48  & \textit{\textbf{0.402 }}    & 0.403    & 0.467  & 0.625    & 0.757    & 0.617 & 0.577 & 0.960  & 0.750    & 0.618   \\
			& 168 & \textit{\textbf{0.442 }}    & 0.446    & 0.501  & 0.752    & 0.846    & 0.738 & 0.725 & 1.214  & 0.996    & 0.608   \\
			& 336 & \textit{\textbf{0.463 }}     & 0.474    & 0.577  & 0.873    & 0.952    & 0.800 & 0.794 & 1.369  & 1.090    & 0.730   \\
			& 720 &\textit{\textbf{0.509 }}    & 0.515    & 0.667  & 0.896    & 1.291    & 1.311 & 1.018 & 1.380  & 1.218    &---      \\
			\hline
			\multirow{5}{*}{ETTh$_2$} & 24  & 0.330    & \textit{\textbf{0.307 }}     & 0.519  & 0.665    & 0.750    & 0.754 & 0.689 & 1.457  & 1.475    & 0.883   \\
			& 48  & 0\textit{\textbf{0.432 }}   & 0.446    & 0.527  & 1.001    & 1.034    & 0.911 & 0.805 & 1.687  & 1.585    & 0.847   \\
\end{tabular}
	\end{threeparttable}
\end{table*}
\subsubsection{MAE Results of Comparison Experiments}
The whole results of comparison experiments are shown in Tab.\ref{taba6}-\ref{taba13}.  In these tables, we present the best results and attach error bars of RTNet. Note that different from results in validation experiments where average performances of RTNet are shown, we put the best results of RTNet here to highlight its promising forecasting capability. Though average forecasting performance of SCINet is better than RTNet as for multivariate forecasting on ETTh$_2$ shown in Tab.7 of the main text, it is obvious that best results of RTNet outperform other beselines in all situations even compared with SCINet. Moreover, RTNet breaks plenty of accurancy records. For instance, the MSE result of RTNet reaches 0.028 which is the first time below 0.03 under univariate forecasting in ETTh$_1$ with prediction length of 24; reaches 0.088 which is the first time below 0.09 under univariate forecasting in WTH with prediction length of 24 and reaches 0.530 which is the first time below 0.7 under multivariate forecasting in ETTh$_1$ with prediction length of 720.\par
\begin{figure}[]
	\centering
	\includegraphics[width=3in]{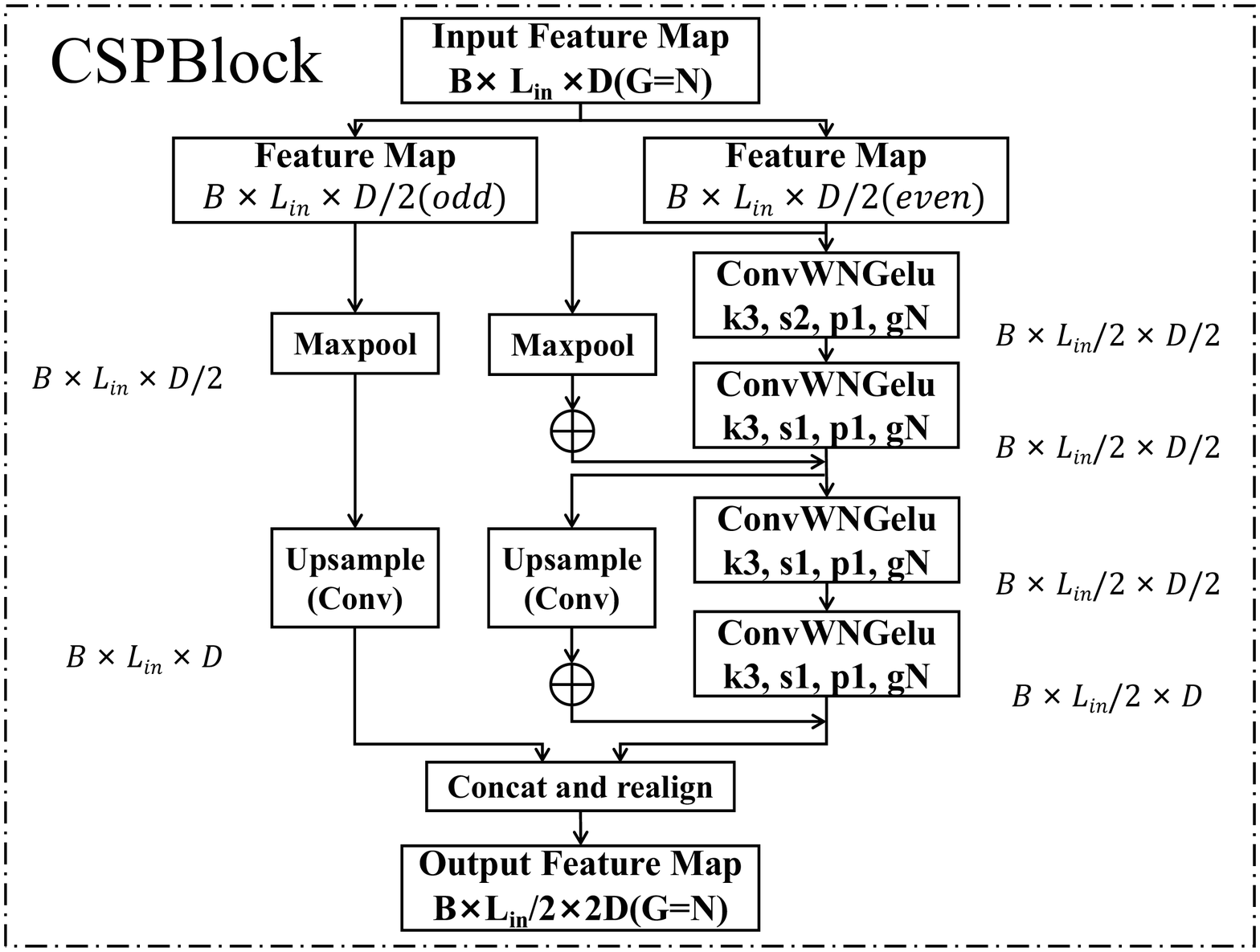}\label{fig17a}
	
	\caption{An example of CSPBlock derived from CSPNet. We split the feature map by odd-even to keep the independence of each group in both two CSP streams. The `concat and realign' block is to reverse the splitting process of odd-even streams. We replace RTBlocks with CSPBlocks to combine RTNet with CSPNet.}
	\label{fig17}
\end{figure}
\begin{figure}[]
	\centering
	\includegraphics[width=2.7in]{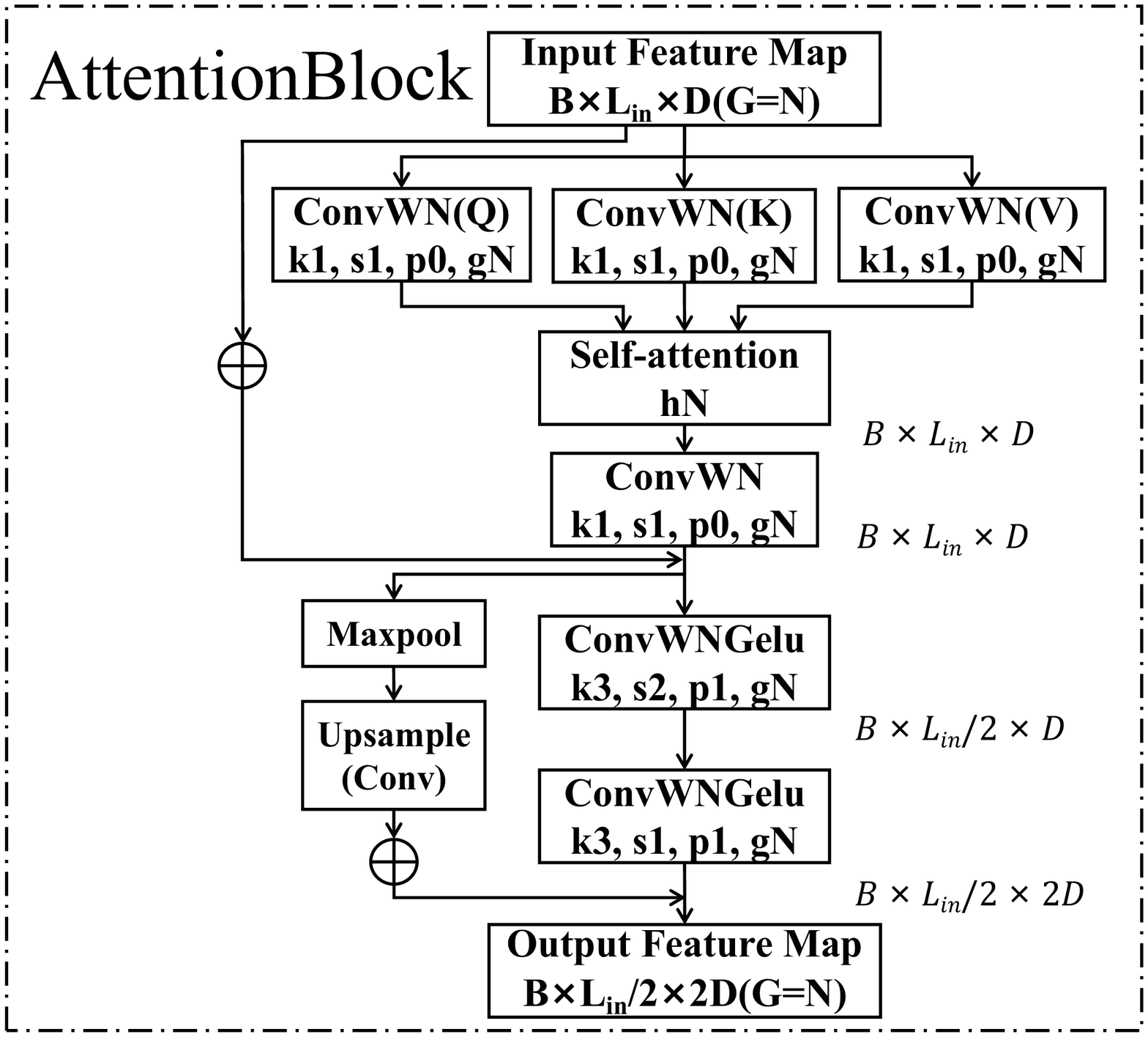}
	\caption{An example of AttentionBlock derived from self-attention. All projection layers are changed to $ 1\times1 $ group convolutional layer and the number of heads in self-attention is set to the number of groups to keep independence of each group. Each self-attention module is accompanied with two group convolutional layers to shorten the sequence and deepen the features. Moreover, they limit the position of each sequence element mediately, hence position embedding is not employed. We replace RTBlocks with AttentionBlocks to combine RTNet with self-attention.}
	\label{fig18}
\end{figure}

\subsection{Ablation Study}
We do two ablation studies including the selection of feature extractors and the way to use time embedding.\par
In the first ablation study, RTNet is separately transformed from ResNet-18\cite{ResNet}/CSPNet\cite{CSPNet}/self-attention\cite{attention2017} to use different feature extractors to validate the adaptability of it and we perform this experiment under the setting of univariate forecasting of ECL as Tab.\ref{tab10} and \ref{taba14}. `FE' refers to the feature extractor. The architectures of RTNet derived from CSPNet/self-attention are shown in Fig.\ref{fig17} and Fig.\ref{fig18}. 
The results from Tab.\ref{tab10} and \ref{taba14} illustrate that RTNet attached with ResNet-18/CSPNet, which are widely used backbones in CV \cite{ResCRNet,ResNetSB,ResInceptionv4IA,CSPYolo,CSPLightweightFA}, and self-attention, which is widely used module in NLP \cite{BERT,HierarchicalAttention,AttentionNLP}, share similar and promising performances. These indicate that RTNet not only has splendid adaptability but also has great potential to couple with SOTA architectures in other AI fields.\par
\begin{table}[]
	\renewcommand{\arraystretch}{1.2}
	\caption{MSE of Ablation Study on  Feature Extractors Under ECL}
	\label{tab10}
	\centering
	\setlength\tabcolsep{1.8pt}
	\begin{threeparttable}
\begin{tabular}{ccccccc}
			\toprule[1.5pt]
			Format                                & \multicolumn{3}{c}{RTNet(E)}       & \multicolumn{3}{c}{RTNet(C) }    \\
			\cmidrule(lr){2-4}
			\cmidrule(lr){5-7}
			Pred\textbackslash{}FE & RTBlock & CSPBlock & Attention & RTBlock & CSPBlock & Attention \\
			\midrule[1pt]
			\multirow{2}{*}{24}                                    & 0.112   & 0.112  & \textit{\textbf{0.108}}    & 0.121   & \textit{\textbf{0.118}}  & 0.121     \\
			& $\pm$0.80\%  & $\pm$2.40\% &$\pm$2.70\%    &$\pm$4.90\%  &$\pm$5.20\% &$\pm$4.00\%    \\
			\multirow{2}{*}{48}                                     & 0.121   & 0.118  & \textit{\textbf{0.115}}    & 0.124   & \textit{\textbf{0.123}}  & 0.130      \\
			&$\pm$0.50\%  &$\pm$1.30\% &$\pm$2.60\%    &$\pm$3.50\%  &$\pm$3.70\% &$\pm$2.70\%    \\
			\multirow{2}{*}{168}                                   & 0.115   & \textit{\textbf{0.111}}  & 0.113     & 0.130    & \textit{\textbf{0.129}} & 0.138     \\
			&$\pm$2.60\%  &$\pm$0.30\% &$\pm$3.60\%    &$\pm$3.60\%  &$\pm$3.60\% &$\pm$5.70\%    \\
			\multirow{2}{*}{336}                                    & \textit{\textbf{0.118}}   & 0.118  & 0.121     & 0.133   & \textit{\textbf{0.131}} & 0.140      \\
			&$\pm$2.50\%  &$\pm$2.60\% &$\pm$2.50\%    &$\pm$6.50\%  &$\pm$3.20\% &$\pm$3.70\%    \\
			\multirow{2}{*}{720}                                    & \textit{\textbf{0.115}}   & 0.117  & 0.123     & 0.134   & \textit{\textbf{0.129}}  & 0.146     \\
			&$\pm$2.20\%  &$\pm$1.40\% &$\pm$4.20\%    &$\pm$5.90\%  &$\pm$2.60\% &$\pm$4.10\% \\
			\bottomrule[1.5pt]  
		\end{tabular}
	\end{threeparttable}
\end{table}
\begin{table}[]
	\renewcommand{\arraystretch}{1.2}
	\caption{MAE of Ablation Study on  Feature Extractors Under ECL}
	\label{taba14}
	\centering
	\setlength\tabcolsep{1.8pt}
	\begin{threeparttable}
		\begin{tabular}{ccccccc}
			\toprule[1.5pt]
			Format                                & \multicolumn{3}{c}{RTNet(E)}       & \multicolumn{3}{c}{RTNet(C) }    \\
			\cmidrule(lr){2-4}
			\cmidrule(lr){5-7}
			Pred\textbackslash{}FE & RTBlock & CSPBlock & Attention & RTBlock & CSPBlock & Attention \\
			\midrule[1pt]
			\multirow{2}{*}{24}  & 0.235   & 0.236  & \textit{\textbf{0.234}} & 0.250   & \textit{\textbf{0.245}}  & 0.254     \\
			&$\pm$0.5\%   &$\pm$1.9\%  &$\pm$1.7\%     &$\pm$4.1\%   &$\pm$3.7\%  &$\pm$2.8\%     \\
			48    & 0.244   & 0.240  & \textit{\textbf{0.239}}    & 0.259   & \textit{\textbf{0.248}} & 0.261     \\
			&$\pm$0.5\%   &$\pm$0.7\%  &$\pm$1.9\%     &$\pm$6.2\%   &$\pm$3.0\%  &$\pm$1.4\%     \\
			168   & 0.246   & \textit{\textbf{0.238}} & 0.242     & 0.267   & \textit{\textbf{0.253}} & 0.270     \\
			&$\pm$1.5\%   &$\pm$0.2\%  &$\pm$2.1\%     &$\pm$7.3\%   &$\pm$2.5\%  &$\pm$3.9\%     \\
			336   & 0.247   & \textit{\textbf{0.243}}  & 0.250     & 0.269   & \textit{\textbf{0.261}}  & 0.275     \\
			&$\pm$1.8\%   &$\pm$1.8\%  &$\pm$1.2\%     &$\pm$6.6\%   &$\pm$2.5\%  &$\pm$2.4\%     \\
			720   & \textit{\textbf{0.246}} & 0.248  & 0.258     & 0.273   & \textit{\textbf{0.261}} & 0.288     \\
			&$\pm$1.7\%   &$\pm$1.0\%  &$\pm$2.9\%     &$\pm$5.9\%   &$\pm$2.2\%  &$\pm$2.6\%   \\
			\bottomrule[1.5pt] 
		\end{tabular}
	\end{threeparttable}
\end{table}
\begin{table*}[]
	\renewcommand{\arraystretch}{1.1}
	\caption{MSE of Ablation Study on Time Embedding}
	\label{tab11}
	\centering
	\setlength\tabcolsep{1.8pt}
	\begin{threeparttable}
		\begin{tabular}{ccccccccccccccccc}
			\toprule[1.5pt]	
			\multicolumn{2}{c}{Dataset}   & \multicolumn{5}{c}{ETTh$ _{1} $}  & \multicolumn{5}{c}{WTH} & \multicolumn{5}{c}{ECL}     \\
			\cmidrule(lr){3-7}
			\cmidrule(lr){8-12}
			\cmidrule(lr){13-17}
			Format& Pred      & 24     & 48     & 168    & 336    & 720    & 24    & 48    & 168   & 336   & 720   & 24    & 48    & 168   & 336   & 720   \\
			\midrule[1pt]
			\multirow{6}{*}{E}&\multirow{2}{*}{w-o}  & 0.037                                          & 0.056                                          & 0.149                                          & 0.145                                          & 0.175                                          & 0.089                                          & 0.131                                          & 0.192                                          & 0.207                                          & 0.199                                          & \textit{\textbf{0.111}} & 0.121                                          & 0.128                                          & 0.137                                          & 0.135                                          \\
			&    & $\pm$17.7\%                                    & $\pm$13.3\%                                    & $\pm$30.0\%                                    & $\pm$18.5\%                                    & $\pm$20.0\%                                    & $\pm$1.1\%                                     & $\pm$1.5\%                                     & $\pm$2.0\%                                     & $\pm$1.2\%                                     & $\pm$2.0\%                                     & $\pm$1.2\%                                     & $\pm$2.8\%                                     & $\pm$2.9\%                                     & $\pm$1.9\%                                     & $\pm$4.2\%                                     \\
\end{tabular}
	\end{threeparttable}
\end{table*}
\begin{table*}[]
	\renewcommand{\arraystretch}{1.1}
	\caption{MAE of Ablation Study on Time Embedding}
	\label{taba15}
	\centering
	\setlength\tabcolsep{1.8pt}
	\begin{threeparttable}
		\begin{tabular}{ccccccccccccccccc}
			\toprule[1.5pt]	
			\multicolumn{2}{c}{Dataset}   & \multicolumn{5}{c}{ETTh$ _{1} $}  & \multicolumn{5}{c}{WTH} & \multicolumn{5}{c}{ECL}     \\
			\cmidrule(lr){3-7}
			\cmidrule(lr){8-12}
			\cmidrule(lr){13-17}
			Format& Pred      & 24     & 48     & 168    & 336    & 720    & 24    & 48    & 168   & 336   & 720   & 24    & 48    & 168   & 336   & 720   \\
			\midrule[1pt]
			\multirow{6}{*}{E}&\multirow{2}{*}{w-o}      & 0.150  & 0.184  & 0.310  & 0.307  & 0.340  & \textit{\textbf{0.207}}& \textit{\textbf{0.253}} & 0.321 & 0.336 & \textit{\textbf{0.333}} & 0.236 & 0.245 & 0.262 & 0.274 & 0.276 \\
			& &$\pm$8.3\%  &$\pm$6.2\%  &$\pm$18.1\% &$\pm$9.6\%  &$\pm$12.3\% &$\pm$0.3\% &$\pm$0.5\% &$\pm$2.0\% &$\pm$1.0\% &$\pm$1.3\% &$\pm$0.7\% &$\pm$1.4\% &$\pm$2.0\% &$\pm$1.5\% &$\pm$2.7\% \\
			&\multirow{2}{*}{w-i}       & 0.230  & 0.321  & 0.419  & 0.450  & 0.503  & 0.207 & 0.253 & \textit{\textbf{0.317}} & \textit{\textbf{0.327}} & 0.334 & ---      & ---      & ---      & ---      & ---      \\
			& &$\pm$11.4\% &$\pm$17.8\% &$\pm$35.0\% &$\pm$32.7\% &$\pm$13.6\% &$\pm$0.9\% &$\pm$1.5\% &$\pm$2.4\% &$\pm$0.1\% &$\pm$0.9\% & ---      & ---      & ---      & ---      & ---      \\
			&\multirow{2}{*}{Without}   & \textit{\textbf{0.132}}  & \textit{\textbf{0.164}}  & \textit{\textbf{0.204}} & \textit{\textbf{0.224}} & \textit{\textbf{0.276}}  & 0.217 & 0.265 & 0.357 & 0.396 & 0.462 & \textit{\textbf{0.235}} &\textit{\textbf{0.244}} & \textit{\textbf{0.246}} & \textit{\textbf{0.247}} & \textit{\textbf{0.246}} \\
			& &$\pm$2.8\%  &$\pm$3.9\%  &$\pm$5.9\%  &$\pm$3.6\%  &$\pm$9.9\%  &$\pm$2.0\% &$\pm$0.8\% &$\pm$1.6\% &$\pm$0.7\% &$\pm$0.3\% &$\pm$0.5\% &$\pm$0.5\% &$\pm$1.5\% &$\pm$1.8\% &$\pm$1.7\% \\
\end{tabular}
	\end{threeparttable}
\end{table*}
\begin{table*}[]
	\renewcommand{\arraystretch}{1.1}
	\caption{Parameter Sensitivity Results of Numbers of Pyramid Networks and RTblocks}
	\label{taba16}
	\centering
	\setlength\tabcolsep{1.8pt}
	\renewcommand{\multirowsetup}{\centering}	
	\begin{threeparttable}
		\begin{tabular}{ccccccccccccc}
			\toprule[1.5pt]
			$ R $ & \multicolumn{4}{c}{1}  & \multicolumn{4}{c}{2}       & \multicolumn{4}{c}{3}\\
			\cmidrule(lr){2-5}
			\cmidrule(lr){6-9}\cmidrule(lr){10-13}
			Format & \multicolumn{2}{c}{E}          & \multicolumn{2}{c}{C}  & \multicolumn{2}{c}{E}          & \multicolumn{2}{c}{C}  & \multicolumn{2}{c}{E}          & \multicolumn{2}{c}{C}            \\
			\cmidrule(lr){2-3}
			\cmidrule(lr){4-5}\cmidrule(lr){6-7}
			\cmidrule(lr){8-9}
			\cmidrule(lr){10-11}\cmidrule(lr){12-13}
			$ P $&MSE&MAE&MSE&MAE&MSE&MAE&MSE&MAE&MSE&MAE&MSE&MAE\\
			\midrule[1pt]
			\multirow{2}{*}{1}                      & 0.157      & 0.311      & 0.172      & 0.321      & 0.156       & 0.312      & 0.171       & 0.322      & 0.151                                          & 0.307                                          & 0.181       & 0.332       \\
			& $\pm$9.4\% & $\pm$4.8\% & $\pm$8.6\% & $\pm$3.6\% & $\pm$16.5\% & $\pm$9.0\% & $\pm$14.1\% & $\pm$6.3\% & $\pm$11.7\%                                    & $\pm$5.9\%                                     & $\pm$20.7\% & $\pm$10.4\% \\
			\multirow{2}{*}{2}  & ---        & ---        & ---        & ---        & 0.153       & 0.309      & 0.151       & 0.303      & 0.155                                          & 0.309                                          & 0.157       & 0.310        \\
			& ---        & ---        & ---        & ---        & $\pm$10.0\% & $\pm$5.4\% & $\pm$6.7\%  & $\pm$2.8\% & $\pm$14.9\%                                    & $\pm$7.3\%                                     & $\pm$8.1\%  & $\pm$3.4\%  \\
			\multirow{2}{*}{3}                           & ---        & ---        & ---        & ---        & ---         & ---        & ---         & ---        & \textit{\textbf{0.146}} & \textit{\textbf{0.300}} & 0.156       & 0.310        \\
\end{tabular}
	\end{threeparttable}
\end{table*}
\begin{table*}[]
	\renewcommand{\arraystretch}{1.1}
	\caption{Parameter Sensitivities Results of $\alpha$ and $\theta$}
	\label{taba17}
	\centering
	\setlength\tabcolsep{1.8pt}
	\renewcommand{\multirowsetup}{\centering}	
	\begin{threeparttable}
		\begin{tabular}{ccccccccccccccccc}
			\toprule[1.5pt]Parameter   & \multicolumn{8}{c}{$\alpha$} & \multicolumn{8}{c}{$\theta$}         \\
			\cmidrule(lr){2-9}\cmidrule(lr){10-17}Value & \multicolumn{2}{c}{8}  &     \multicolumn{2}{c}{6}    &            \multicolumn{2}{c}{4}   &                                                \multicolumn{2}{c}{2}   &                                                \multicolumn{2}{c}{15}     &  \multicolumn{2}{c}{30}   & \multicolumn{2}{c}{60}              & \multicolumn{2}{c}{89}       \\
			\cmidrule(lr){2-3}\cmidrule(lr){4-5}
			\cmidrule(lr){6-7}\cmidrule(lr){8-9}
			\cmidrule(lr){10-11}\cmidrule(lr){12-13}
			\cmidrule(lr){14-15}\cmidrule(lr){16-17}
			Format& MSE      & MAE        & MSE   & MAE        & MSE   & MAE                                            & MSE   & MAE                                            & MSE      & MAE         & MSE   & MAE                                            & MSE   & MAE         & MSE   & MAE         \\
			\midrule[1pt]
			\multirow{2}{*}{E}                                         & 0.160    & 0.315      & 0.151 & 0.307      & \textit{\textbf{0.146}} & \textit{\textbf{0.300}} & 0.158 & 0.313                                          & 1.409    & 0.703       & \textit{\textbf{1.357}} & \textit{\textbf{0.703}} & 1.735 & 0.938       & 2.049 & 1.057       \\
			& $\pm$16.9\%    & $\pm$8.3\% & $\pm$12.0\% & $\pm$6.0\% & $\pm$5.0\%  & $\pm$2.0\%                                     & $\pm$16.3\% & $\pm$8.4\%                                     & $\pm$20.8\%    & $\pm$8.4\%  & $\pm$7.2\%  & $\pm$3.4\%                                     & $\pm$22.0\% & $\pm$9.0\%  & $\pm$17.8\% & $\pm$10.1\% \\
\end{tabular}
	\end{threeparttable}
\end{table*}

In addition, we perform another ablation study on the way to use time embedding. To examine the adaptability and robustness of decoupled time embedding, we do this experiment under the setting of univariate forecasting of ETTh$ _{1} $ and WTH where adding time embedding or not affects heavily as shown in Tab.\ref{tab11}. The MAE results are shown in Tab.\ref{taba15}. Besides, to validate the analysis of time-related properties of different datasets in Section 4.1 of the main text, we do extra experiment under the setting of univariate forecasting of ECL where adding time embedding or not matters a little as shown in Tab.\ref{tab11} and \ref{taba15}. `w-o' refers to the decoupled time embedding method proposed by us; `w-i' is the conventional time embedding method which adds time embedding into input sequence; `without' denotes models without time embedding. \par
Combined with their different occasions mentioned in Section 3.3 of the main text, we analyze the results of experiments on ETTh$ _{1} $ and WTH separately. As for ETT where time embedding is not relative to normal variates on account of the analysis in Section 4.1, the performances of RTNet with two time embedding methods all drop compared with RTNet without time embedding. Nevertheless, the performance of RTNet using the decoupled time embedding exceeds itself with time embedding in input sequence a lot under the end-to-end format. While under the contrastive learning based format, the performance of RTNet using the decoupled time embedding is only slightly inferior to itself with time embedding in input sequence. In fact, during the second stage in contrastive learning based format, the time embedding network is more complex than the one linear projection layer whose input originates from the first stage. Therefore, when handling datasets like ETT whose variates are time-embedding irrelevant, the disadvantage brought by unnecessary time embedding will influence more heavily, which neutralizes the benefits brought by decoupled method. However, it is obvious that RTNet with end-to-end format overcomes this deficiency in that it has only one stage for auto-regressive inference. As to WTH where time embedding is truly relative to normal variates because of priori knowledge, the performances of RTNet using the decoupled time embedding or the conventional method are half a dozen the other where the former one generally outperforms a little. This phenomenon shows that both two methods are feasible for datasets like WTH whose variates are time-embedding relevant. Meanwhile, they all surpass RTNet without time embedding, which proves the analysis of WTH in Section 4.1. In summary, decoupled time embedding is more flexible and robust on the whole. Besides, similar forecasting performances on ECL (Tab.\ref{tab11} and \ref{taba15}) of RTNet with/without time embedding verifies the previous analysis in Section 4.1 that time embedding is not essential for ECL.\par
\subsection{Parameter Sensitivity}
We examine the sensitivity of core hyper-parameters we proposed in this group of experiments under ETTh$ _{2} $ and choose 168 as the prediction length.\par
Above all, we check out the connection between the number of pyramid networks $ P $ and the number of RTBlocks $ R $ in the dominant pyramid network through univariate forecasting. The results are shown in Tab.\ref{taba16}. It is obvious that stacking more pyramid networks and RTBlocks will improve the performance of RTNet. This phenomenon demonstrates the function of CPN and RTBlocks.\par
Moreover, we experiment on the effect of the maximum overlap degree of adjacent input sequences, i.e. $(1 - 1/ \alpha) $ in Section 3.5 of the main text, via univariate forecasting shown in Tab.\ref{taba17}. It could be observed that $\alpha=4$ is more suitable for RTNet with end-to-end format and $\alpha=2$ is more appropriate for RTNet with contrastive learning based format. It means that $\alpha$ needs to be selected cautiously for different formats. \par
Furthermore, we examine the effect of threshold $ \theta $ in cos-relation matrix, which determines whether the values in cos-relation matrix shall be padded into zero, through multivariate forecasting on ETTh$_2$. The results are shown in Tab.\ref{taba17}. Obviously, performances of RTNet with end-to-end/contrastive learning based format are improved after choosing a more appropriate $ \theta $, e.g. $ 30\degree $. Meanwhile, the performance will drop when $\theta\geq 60\degree $. Thus, it could be deduced that variates with weak relevance with a certain other variate should also donate small or even none contributions to its forecasting.\par

\section{Hyper-parameters and Settings}
\label{D}
\begin{table*}[]
	\renewcommand{\arraystretch}{1.2}
	\caption{Hyper-parameters and Settings}
	\label{taba21}
	\centering
	\setlength\tabcolsep{1.8pt}
	\begin{threeparttable}
		\begin{tabular}{c|ccccc}
			\toprule[1.5pt]
			\multirow{2}{*}{Hyper-Parameters/Settings } & \multicolumn{5}{c}{Values/Mechanisms}   \\
			\cmidrule{2-6}
			& \multicolumn{1}{c|}{ETTh$ _1 $}&\multicolumn{1}{c|}{ETTm$ _{1} $ }           & \multicolumn{1}{c|}{ETTh$ _{2} $} & \multicolumn{1}{c|}{WTH} & ECL \\
			\midrule[1pt] 
			Input sequence length   & \multicolumn{1}{c|}{168}   &\multicolumn{1}{c|}{336}  & \multicolumn{1}{c|}{168}  & \multicolumn{1}{c|}{48}  & 336 \\
			\hline
			The number of pyramid networks                                           & \multicolumn{3}{c|}{3}   & \multicolumn{1}{c|}{4}   & 3   \\
			\hline
			The number of RTBlocks in the main pyramid network   &\multicolumn{3}{c|}{3}  & \multicolumn{1}{c|}{4}   & 3   \\
			\hline
			The number of TimeBlocks in time embedding network     &\multicolumn{5}{c}{2}  \\
			\hline
			The kernel size of convolutional layers   &\multicolumn{5}{c}{3}  \\
			\hline
			Threshold angle $\theta$ of cos-relation matrix  & \multicolumn{2}{c|}{45$\degree$}                  & 30$\degree$    & \multicolumn{1}{|c}{---}& \multicolumn{1}{|c}{45$\degree$} \\
			\hline
			Optimizer    & \multicolumn{5}{c}{Adam}    \\
			\hline
			Dropout   & \multicolumn{3}{c|}{0.1} & \multicolumn{1}{c|}{0}  & 0.1 \\
			\hline
			Learning rate                                                            &\multicolumn{5}{c}{1e-04}   \\
			\hline
			End-to-end batch size of train input data                                & \multicolumn{5}{c}{16}  \\
			\hline
			Self-supervised batch size of train input data of the first stage        & \multicolumn{5}{c}{64} \\
			\hline
			Self-supervised batch size of train input data of the second stage       & \multicolumn{5}{c}{16}  \\
			\hline
			Supremum data augmentation amplification   & \multicolumn{5}{c}{0.2} \\
			\hline
			The number of sequences with data augmentation, including initial sequences               & \multicolumn{5}{c}{4} \\
			\hline
			$ \alpha $, controling the maximum overlap degree of adjacent input sequences & \multicolumn{5}{c}{4}  \\
			\hline
			Device &\multicolumn{5}{c}{A single NVIDIA GeForce RTX 3090 24GB GPU} \\
			\bottomrule[1.5pt]
		\end{tabular}
	\end{threeparttable}
\end{table*}

Hyper-parameters and settings of RTNet under each dataset are shown in Tab.\ref{taba21}.\par
\ifCLASSOPTIONcaptionsoff
  \newpage
\fi

\bibliographystyle{IEEEtran}
\bibliography{reference.bib}
\vfill

\end{document}